\documentclass[11pt]{article}

\usepackage{acl}

\usepackage{times}
\usepackage{latexsym}
\usepackage{hyperref}
\usepackage{color, colortbl}
\definecolor{Gray}{gray}{0.9}
\definecolor{high}{HTML}{c2d9d6}
\definecolor{low}{HTML}{f9f0c8}
\definecolor{total}{HTML}{dbeda5}
\definecolor{hard}{HTML}{ffebad}
\usepackage{array}
\usepackage{booktabs}
\usepackage{pifont}
\usepackage{multirow}
\usepackage{amsmath}
\usepackage{amssymb}
\usepackage{amsfonts}
\usepackage{graphicx}
\usepackage{subcaption}
\usepackage{enumerate}
\usepackage{diagbox}
\usepackage{mathtools}
\usepackage{paralist}
\usepackage{tabularx}
\usepackage{lipsum}
\usepackage{ragged2e}
\usepackage{makecell}
\usepackage{siunitx}
\usepackage[table]{xcolor}
\usepackage[most]{tcolorbox}
\definecolor{lightgraytext}{gray}{0.5}
\definecolor{markgreen}{HTML}{009E3D}
\definecolor{markred}{HTML}{D60000}
\definecolor{markorange}{HTML}{F2A000}
\usepackage{arydshln}
\usepackage{algorithm}
\usepackage{algpseudocode}
\usepackage{enumitem}
\usepackage{nicefrac}

\NewDocumentCommand{\qing}
{ mO{} }{\textcolor{blue}{\textsuperscript{\textit{Qing}}\textsf{\textbf{\small[#1]}}}}
\newcommand{\ours}{EvolvingWorld}
\newcommand{\cmark}{\textcolor{markgreen}{\ding{51}}}
\newcommand{\xmark}{\textcolor{markred}{\ding{55}}}
\newcommand{\partialmark}{\textcolor{markorange}{\ensuremath{\bullet}}}
\newcommand{\yue}[1]{\textcolor{brown}{#1}}

\usepackage[T1]{fontenc}

\usepackage[utf8]{inputenc}

\usepackage{microtype}

\usepackage{inconsolata}
\usepackage{listings}
\title{EvolvingWorld: An Open-Schema Framework for Co-Evolving Role-Play Agents and World Model in Interactive Literary World}

\author{
\textbf{Qing Zong}\textsuperscript{1},
\textbf{Yue Guo}\textsuperscript{2},
\textbf{Mengxin Yang}\textsuperscript{3},
\textbf{Yiwen Guo}\textsuperscript{4},
\textbf{Yangqiu Song}\textsuperscript{1} \\
\textsuperscript{1}Hong Kong University of Science and Technology \quad
\textsuperscript{2}LIGHTSPEED \\
\textsuperscript{3}Huazhong University of Science and Technology \quad
\textsuperscript{4}Independent Researcher \\
\texttt{qzong@cse.ust.hk}
}

\begin{document}
\maketitle

\begin{abstract}
This paper introduces \textbf{\ours{}}, a framework and benchmark for character and world co-evolution in interactive literary worlds. 
Existing systems either treat interactive literary simulation as static persona imitation or isolated scene generation, failing to capture how characters and worlds evolve together over time. To address this, \ours{} models literary simulation as a long-horizon process where characters interact, scenes progress, and character and world states are persistently updated. Unlike prior systems relying on fixed schemas,  \ours{} adopts an open-schema framework to support simulation across diverse literary worlds. The framework consists of two coupled modules: a \textbf{Character Agent} for multi-character role-play and persistent profile evolution, and an LLM-based \textbf{World Model} for global and location/entity-level state maintenance and scene progression.
Based on this architecture, we formulate 7 trainable tasks for scene initialization, interaction generation, and state update. We construct a dataset from 57 books, producing 138,596 supervised training samples and 222 snapshots for testing. Furthermore, we introduce a trajectory-level LLM-as-Judge evaluation protocol spanning 10 dimensions and 20 metrics. Experiments show that \ours{} can improve long-horizon simulation by effectively maintaining persistent, coherent character and world development. 
\footnote{\url{https://github.com/HKUST-KnowComp/EvolvingWorld}} 
\end{abstract}

\section{Introduction}

\begin{figure}[t]
    \centering
    \includegraphics[width=0.9\columnwidth]{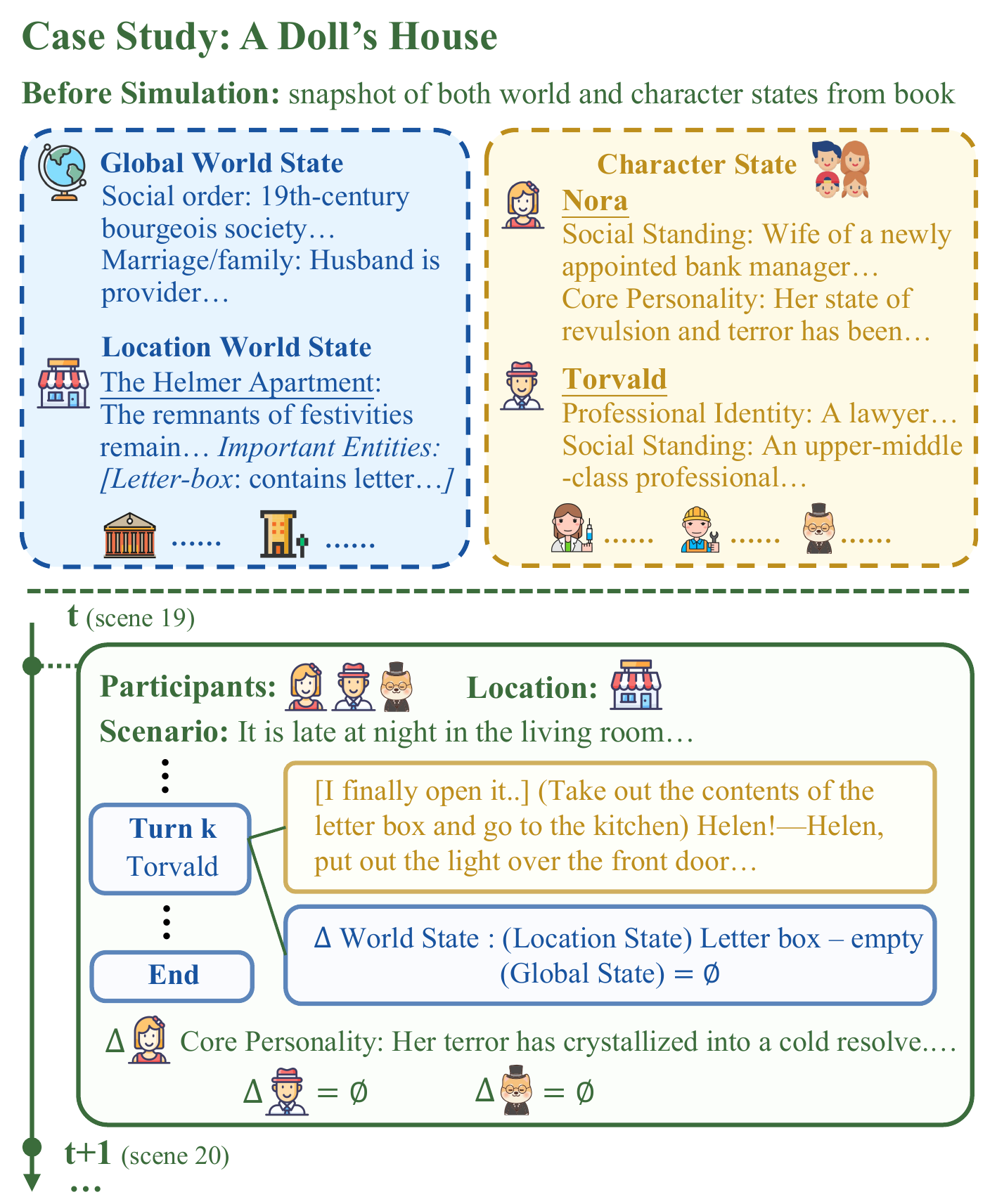}
    \vspace{-0.1in}
\caption{A simulation case from \ours{}. Starting from a snapshot derived from a book, multiple characters interact as both character and world states evolve.}
    \label{fig:casestudy}
\vspace{-0.2in}
\end{figure}

\begin{table*}[t]
\vspace{-0.1in}
\centering
\small
\setlength{\tabcolsep}{2.0pt}
\resizebox{\textwidth}{!}{%
\begin{tabular}{lcccccccccccccccc}
\toprule
\multirow{3}{*}{\textbf{Framework}} &
\multicolumn{3}{c}{\textbf{Character}} &
\multicolumn{2}{c}{\textbf{Interaction}} &
\multicolumn{5}{c}{\textbf{Message Components}} &
\multicolumn{5}{c}{\textbf{World}} &
\multicolumn{1}{c}{\textbf{Data}} \\
\cmidrule(lr){2-4}
\cmidrule(lr){5-6}
\cmidrule(lr){7-11}
\cmidrule(lr){12-16}
\cmidrule(lr){17-17}
&
\textbf{Profile} &
\textbf{Prof. Upd.} &
\textbf{Open-Sch.} &
\textbf{Multi-Char.} &
\textbf{Init Scene} &
\textbf{Speech} &
\textbf{Thought} &
\textbf{Action} &
\textbf{Env.} &
\textbf{Group Act} &
\textbf{Multi-W.} &
\textbf{Global} &
\textbf{Glob. Upd.} &
\textbf{Open-Sch.} &
\textbf{Loc./Ent.} &
\textbf{Training} \\
\midrule
ChatHaruhi~\citep{ChatHaruhi} & \cmark & \xmark & \xmark & \xmark & \xmark & \cmark & \xmark & \xmark & \xmark & \xmark & \cmark & \xmark & \xmark & \xmark & \xmark & \cmark \\
CharacterLLM~\citep{CharacterLLM} & \cmark & \xmark & \xmark & \xmark & \xmark & \cmark & \xmark & \xmark & \xmark & \xmark & \cmark & \xmark & \xmark & \xmark & \xmark & \cmark \\
HPD~\citep{HPD} & \cmark & \xmark & \xmark & \cmark & \xmark & \cmark & \xmark & \xmark & \xmark & \xmark & \xmark & \xmark & \xmark & \xmark & \xmark & \cmark \\
RoleLLM~\citep{RoleLLM} & \cmark & \xmark & \xmark & \xmark & \xmark & \cmark & \xmark & \xmark & \xmark & \xmark & \cmark & \xmark & \xmark & \xmark & \xmark & \cmark \\
CharacterGLM~\citep{CharacterGLM} & \cmark & \xmark & \xmark & \xmark & \xmark & \cmark & \xmark & \cmark & \xmark & \xmark & \cmark & \xmark & \xmark & \xmark & \xmark & \cmark \\
CharacterEval~\citep{CharacterEval} & \cmark & \xmark & \xmark & \xmark & \xmark & \cmark & \xmark & \cmark & \xmark & \xmark & \cmark & \xmark & \xmark & \xmark & \xmark & \cmark \\
CharacterBench~\citep{CharacterBench} & \cmark & \xmark & \xmark & \xmark & \xmark & \cmark & \xmark & \xmark & \xmark & \xmark & \cmark & \xmark & \xmark & \xmark & \xmark & \cmark \\
MMRole~\citep{MMRole} & \cmark & \xmark & \xmark & \cmark & \xmark & \cmark & \xmark & \xmark & \xmark & \xmark & \cmark & \xmark & \xmark & \xmark & \xmark & \cmark \\
DITTO~\citep{DITTO} & \cmark & \xmark & \xmark & \xmark & \xmark & \cmark & \xmark & \xmark & \xmark & \xmark & \cmark & \xmark & \xmark & \xmark & \xmark & \cmark \\
SimsChat~\citep{SimsChat} & \cmark & \xmark & \xmark & \xmark & \xmark & \cmark & \xmark & \xmark & \xmark & \xmark & \xmark & \xmark & \xmark & \xmark & \xmark & \cmark \\
BeyondDialogue~\citep{BeyondDialogue} & \cmark & \xmark & \xmark & \xmark & \xmark & \cmark & \xmark & \xmark & \xmark & \xmark & \cmark & \xmark & \xmark & \xmark & \xmark & \cmark \\
Crab~\citep{Crab} & \cmark & \xmark & \xmark & \xmark & \xmark & \cmark & \xmark & \cmark & \xmark & \xmark & \cmark & \xmark & \xmark & \xmark & \xmark & \cmark \\
TailorRPA~\citep{TailorRPA} & \cmark & \xmark & \xmark & \xmark & \xmark & \cmark & \xmark & \xmark & \xmark & \xmark & \cmark & \xmark & \xmark & \xmark & \xmark & \cmark \\
CoSER~\citep{CoSER} & \cmark & \xmark & \xmark & \cmark & \xmark & \cmark & \cmark & \cmark & \cmark & \xmark & \cmark & \xmark & \xmark & \xmark & \xmark & \cmark \\
AdaMARP~\citep{AdaMARP} & \cmark & \xmark & \xmark & \cmark & \cmark & \cmark & \cmark & \cmark & \cmark & \xmark & \cmark & \xmark & \xmark & \xmark & \xmark & \cmark \\
BookWorld~\citep{BookWorld} & \cmark & \partialmark & \xmark & \cmark & \cmark & \cmark & \cmark & \cmark & \cmark & \xmark & \cmark & \cmark & \partialmark & \xmark & \partialmark & \xmark \\
GenerativeAgents~\citep{GenerativeAgents} & \cmark & \partialmark & \xmark & \cmark & \partialmark & \cmark & \cmark & \cmark & \cmark & \xmark & \xmark & \xmark & \xmark & \xmark & \cmark & \xmark \\
LARP~\citep{LARP} & \cmark & \xmark & \xmark & \cmark & \xmark & \cmark & \cmark & \cmark & \cmark & \xmark & \xmark & \xmark & \xmark & \xmark & \cmark & \cmark \\
CharacterBox~\citep{CharacterBox} & \cmark & \partialmark & \xmark & \cmark & \xmark & \cmark & \cmark & \cmark & \cmark & \xmark & \cmark & \xmark & \xmark & \xmark & \partialmark & \xmark\\
    \midrule
\textit{\textbf{\ours{} (Ours)}} & \cmark & \cmark & \cmark & \cmark & \cmark & \cmark & \cmark & \cmark & \cmark & \cmark & \cmark & \cmark & \cmark & \cmark & \cmark & \cmark \\
\bottomrule
\end{tabular}%
}
\vspace{-0.05in}
\caption{Comparison with representative role-play frameworks. Character columns cover explicit profiles, automatic profile updates, and whether character states use open rather than fixed schema. Interaction columns cover multi-character scenes and automatic scene initialization or switching. Message columns indicate whether generated messages include speech, thought, action, environmental messages, and collective messages by multiple characters. World columns cover support for diverse worlds, global world settings, global world state evolution, open-schema world representations, and location/entity-level state modeling. Symbols indicate full support (\cmark), partial support (\partialmark), or no support (\xmark). Detailed clarifications for partial support are provided in Appendix~\ref{app:partial_support_clarifications}.}
\label{tab:framework_comparison}
\vspace{-0.15in}
\end{table*}

Large language models (LLMs) have enabled fluent role-playing agents that imitate fictional characters and sustain persona-grounded dialogue~\citep{CharacterLLM, RoleLLM, CoSER, CharacterGLM, xu2026improvinggeneralroleplayingagents, AdaMARP}. Yet simulating a literary world poses a harder long-horizon challenge: as a story unfolds, characters revise beliefs, motivations, and relationships, while locations, objects, and background conditions also change. The goal is therefore not only to produce the next plausible utterance, but to maintain coherent character and world states across scenes.

Existing role-playing systems fall short in three ways. First, most persona-based agents rely only on static profiles or short dialogue contexts~\citep{ChatHaruhi, DITTO, SimsChat, CharacterBench, liu-etal-2026-costbench}. Second, multi-agent environments often use manually specified sandboxes, making them hard to scale to diverse literary worlds~\citep{GenerativeAgents, LARP, GenLARP}. Third, prior book-grounded systems focus on single-scene role-play~\citep{CoSER, AdaMARP} or partial long-horizon updates. For example, BookWorld~\citep{BookWorld} updates characters' goals and states, and global events, but lacks full profile evolution, location/entity-level world updates, and trainable subtask supervision.

We argue that literary world simulation requires \textit{open-schema co-evolution}. Open-schema means that the system infers the relevant character and world dimensions from each book rather than forcing all stories into fixed slots: character dimensions can differ between a detective's investigative habits and a Victorian orphan's social position, while world dimensions may shift from school rules and class hierarchies to political orders or supernatural systems. Once these dimensions are constructed, co-evolution keeps character and world states coupled: character actions can reshape locations or social orders, while world changes can in turn alter motivations and profiles. The key difficulty is deciding which dimensions to track, when evidence justifies profile updates, and how local events propagate to global, location, and entity states. Thus, we reframe interactive literary world from characters and worlds that remain largely static to ones that can fully develop beyond their initial descriptions as long as supported by ongoing interactions.

We present \textbf{\ours{}}, a framework for \textbf{open-schema} character and world \textbf{evolution} in interactive multi-agent literary worlds. Given a book snapshot, \ours{} simulates the story forward with two coupled modules: an open-schema \textbf{Character Agent} for multi-character role-play and profile evolution, and an LLM-based \textbf{World Model} for global, location, and entity-level state tracking and scene progression (Figure~\ref{fig:casestudy}). Within the Character Agent, profile dimensions may evolve at different speeds: emotions can shift quickly, while personality traits often require accumulated evidence. We therefore use a \textbf{hidden tracker} to store weak or emerging evidence separately before they justify profile updates. We decompose the framework into 7 supervised tasks covering scene initialization, interaction generation, and state update.

We construct a dataset from 57 books, yielding \textbf{138,596} supervised training samples and \textbf{222} test snapshots. We also introduce a trajectory-level LLM-as-Judge evaluation framework covering 20 metrics, measuring persistent character and world development beyond local role-playing quality. Experiments show that \ours{} reduces the long-horizon performance degradation often observed in previous frameworks (Figure~\ref{fig:length_eval_frameworks}).

\begin{figure*}[t]
\vspace{-0.2in}
\centering
\includegraphics[width=\textwidth]{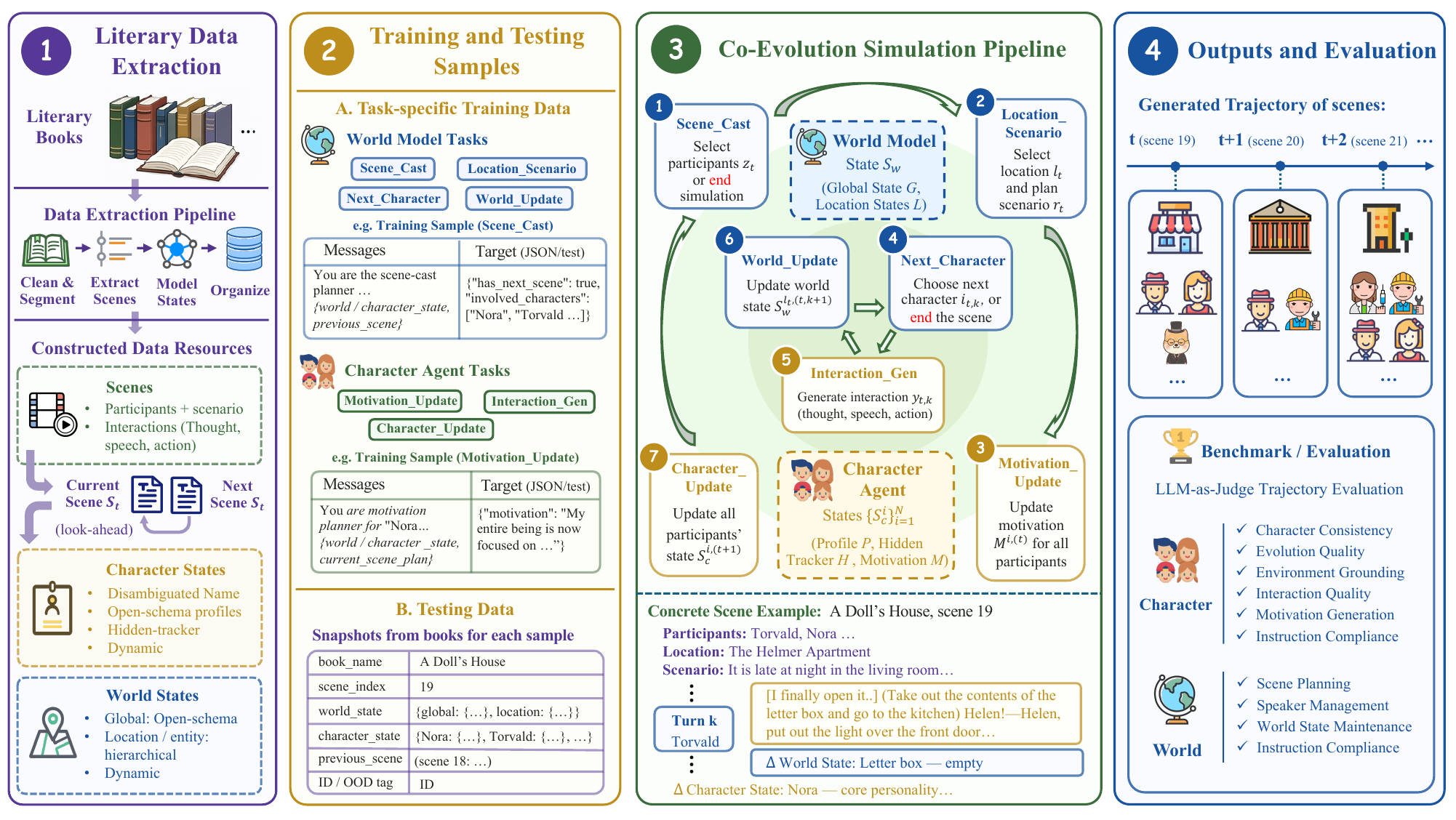}
\vspace{-0.25in}
\caption{Overview of our \ours{} framework. It contains  following stages: Dataset Construction, parts 1-2 (Section~\ref{sec:dataset_construction}); Simulation Pipeline, part 3 (Section~\ref{sec:task_formalization}); and Evaluation Method, part 4 (Section~\ref{sec:evaluation_framework}). }
\vspace{-0.15in}
\label{fig:evolvingbook_framework1}
\end{figure*}

Overall, we make 4 contributions: \textbf{(1)} a new formulation of interactive literary world simulation as a long-horizon process in which characters and the world can fully develop beyond their initial descriptions based on interactions rather than updating only a few predefined fields, shifting the objective from reproducing the original book toward sustaining a grounded, continuously developing world; \textbf{(2)} \ours{}, a simulation framework that couples an open-schema Character Agent with hidden trackers for multi-timescale profile evolution and an LLM-based World Model that maintains open-schema global and location/entity-level states while guiding scene progression, decomposed into 7 trainable tasks; \textbf{(3)} a benchmark built from 57 books with \textbf{138,596} training samples and \textbf{222} test snapshots, together with trajectory-level evaluation framework covering 20 metrics; and \textbf{(4)} empirical evidence that \ours{} reduces long-horizon performance degradation, demonstrating the value of co-evolving character and world states.

\section{Related Work}

\textbf{Role-Play Agents.} Role-playing language agents have progressed from scripts~\citep{RoleLLM} and self-alignment~\citep{DITTO, SimsChat} to training-based personality simulation~\citep{CharacterLLM, CharacterGLM, CharacterBot}, multi-modality~\citep{MMRole} and memory retrieval~\citep{TailorRPA, BeyondDialogue}. While these systems typically treat personas as static anchors, we enables open-schema character evolution over long-horizon narratives.

\textbf{Multi-Agent and World Model.} Believable simulation needs grounded interaction. While social~\citep{GenerativeAgents, AgentSociety, TrendSim} and gaming~\citep{LARP, CharacterBox} environments rely on fixed sandboxes, which do not scale to diverse worlds, role-play agents like CoSER~\citep{CoSER} and  AdaMARP~\citep{AdaMARP} lack world modeling. BookWorld~\citep{BookWorld} contains a world agent but uses a predefined fixed schema and only static world states. Recent advances explore LLM-based world models~\citep{AWMsurvey, WordtoWorld} and also open-schema in event extraction~\citep{AutoSchemaKG, KARMA}, which haven't been studied in interactive literary worlds.
We bridge this gap by supporting LLM-based open-schema world evolutions for diverse book worlds.

\textbf{Role-Play Evaluation.} Evaluation has shifted from fluency toward behavior~\citep{CharacterEval, CharacterBench, RPEval}. While some recent works have introduced trajectory-level evaluation~\citep{AdaMARP, CPO}, the quality of state evolution remains underexplored. \ours{} introduces a trajectory-level framework to further quantify persistent character and world development in open-ended environments.

\section{The \ours{} Framework}

Figure~\ref{fig:evolvingbook_framework1} provides an overview of \ours{}, with data construction, simulation pipeline, and evaluation method. It unfolds a book from a chosen narrative point into an evolving interactive literary world. Given a structured state snapshot at that point, it simulates scene-by-scene interactions over persistent character and world states modeling.


\subsection{Design Principles of \ours{}}


Table~\ref{tab:framework_comparison} compares \ours{} with existing role-play frameworks. Most prior systems face two main limitations: their characters rely on fixed, static profiles that cannot evolve over time; and their worlds are either rigid sandboxes or lack detailed entity-level state tracking. To address these gaps, \ours{} introduces two core components, a \textbf{Character Agent} and a \textbf{World Model}, to drive open-schema, long-horizon evolution.

\subsubsection{Character Agent}
The Character Agent represents each character with an open-schema profile, since characters from different books vary a lot: a detective may have investigative habits, while a Victorian orphan is better characterized by social position. Prior literary role-play frameworks~\citep{CoSER, AdaMARP, BookWorld} rely on fixed profile dimensions. In contrast, we provide only reference dimensions when constructing profiles, allowing LLM to select, merge, or introduce new fields according to the book's genre, setting, and style.

Beyond individual characters, the Character Agent also supports the environment and character groups as special acting units. They enable environmental events and shared group interactions within the same interaction loop.

The Character Agent further supports persistent profile evolution. Unlike prior works~\citep{BookWorld, GenerativeAgents} updating only several predefined dimensions such as memory or psychological state, we treat every dimension in the open-schema profile as a potentially evolvable part of the character state. Moreover, profile dimensions may evolve at different speeds: mood may change quickly, but personality requires accumulated evidence. To model this multi-timescale evolution, we introduce a hidden tracker that records weak or emerging evidence separately from the profile. This design prevents premature profile updates while allowing repeated signals across scenes to accumulate into later changes. During evolution, the Character Agent considers both dimension-level changeability and accumulated hidden evidence before updating states.

\subsubsection{World Model}
The World Model maintains both a global world state and location-level physical states. The global state captures world-level settings, such as historical background and social institutions. Since literary works construct vastly different worlds, ranging from school narratives to post-apocalyptic societies, we adopt an open-schema design that avoids reducing the global state to fixed dimensions, unlike prior systems like BookWorld~\citep{BookWorld}.

In addition to global state, we explicitly model the physical state of each location. Sandbox-based environments such as Generative Agents~\citep{GenerativeAgents} and LARP~\citep{LARP} depend on a manually predefined single world, which limits scalability across diverse worlds. We instead use an LLM-based World Model to construct detailed physical states for all locations and update them throughout simulation. Locations can be nested with sub-locations, such as a house with rooms, or separate atomic locations, such as the road outside the house. For each level of location, the framework maintains a detailed description and tracks all important non-character entities and their states, such as a Christmas tree standing by the window. Both global and location-level states are automatically updated through character interactions.

Together, these evolving character and world states enable long-horizon, cross-scene simulation in interactive multi-agent literary worlds, where characters are not constrained to static personas and worlds are not treated as passive backdrops.

\subsection{Simulation Pipeline and Task Formulation}
\label{sec:task_formalization}

We now formulate how Character Agent and World Model are composed into a co-evolution simulation pipeline. Starting from a snapshot of a book, the simulator repeatedly plans a scene, generates multi-character interactions, updates the world during the scene, and revises character states after the scene. This turns the design principles above into a sequence of tasks, supporting end-to-end long-horizon simulation. To keep it clear, we define the states and modules' observations, and present the seven tasks in their execution order below.

\begingroup
\setlength{\abovedisplayskip}{4pt}
\setlength{\belowdisplayskip}{5pt}
\setlength{\abovedisplayshortskip}{3pt}
\setlength{\belowdisplayshortskip}{4pt}

\paragraph{States and observations.}
At scene step $t$, let $\mathcal{I}$ denote the full character set and $\mathcal{L}$ the set of locations. The simulator maintains a global world state, one location state for each location $\ell \in \mathcal{L}$, and one character state for each character $i \in \mathcal{I}$:
\[
\begin{aligned}
S_w^{\ell,(t)} &= \bigl(G^{(t)}, L^{\ell,(t)}\bigr), \quad \ell \in \mathcal{L},\\
S_c^{i,(t)} &= \bigl(P^{i,(t)}, H^{i,(t)}, M^{i,(t)}\bigr), \quad i \in \mathcal{I}.
\end{aligned}
\]
Here $S_w^{\ell,(t)}$ denotes the world state at location $\ell$, while $S_c^{i,(t)}$ denotes the state of character $i$. In particular, $G^{(t)}$ is the open-schema global world state, and $L^{\ell,(t)}$ stores the description of location $\ell$ and its important entity states. For character $i$, $P^{i,(t)}$ is the open-schema profile, $H^{i,(t)}$ is the hidden tracker, and $M^{i,(t)}$ is the scene-level motivation.

The observations are module-specific views constructed from these states. For any character subset $A \subseteq \mathcal{I}$ and location subset $B \subseteq \mathcal{L}$, the World Model observation $O_w^{A,B,(t)}$ contains the global state, the location states in $B$, and the character states in $A$. For any character $i$ and location $\ell$, the Character Agent observation $O_c^{i,\ell,(t)}$ contains the corresponding location-specific world state and character state:
\[
\begin{aligned}
O_w^{A,B,(t)} &= \bigl(G^{(t)}, \{L^{\ell,(t)}\}_{\ell \in B}, \{S_c^{i,(t)}\}_{i \in A}\bigr),\\
O_c^{i,\ell,(t)} &= \bigl(S_w^{\ell,(t)}, S_c^{i,(t)}\bigr).
\end{aligned}
\]
\noindent\textbf{Task sequence.}
Each scene is generated by the following ordered tasks.

\noindent\textbf{Task 1: \texttt{scene\_cast}.} The World Model selects the participating character set $z_t \subseteq \mathcal{I}$ from the full character set:
\[
O_w^{\mathcal{I},\varnothing,(t)} \rightarrow z_t.
\]
\noindent\textbf{Task 2: \texttt{location\_scenario}.} Given the selected cast, the World Model produces a scene plan that specifies the location $\ell_t$ and scenario $r_t$:
\[
(O_w^{z_t,\mathcal{L},(t)}, z_t) \rightarrow \ell_t, r_t.
\]
\noindent\textbf{Task 3: \texttt{motivation\_update}.} Before the scene begins, the Character Agent prepares each participant with a scene-specific motivation:
\[
(O_c^{i,\ell_t,(t)}, r_t) \rightarrow M^{i,(t)}, \quad i \in z_t.
\]
\noindent\textbf{Task 4: \texttt{next\_character}.} During the scene, let $Y_{t,<k}$ be the interaction history before turn $k$. The World Model chooses the next acting character from the selected cast:
\[
(O_w^{z_t,\{\ell_t\},(t)}, r_t, Y_{t,<k}) \rightarrow i_{t,k}.
\]
\noindent\textbf{Task 5: \texttt{interaction\_gen}.} Given the selected actor $i_{t,k}$, which may be a character, the environment, or a character group, the Character Agent generates $y_{t,k}$. Its content may mix thought \texttt{[...]}, speech as plain text, and action \texttt{(...)}. Only the character's own thoughts in $Y_{t,<k}$ are visible:
\[
(O_c^{i_{t,k},\ell_t,(t)}, r_t, Y_{t,<k}) \rightarrow y_{t,k}.
\]
\noindent\textbf{Task 6: \texttt{world\_update}.} After each interaction, the World Model updates the global and location state:
\[
(O_w^{z_t,\{\ell_t\},(t)}, y_{t,k}, Y_{t,<k}) \rightarrow S_w^{\ell_t,(t,k+1)}.
\]
The completed scene history is denoted by $Y_t = (y_{t,1}, \ldots, y_{t,K_t})$, where $K_t$ is the number of generated turns before the scene ends.

\noindent\textbf{Task 7: \texttt{character\_update}.} After the scene, the final in-scene world state becomes $S_w^{\ell_t,(t+1)}$, and each participating character state is updated:
\[
(O_c^{i,\ell_t,(t)}, Y_t, S_w^{\ell_t,(t+1)}) \rightarrow S_c^{i,(t+1)}, \quad i \in z_t.
\]
Non-participating characters and non-current locations keep their previous states.

Together, these tasks define one scene-level transition over the persistent states and interactions,
\[
\begin{aligned}
\bigl(S_w^{\mathcal{L},(t)}, S_c^{\mathcal{I},(t)}\bigr)
&\rightarrow
\bigl(S_w^{\mathcal{L},(t+1)}, S_c^{\mathcal{I},(t+1)}, Y_t\bigr) \\
&= \bigl(O_w^{\mathcal{I},\mathcal{L},(t+1)}, Y_t\bigr),
\end{aligned}
\]
and repeated transitions produce the full trajectory,
\[
\begin{aligned}
\tau = & \bigl(O_w^{\mathcal{I},\mathcal{L},(0)}, z_1, r_1, Y_1, O_w^{\mathcal{I},\mathcal{L},(1)}, \dots, \\
& z_T, r_T, Y_T, O_w^{\mathcal{I},\mathcal{L},(T)}\bigr).
\end{aligned}
\]
The goal is to produce a grounded long-horizon trajectory in which scene content, character evolution, and world-state changes remain mutually consistent. See Appendix~\ref{sec:app_simulation_protocol} for pseudocode.
\endgroup

\subsection{Dataset Construction}
\label{sec:dataset_construction}

Following \citet{CoSER}, we select 57 chronologically narrated books. We use \textit{Gemini-2.5-Pro} as extraction LLM. Chronological narratives allow later scenes to serve as look-ahead references, so character and world-state changes are grounded in textual evidence rather than LLM's knowledge. Detailed construction and prompts are provided in Appendices~\ref{app:data_construction_details} and~\ref{app:data_prompts}.

\begin{table*}[t]
\vspace{-0.2in}
\centering
\huge
\resizebox{\linewidth}{!}{
\begin{tabular}{lcccccccccccc}
\toprule
\multicolumn{13}{l}{\textbf{CC}: Character Consistency \quad \textbf{EQ}: Evolution Quality \quad \textbf{EG}: Environmental Grounding \quad \textbf{IQ}: Interaction Quality \quad \textbf{MG}: Motivation Generation} \\
\multicolumn{13}{l}{\textbf{IC}: Instruction Compliance \quad \textbf{PF}: Profile Fidelity \quad \textbf{SSF}: Speaking Style Fidelity \quad \textbf{MDB}: Motivation-Driven Behavior \quad \textbf{PUF}: Profile Update Fidelity} \\
\multicolumn{13}{l}{\textbf{PES}: Profile Evolution Smoothness \quad \textbf{EA}: Environment Awareness \quad \textbf{EU}: Environmental Utilization} \\
\multicolumn{13}{l}{\textbf{CR}: Contextual Responsiveness \quad \textbf{NP}: Narrative Progression \quad \textbf{MQ}: Motivation Quality} \\
\midrule
\multirow{2}{*}{\textbf{Models}} & \multicolumn{3}{c}{\textbf{CC}} & \multicolumn{2}{c}{\textbf{EQ}} & \multicolumn{2}{c}{\textbf{EG}} & \multicolumn{2}{c}{\textbf{IQ}} & \multicolumn{1}{c}{\textbf{MG}} & \multicolumn{1}{c}{\textbf{IC}} & \multirow{2}{*}{\textbf{Avg.}} \\
\cmidrule(lr){2-4} \cmidrule(lr){5-6} \cmidrule(lr){7-8} \cmidrule(lr){9-10} \cmidrule(lr){11-11} \cmidrule(lr){12-12}
 & \textbf{PF} & \textbf{SSF} & \textbf{MDB} & \textbf{PUF} & \textbf{PES} & \textbf{EA} & \textbf{EU} & \textbf{CR} & \textbf{NP} & \textbf{MQ} & \textbf{IC} & \\
\midrule
\rowcolor{gray!15} \textbf{Closed-source} &  &  &  &  &  &  &  &  &  &  &  &  \\
Kimi-K2.5  &  80.79{\Large$\pm$10.84}  &  61.45{\Large$\pm$19.00}  &  87.52{\Large\phantom{0}$\pm$9.78}  &  81.07{\Large\phantom{0}$\pm$8.63}  &  54.51{\Large$\pm$16.38}  &  81.29{\Large$\pm$10.59}  &  83.86{\Large$\pm$13.00}  &  88.31{\Large\phantom{0}$\pm$8.27}  &  56.02{\Large$\pm$16.66}  &  77.12{\Large$\pm$12.25}  &  82.83{\Large\phantom{0}$\pm$5.68}  &  75.89 \\
Gemini-2.5-Flash  &  75.34{\Large\phantom{0}$\pm$3.94}  &  61.68{\Large\phantom{0}$\pm$6.80}  &  75.67{\Large\phantom{0}$\pm$5.96}  &  68.61{\Large\phantom{0}$\pm$5.91}  &  58.02{\Large\phantom{0}$\pm$7.68}  &  67.02{\Large\phantom{0}$\pm$6.14}  &  57.10{\Large\phantom{0}$\pm$8.90}  &  75.72{\Large\phantom{0}$\pm$4.02}  &  51.86{\Large\phantom{0}$\pm$9.17}  &  74.61{\Large\phantom{0}$\pm$8.04}  &  51.45{\Large$\pm$26.93}  &  65.19\\
Gemini-2.5-Pro  &  93.88{\Large\phantom{0}$\pm$3.47}  &  85.94{\Large\phantom{0}$\pm$5.90}  &  94.95{\Large\phantom{0}$\pm$3.44}  &  83.28{\Large\phantom{0}$\pm$5.33}  &  70.51{\Large\phantom{0}$\pm$8.65}  &  89.53{\Large\phantom{0}$\pm$5.19}  &  90.12{\Large\phantom{0}$\pm$6.15}  &  93.63{\Large\phantom{0}$\pm$3.25}  &  72.10{\Large\phantom{0}$\pm$8.19}  &  77.20{\Large\phantom{0}$\pm$6.36}  &  \underline{86.06}{\Large\phantom{0}$\pm$3.08}  &  85.20\\
Gemini-3.1-Pro-P  &  89.79{\Large\phantom{0}$\pm$5.23}  &  79.40{\Large\phantom{0}$\pm$6.91}  &  93.09{\Large\phantom{0}$\pm$3.57}  &  78.90{\Large\phantom{0}$\pm$5.87}  &  66.29{\Large\phantom{0}$\pm$9.98}  &  85.35{\Large\phantom{0}$\pm$6.17}  &  83.90{\Large\phantom{0}$\pm$8.33}  &  91.87{\Large\phantom{0}$\pm$3.62}  &  67.74{\Large\phantom{0}$\pm$9.63}  &  83.04{\Large\phantom{0}$\pm$6.41}  &  84.12{\Large$\pm$12.90}  &  82.14\\
GPT-4o  &  71.45{\Large\phantom{0}$\pm$6.84}  &  54.13{\Large\phantom{0}$\pm$9.15}  &  76.36{\Large\phantom{0}$\pm$6.76}  &  57.66{\Large\phantom{0}$\pm$8.60}  &  54.46{\Large\phantom{0}$\pm$8.62}  &  75.80{\Large\phantom{0}$\pm$7.04}  &  76.32{\Large\phantom{0}$\pm$9.16}  &  77.94{\Large\phantom{0}$\pm$4.63}  &  55.89{\Large$\pm$11.73}  &  69.94{\Large\phantom{0}$\pm$8.81}  &  74.15{\Large$\pm$16.39}  &  67.65 \\
GPT-5-Chat  &  80.73{\Large\phantom{0}$\pm$5.60}  &  73.44{\Large\phantom{0}$\pm$9.14}  &  85.89{\Large\phantom{0}$\pm$5.45}  &  71.33{\Large\phantom{0}$\pm$7.30}  &  64.00{\Large\phantom{0}$\pm$9.68}  &  86.74{\Large\phantom{0}$\pm$5.29}  &  92.52{\Large\phantom{0}$\pm$6.49}  &  87.36{\Large\phantom{0}$\pm$4.09}  &  67.37{\Large\phantom{0}$\pm$9.19}  &  81.49{\Large\phantom{0}$\pm$7.85}  &  77.29{\Large$\pm$20.63}  &  78.92 \\
GPT-5.3-Chat  &  94.71{\Large\phantom{0}$\pm$2.75}  &  88.77{\Large\phantom{0}$\pm$4.30}  &  93.71{\Large\phantom{0}$\pm$2.95}  &  77.90{\Large\phantom{0}$\pm$3.74}  &  \underline{80.16}{\Large\phantom{0}$\pm$6.62}  &  91.62{\Large\phantom{0}$\pm$4.88}  &  92.10{\Large\phantom{0}$\pm$7.54}  &  92.27{\Large\phantom{0}$\pm$3.74}  &  59.98{\Large\phantom{0}$\pm$7.55}  &  83.94{\Large\phantom{0}$\pm$6.50}  &  83.85{\Large\phantom{0}$\pm$3.25}  &  85.36\\
Claude-4.6-Sonnet  &  \underline{96.47}{\Large\phantom{0}$\pm$4.16}  &  \underline{95.52}{\Large\phantom{0}$\pm$4.51}  &  \underline{98.49}{\Large\phantom{0}$\pm$3.29}  &  \underline{95.30}{\Large\phantom{0}$\pm$5.61}  &  71.02{\Large$\pm$15.19}  &  \underline{94.33}{\Large\phantom{0}$\pm$5.25}  &  \underline{96.79}{\Large\phantom{0}$\pm$7.27}  &  \underline{97.29}{\Large\phantom{0}$\pm$4.67}  &  \underline{84.03}{\Large\phantom{0}$\pm$9.31}  &  \underline{90.17}{\Large\phantom{0}$\pm$5.07}  &  62.07{\Large$\pm$35.76}  &  \underline{89.23}\\
Claude-4.6-Opus  &  \textbf{97.81}{\Large\phantom{0}$\pm$1.92}  &  \textbf{96.70}{\Large\phantom{0}$\pm$2.79}  &  \textbf{99.00}{\Large\phantom{0}$\pm$1.63}  &  \textbf{96.36}{\Large\phantom{0}$\pm$2.83}  &  \textbf{84.62}{\Large\phantom{0}$\pm$9.15}  &  \textbf{95.91}{\Large\phantom{0}$\pm$3.36}  &  \textbf{98.73}{\Large\phantom{0}$\pm$2.46}  &  \textbf{98.85}{\Large\phantom{0}$\pm$1.49}  &  \textbf{89.76}{\Large\phantom{0}$\pm$6.06}  &  \textbf{95.36}{\Large\phantom{0}$\pm$3.63}  &  \textbf{91.53}{\Large\phantom{0}$\pm$3.25}  &  \textbf{94.97} \\
\midrule
\rowcolor{gray!15} \textbf{Open-source} &  &  &  &  &  &  &  &  &  &  &  &  \\
\rowcolor{gray!15} \multicolumn{13}{l}{\hspace{1em}\textbf{$<$ 14B}} \\
Qwen3-4B-I & 15.31{\Large$\pm$10.34} & \phantom{0}7.25{\Large\phantom{0}$\pm$6.89} & 13.57{\Large$\pm$10.80} & 21.84{\Large$\pm$10.54} & 26.89{\Large$\pm$11.14} & 30.11{\Large$\pm$11.88} & 32.68{\Large$\pm$10.21} & 15.69{\Large$\pm$12.49} & 12.47{\Large\phantom{0}$\pm$9.73} & 39.00{\Large$\pm$18.49} & 24.76{\Large$\pm$17.68} & 21.77\\
Qwen2.5-7B-I  &  15.77{\Large\phantom{0}$\pm$6.68}  &  \phantom{0}7.69{\Large\phantom{0}$\pm$5.06}  &  11.70{\Large\phantom{0}$\pm$6.98}  &  32.25{\Large\phantom{0}$\pm$7.42}  &  25.07{\Large$\pm$10.49}  &  30.45{\Large\phantom{0}$\pm$6.41}  &  28.53{\Large\phantom{0}$\pm$6.75}  &  11.80{\Large\phantom{0}$\pm$7.31}  &  \phantom{0}9.69{\Large\phantom{0}$\pm$5.53}  &  36.20{\Large$\pm$16.06}  &  20.14{\Large$\pm$12.72}  &  20.84 \\
Qwen-7B~(Coser)  &  16.59{\Large$\pm$11.97}  &  12.20{\Large\phantom{0}$\pm$8.33}  &  27.61{\Large$\pm$12.64}  &  18.16{\Large$\pm$16.58}  &  17.91{\Large$\pm$16.25}  &  37.34{\Large\phantom{0}$\pm$8.69}  &  25.20{\Large\phantom{0}$\pm$4.14}  &  13.92{\Large\phantom{0}$\pm$6.96} &  27.60{\Large$\pm$12.86} &  2.06{\Large\phantom{0}$\pm$9.94}  &  10.26{\Large\phantom{0}$\pm$1.55}  &  18.98
\\
Qwen-7B~(Crab)  &  14.25{\Large\phantom{0}$\pm$7.83}  &  \phantom{0}6.34{\Large\phantom{0}$\pm$4.43}  &  15.76{\Large\phantom{0}$\pm$7.80}  &  19.82{\Large\phantom{0}$\pm$7.22}  &  15.89{\Large\phantom{0}$\pm$9.94}  &  32.79{\Large$\pm$10.41}  &  20.28{\Large\phantom{0}$\pm$5.85}  &  16.91{\Large\phantom{0}$\pm$8.45}  &  17.74{\Large\phantom{0}$\pm$5.31}  &   26.82{\Large$\pm$12.05}  &  14.91{\Large\phantom{0}$\pm$8.10}  &  18.32
\\
Llama-3.1-8B-I  &  30.58{\Large$\pm$11.10}  &  22.30{\Large\phantom{0}$\pm$8.36}  &  30.71{\Large$\pm$12.65}  &  21.87{\Large$\pm$16.04}  &  20.36{\Large$\pm$13.89}  &  44.33{\Large\phantom{0}$\pm$8.88}  &  \textbf{46.19}{\Large\phantom{0}$\pm$9.30}  &  40.78{\Large$\pm$12.36}  &  \textbf{40.77}{\Large\phantom{0}$\pm$9.04}  &  27.26{\Large$\pm$22.46}  &  14.55{\Large$\pm$17.16}  &  30.88\\
Llama-8B~(Coser)  &  30.36{\Large$\pm$16.76}  &  22.26{\Large$\pm$15.53}  &  38.28{\Large$\pm$18.63}  &  15.86{\Large$\pm$16.21}  &  15.94{\Large$\pm$15.91}  &  37.61{\Large\phantom{0}$\pm$8.57}  &  31.04{\Large\phantom{0}$\pm$2.85}  &  30.36{\Large$\pm$14.55}  &  37.27{\Large$\pm$11.12}  &   \phantom{0}4.07{\Large$\pm$14.52}  &  11.09{\Large\phantom{0}$\pm$4.60}  &  24.92\\
Llama-8B~(Crab)  &  29.37{\Large$\pm$18.52}  &  20.00{\Large$\pm$16.45}  &  38.10{\Large$\pm$20.05}  &  \phantom{0}1.26{\Large\phantom{0}$\pm$5.04}  &  \phantom{0}2.27{\Large\phantom{0}$\pm$8.62}  &  33.20{\Large\phantom{0}$\pm$5.08}  &  32.67{\Large\phantom{0}$\pm$2.56}  &  26.57{\Large$\pm$12.04}  &  34.03{\Large\phantom{0}$\pm$7.19}  &  10.45{\Large$\pm$20.33}  &  10.25{\Large\phantom{0}$\pm$1.88}  &  21.65\\

\textbf{Qwen-4B~(EW-ours)} & 44.68{\Large$\pm$10.21} & 37.24{\Large\phantom{0}$\pm$9.73} & 37.10{\Large\phantom{0}$\pm$8.45} & 35.24{\Large\phantom{0}$\pm$7.81} & 31.43{\Large$\pm$10.20} & 40.09{\Large\phantom{0}$\pm$6.84} & 28.15{\Large\phantom{0}$\pm$5.41} & 35.64{\Large\phantom{0}$\pm$9.28} & 28.30{\Large\phantom{0}$\pm$5.88} & 40.57{\Large\phantom{0}$\pm$9.72} & 59.12{\Large$\pm$11.15} & 37.96\\
\textbf{Qwen-7B~(EW-ours)}  &  \textbf{54.07}{\Large\phantom{0}$\pm$9.85}  &  \textbf{48.19}{\Large$\pm$10.44}  &  \textbf{47.15}{\Large\phantom{0}$\pm$8.44}  &  \underline{41.24}{\Large\phantom{0}$\pm$7.72}  &  \underline{40.33}{\Large$\pm$10.06}  &  \underline{45.44}{\Large\phantom{0}$\pm$7.27}  &  31.81{\Large\phantom{0}$\pm$6.14}  &  \textbf{48.61}{\Large\phantom{0}$\pm$9.92}  &  36.09{\Large\phantom{0}$\pm$6.68}  &  \underline{42.12}{\Large\phantom{0}$\pm$8.31}  &  \textbf{65.78}{\Large\phantom{0}$\pm$9.45}  &  \underline{45.53} \\
\textbf{Llama-8B~(EW-ours)}  &  \underline{50.94}{\Large$\pm$20.32}  &  \underline{44.90}{\Large$\pm$18.45}  &  \underline{46.43}{\Large$\pm$16.41}  &  \textbf{42.49}{\Large$\pm$10.11}  &  \textbf{42.44}{\Large$\pm$10.66}  &  \textbf{46.47}{\Large$\pm$11.92}  &  \underline{33.75}{\Large\phantom{0}$\pm$7.68}  &  \underline{46.03}{\Large$\pm$19.69}  &  \underline{38.83}{\Large$\pm$13.13}  &  \textbf{47.96}{\Large\phantom{0}$\pm$7.72}  &  \underline{65.62}{\Large\phantom{0}$\pm$8.79}  &  \textbf{45.99} \\
\addlinespace[0.3ex]
\midrule
\addlinespace[0.3ex]
\rowcolor{gray!15} \multicolumn{13}{l}{\hspace{1em}\textbf{$\geq$ 14B}} \\
Qwen2.5-14B-I  &  31.44{\Large$\pm$10.64}  &  14.71{\Large\phantom{0}$\pm$7.63}  &  30.43{\Large$\pm$10.11}  &  43.95{\Large\phantom{0}$\pm$6.51}  &  35.56{\Large$\pm$11.66}  &  43.84{\Large\phantom{0}$\pm$8.65}  &  38.79{\Large\phantom{0}$\pm$9.18}  &  32.83{\Large\phantom{0}$\pm$9.13}  &  15.66{\Large\phantom{0}$\pm$6.22}  &  53.36{\Large$\pm$11.16}  &  39.78{\Large$\pm$18.50}  &  34.58 \\
Qwen2.5-32B-I  &  19.49{\Large$\pm$11.17}  &  \phantom{0}7.42{\Large\phantom{0}$\pm$6.16}  &  20.08{\Large\phantom{0}$\pm$9.81}  &  35.85{\Large\phantom{0}$\pm$9.25}  &  29.82{\Large$\pm$11.19}  &  39.50{\Large\phantom{0}$\pm$8.48}  &  32.63{\Large\phantom{0}$\pm$8.40}  &  24.55{\Large\phantom{0}$\pm$8.20}  &  13.87{\Large\phantom{0}$\pm$6.75}  &  45.12{\Large$\pm$12.65}  &  38.14{\Large$\pm$15.37}  &  27.86 \\
\textbf{Qwen-14B~(EW-ours)}  &  62.39{\Large$\pm$14.45}  &  55.74{\Large$\pm$13.59}  &  55.63{\Large$\pm$13.02}  &  49.82{\Large\phantom{0}$\pm$8.48}  &  \underline{53.60}{\Large$\pm$11.17}  &  50.03{\Large\phantom{0}$\pm$9.79}  &  36.25{\Large\phantom{0}$\pm$7.47}  &  55.36{\Large$\pm$14.26}  &  44.04{\Large\phantom{0}$\pm$9.66}  &  49.85{\Large\phantom{0}$\pm$7.98}  &  \textbf{66.27}{\Large$\pm$14.59}  &  52.63 \\
\textbf{Qwen-32B~(EW-ours)}  &  \textbf{67.81}{\Large$\pm$12.50}  &  \textbf{60.56}{\Large$\pm$12.20}  &  \underline{61.34}{\Large$\pm$10.76}  &  \textbf{57.24}{\Large\phantom{0}$\pm$7.83}  &  \textbf{58.62}{\Large$\pm$10.76}  &  55.43{\Large\phantom{0}$\pm$9.28}  &  39.22{\Large\phantom{0}$\pm$7.91}  &  60.55{\Large$\pm$13.42}  &  \underline{48.48}{\Large\phantom{0}$\pm$9.52}  &  54.93{\Large\phantom{0}$\pm$7.91}  &  63.45{\Large$\pm$22.36}  &  \underline{57.06} \\
Qwen2.5-72B-I  &  25.71{\Large$\pm$10.42}  &  \phantom{0}9.55{\Large\phantom{0}$\pm$5.89}  &  30.88{\Large$\pm$10.99}  &  37.77{\Large\phantom{0}$\pm$8.08}  &  38.90{\Large\phantom{0}$\pm$6.64}  &  46.31{\Large\phantom{0}$\pm$7.83}  &  36.60{\Large\phantom{0}$\pm$7.27}  &  36.50{\Large$\pm$14.50}  &  23.44{\Large$\pm$10.29}  &  43.86{\Large\phantom{0}$\pm$9.17}  &  41.56{\Large$\pm$32.31}  &  33.73 \\
Qwen3-32B  &  42.16{\Large$\pm$13.59}  &  26.76{\Large$\pm$12.91}  &  58.39{\Large$\pm$10.81}  &  51.10{\Large\phantom{0}$\pm$9.07}  &  35.24{\Large$\pm$12.76}  &  62.37{\Large$\pm$10.82}  &  69.64{\Large$\pm$12.52}  &  \underline{65.99}{\Large\phantom{0}$\pm$9.24}  &  47.34{\Large$\pm$11.69}  &  \underline{70.18}{\Large\phantom{0}$\pm$9.74}  &  54.14{\Large$\pm$19.77}  &  53.03 \\
Llama-3.3-70B-I  &  22.66{\Large\phantom{0}$\pm$8.93}  &  \phantom{0}6.23{\Large\phantom{0}$\pm$4.10}  &  27.03{\Large$\pm$11.61}  &  36.36{\Large$\pm$10.95}  &  40.42{\Large\phantom{0}$\pm$5.32}  &  46.90{\Large\phantom{0}$\pm$7.07}  &  39.27{\Large\phantom{0}$\pm$8.92}  &  39.83{\Large$\pm$11.62}  &  22.21{\Large$\pm$12.14}  &  52.85{\Large$\pm$11.61}  &  46.33{\Large$\pm$28.49}  &  34.55 \\
Mistral-Small  &  54.19{\Large\phantom{0}$\pm$9.61}  &  39.24{\Large\phantom{0}$\pm$9.77}  &  57.23{\Large\phantom{0}$\pm$7.62}  &  46.02{\Large\phantom{0}$\pm$5.48}  &  44.53{\Large\phantom{0}$\pm$8.14}  &  \underline{68.46}{\Large\phantom{0}$\pm$6.27}  &  \underline{70.89}{\Large\phantom{0}$\pm$7.01}  &  61.67{\Large\phantom{0}$\pm$6.76}  &  40.48{\Large\phantom{0}$\pm$6.78}  &  52.40{\Large\phantom{0}$\pm$9.10}  &  \underline{63.85}{\Large\phantom{0}$\pm$5.01}  &  54.45
 \\
DeepSeek-V3-0324  &  \underline{66.22}{\Large$\pm$13.05}  &  \underline{56.23}{\Large$\pm$12.75}  &  \textbf{75.59}{\Large\phantom{0}$\pm$8.32}  &  \underline{55.73}{\Large\phantom{0}$\pm$7.73}  &  53.28{\Large$\pm$11.00}  &  \textbf{71.96}{\Large\phantom{0}$\pm$9.04}  &  \textbf{76.43}{\Large$\pm$10.32}  &  \textbf{72.48}{\Large\phantom{0}$\pm$8.38}  &  \textbf{51.89}{\Large$\pm$14.65}  &  \textbf{70.20}{\Large\phantom{0}$\pm$8.05}  &  55.04{\Large$\pm$16.32}  &  \textbf{64.10} \\
\bottomrule
\end{tabular}
}
\caption{Main benchmark results for \textbf{Character Agent} evaluation. I and P denote Instruct and Preview in model names. The best performances within the same model scale are \textbf{bold-faced}, and the second-best are \underline{underlined}.}
\label{tab:character_results_claude}
\vspace{-0.15in}
\end{table*}

\begin{table*}[t]
\vspace{-0.2in}
\centering
\huge
\resizebox{0.85\linewidth}{!}{
\begin{tabular}{lcccccccccc}
\toprule
\multicolumn{11}{l}{\textbf{SP}: Scene Planning \quad \textbf{SM}: Speaker Management \quad \textbf{WSM}: World State Maintenance \quad \textbf{IC}: Instruction Compliance} \\
\multicolumn{11}{l}{\textbf{CSR}: Cast Selection Rationality \quad \textbf{LSR}: Location \& Scenario Rationality \quad \textbf{SCC}: Scene Continuity \& Coherence } \\
\multicolumn{11}{l}{ \textbf{TSO}: Turn \& Scene Orchestration \quad \textbf{GUS}: Global Update Sensitivity \quad \textbf{GSA}: Global State Accuracy } \\
\multicolumn{11}{l}{\textbf{LUS}: Location Update Sensitivity \quad \textbf{LSA}: Location State Accuracy} \\
\midrule
\multirow{2}{*}{\textbf{Models}} & \multicolumn{3}{c}{\textbf{SP}} & \multicolumn{1}{c}{\textbf{SM}} & \multicolumn{4}{c}{\textbf{WSM}} & \multicolumn{1}{c}{\textbf{IC}} & \multirow{2}{*}{\textbf{Avg.}} \\
\cmidrule(lr){2-4} \cmidrule(lr){5-5} \cmidrule(lr){6-9} \cmidrule(lr){10-10}
 & \textbf{CSR} & \textbf{LSR} & \textbf{SCC} & \textbf{TSO} & \textbf{GUS} & \textbf{GSA} & \textbf{LUS} & \textbf{LSA} & \textbf{IC} & \\
\midrule
\rowcolor{gray!15} \textbf{Closed-source} &  &  &  &  &  &  &  &  &  &  \\
Kimi-K2.5  &  78.36{\Large\phantom{0}$\pm$7.78}  &  80.61{\Large$\pm$15.01}  &  51.72{\Large$\pm$25.95}  &  58.84{\Large$\pm$10.95}  &  \underline{66.64}{\Large\phantom{0}$\pm$7.53}  &  \textbf{59.85}{\Large\phantom{0}$\pm$8.22}  &  \underline{68.90}{\Large\phantom{0}$\pm$6.87}  &  \underline{70.54}{\Large\phantom{0}$\pm$9.77}  &  77.94{\Large\phantom{0}$\pm$9.94}  &  68.16\\
Gemini-2.5-Flash  &  73.18{\Large\phantom{0}$\pm$6.95}  &  84.09{\Large\phantom{0}$\pm$6.10}  &  66.36{\Large$\pm$14.98}  &  43.02{\Large$\pm$24.56}  &  60.41{\Large\phantom{0}$\pm$7.49}  &  56.78{\Large\phantom{0}$\pm$4.10}  &  54.22{\Large$\pm$11.27}  &  56.25{\Large\phantom{0}$\pm$6.31}  &  43.49{\Large$\pm$23.83}  &  59.76\\
Gemini-2.5-Pro  &  78.86{\Large\phantom{0}$\pm$4.73}  &  88.33{\Large\phantom{0}$\pm$6.27}  &  \underline{70.34}{\Large$\pm$20.14}  &  75.66{\Large\phantom{0}$\pm$5.50}  &  65.89{\Large\phantom{0}$\pm$4.69}  &  56.99{\Large\phantom{0}$\pm$5.84}  &  54.96{\Large\phantom{0}$\pm$8.66}  &  68.20{\Large\phantom{0}$\pm$9.76}  &  80.36{\Large\phantom{0}$\pm$7.15}  &  71.07\\
Gemini-3.1-Pro-P  &  \underline{83.50}{\Large\phantom{0}$\pm$3.17}  &  91.60{\Large\phantom{0}$\pm$4.81}  &  69.16{\Large$\pm$20.97}  &  72.70{\Large$\pm$10.18}  &  65.60{\Large\phantom{0}$\pm$9.89}  &  55.13{\Large$\pm$10.68}  &  66.51{\Large\phantom{0}$\pm$9.67}  &  61.54{\Large$\pm$10.84}  &  85.55{\Large$\pm$13.49}  &  72.37\\
GPT-4o  &  77.80{\Large\phantom{0}$\pm$8.81}  &  85.33{\Large\phantom{0}$\pm$6.55}  &  51.02{\Large$\pm$16.90}  &  61.44{\Large\phantom{0}$\pm$7.34}  &  61.40{\Large\phantom{0}$\pm$8.32}  &  52.86{\Large\phantom{0}$\pm$5.91}  &  54.12{\Large\phantom{0}$\pm$9.96}  &  51.44{\Large\phantom{0}$\pm$7.07}  &  73.50{\Large$\pm$18.44}  &  63.21\\
GPT-5-Chat  &  80.44{\Large\phantom{0}$\pm$6.88}  &  88.30{\Large\phantom{0}$\pm$6.12}  &  68.38{\Large$\pm$16.61}  &  64.69{\Large$\pm$17.27}  &  60.57{\Large\phantom{0}$\pm$9.14}  &  53.97{\Large\phantom{0}$\pm$6.21}  &  52.59{\Large\phantom{0}$\pm$7.83}  &  57.85{\Large\phantom{0}$\pm$7.87}  &  77.51{\Large$\pm$22.07}  &  67.14\\
GPT-5.3-Chat  &  80.04{\Large\phantom{0}$\pm$5.09}  &  89.35{\Large\phantom{0}$\pm$4.81}  &  68.40{\Large$\pm$15.48}  &  69.04{\Large\phantom{0}$\pm$6.41}  &  \textbf{70.02}{\Large\phantom{0}$\pm$2.12}  &  51.11{\Large\phantom{0}$\pm$5.48}  &  \textbf{75.21}{\Large\phantom{0}$\pm$6.30}  &  64.77{\Large$\pm$10.96}  &  \underline{85.57}{\Large\phantom{0}$\pm$4.92}  &  \underline{72.61}\\
Claude-4.6-Sonnet  &  81.89{\Large\phantom{0}$\pm$5.46}  &  \underline{92.90}{\Large\phantom{0}$\pm$4.58}  &  60.77{\Large$\pm$25.33}  &  \underline{79.84}{\Large$\pm$13.54}  &  38.93{\Large$\pm$24.04}  &  33.26{\Large$\pm$20.37}  &  30.33{\Large$\pm$21.21}  &  41.12{\Large$\pm$26.94}  &  58.83{\Large$\pm$34.31}  &  57.54 \\
Claude-4.6-Opus  &  \textbf{84.86}{\Large\phantom{0}$\pm$5.34}  &  \textbf{96.66}{\Large\phantom{0}$\pm$2.56}  &  \textbf{82.28}{\Large$\pm$13.37}  &  \textbf{84.98}{\Large\phantom{0}$\pm$4.76}  &  66.05{\Large\phantom{0}$\pm$5.74}  &  \underline{58.48}{\Large\phantom{0}$\pm$6.75}  &  61.72{\Large\phantom{0}$\pm$9.46}  &  \textbf{75.49}{\Large\phantom{0}$\pm$8.93}  &  \textbf{89.34}{\Large\phantom{0}$\pm$6.50}  &  \textbf{77.76}\\
\midrule
\rowcolor{gray!15} \textbf{Open-source} &  &  &  &  &  &  &  &  &  &  \\
\rowcolor{gray!15} \multicolumn{11}{l}{\hspace{1em}\textbf{$<$ 14B}} \\
Qwen3-4B-I & 60.20{\Large$\pm$15.88} & 52.50{\Large$\pm$23.75} & 20.71{\Large$\pm$15.06} & \phantom{0}5.65{\Large\phantom{0}$\pm$6.64} & 41.98{\Large$\pm$23.54} & 35.56{\Large$\pm$20.73} & 22.68{\Large$\pm$14.07} & 23.31{\Large$\pm$13.57} & 25.78{\Large$\pm$18.28} & 32.06 \\
Qwen2.5-7B-I  &  61.07{\Large$\pm$12.78}  &  51.25{\Large$\pm$14.71}  &  14.09{\Large$\pm$14.59}  &  \phantom{0}5.82{\Large\phantom{0}$\pm$4.92}  &  53.53{\Large$\pm$11.52}  &  \underline{44.53}{\Large\phantom{0}$\pm$9.69}  &  28.65{\Large\phantom{0}$\pm$9.43}  &  31.53{\Large\phantom{0}$\pm$8.02}  &  35.58{\Large$\pm$16.83}  &  36.23\\
Llama-3.1-8B-I  &  60.56{\Large$\pm$17.34}  &  \textbf{68.01}{\Large$\pm$17.31}  &  \underline{34.16}{\Large$\pm$13.44}  &  28.65{\Large$\pm$11.65}   &  49.02{\Large$\pm$19.99}  &  43.94{\Large$\pm$17.78}  &  45.89{\Large$\pm$19.33}  &  41.12{\Large$\pm$16.92}  &  30.82{\Large$\pm$36.53}  &  44.69\\

\textbf{Qwen-4B~(EW-ours)} & 61.78{\Large$\pm$10.20} & 62.42{\Large$\pm$11.14} & 23.20{\Large$\pm$16.20} & 30.81{\Large\phantom{0}$\pm$6.65} & 57.62{\Large\phantom{0}$\pm$7.97} & 44.15{\Large\phantom{0}$\pm$8.01} & 58.09{\Large\phantom{0}$\pm$7.59} & 46.55{\Large\phantom{0}$\pm$6.40} & 72.24{\Large$\pm$13.19} & 50.76\\
\textbf{Qwen-7B~(EW-ours)}  &  \textbf{63.90}{\Large\phantom{0}$\pm$8.89}  &  63.32{\Large$\pm$11.66}  &  32.02{\Large$\pm$17.41}  &  \textbf{39.98}{\Large\phantom{0}$\pm$6.79}  &  \textbf{61.21}{\Large\phantom{0}$\pm$5.41}  &  \textbf{47.52}{\Large\phantom{0}$\pm$6.02}  &  \textbf{61.20}{\Large\phantom{0}$\pm$4.10}  &  \textbf{49.10}{\Large\phantom{0}$\pm$4.44}  &  \textbf{76.59}{\Large\phantom{0}$\pm$9.78}  &  \textbf{54.98} \\
\textbf{Llama-8B~(EW-ours)}  &  \underline{63.58}{\Large\phantom{0}$\pm$7.91}  &  \underline{67.57}{\Large$\pm$11.51}  &  \textbf{38.41}{\Large$\pm$23.29}  &  \underline{39.79}{\Large$\pm$11.31}  &  \underline{56.59}{\Large\phantom{0}$\pm$7.62}  &  42.89{\Large\phantom{0}$\pm$7.78}  &  \underline{60.19}{\Large\phantom{0}$\pm$5.80}  &  \underline{46.75}{\Large\phantom{0}$\pm$4.99}  &  \underline{72.90}{\Large\phantom{0}$\pm$7.92}  &  \underline{54.30}\\
\addlinespace[0.3ex]
\midrule
\addlinespace[0.3ex]
\rowcolor{gray!15} \multicolumn{11}{l}{\hspace{1em}\textbf{$\geq$ 14B}} \\
Qwen2.5-14B-I  &  70.66{\Large$\pm$10.08}  &  67.37{\Large$\pm$13.98}  &  21.47{\Large$\pm$15.64}  &  16.95{\Large\phantom{0}$\pm$6.74}  &  50.64{\Large$\pm$21.59}  &  41.28{\Large$\pm$17.15}  &  23.71{\Large$\pm$12.53}  &  29.82{\Large$\pm$13.41}  &  41.89{\Large$\pm$18.78}  &  40.42 \\
Qwen2.5-32B-I  &  69.55{\Large\phantom{0}$\pm$9.80}  &  60.18{\Large$\pm$14.46}  &  17.69{\Large$\pm$14.67}  &  18.03{\Large\phantom{0}$\pm$7.14}  &  61.21{\Large\phantom{0}$\pm$6.80}  &  \underline{49.53}{\Large\phantom{0}$\pm$7.00}  &  42.35{\Large$\pm$13.03}  &  42.08{\Large\phantom{0}$\pm$7.58}  &  51.12{\Large$\pm$17.99}  &  45.75\\
\textbf{Qwen-14B~(EW-ours)}  &  68.18{\Large\phantom{0}$\pm$7.96}  &  70.67{\Large$\pm$10.66}  &  39.24{\Large$\pm$18.05}  &  \underline{45.19}{\Large$\pm$10.48}  &  59.45{\Large\phantom{0}$\pm$7.84}  &  46.76{\Large\phantom{0}$\pm$6.86}  &  \textbf{60.84}{\Large\phantom{0}$\pm$6.83}  &  49.08{\Large\phantom{0}$\pm$6.37}  &  \textbf{77.02}{\Large$\pm$15.15}  &  57.38\\
\textbf{Qwen-32B~(EW-ours)}  &  69.77{\Large\phantom{0}$\pm$7.99}  &  77.26{\Large\phantom{0}$\pm$9.58}  &  \underline{45.42}{\Large$\pm$17.69}  &  \textbf{48.21}{\Large\phantom{0}$\pm$9.83}  &  \underline{61.67}{\Large\phantom{0}$\pm$3.38}  &  \textbf{49.92}{\Large\phantom{0}$\pm$5.06}  &  \underline{60.65}{\Large\phantom{0}$\pm$4.40}  &  \textbf{51.95}{\Large\phantom{0}$\pm$3.96}  &  \underline{73.94}{\Large$\pm$24.58}  &  \textbf{59.87}\\
Qwen2.5-72B-I  &  70.29{\Large\phantom{0}$\pm$9.27}  &  69.52{\Large\phantom{0}$\pm$8.44}  &  37.89{\Large$\pm$13.99}  &  29.88{\Large$\pm$11.59}  &  39.85{\Large$\pm$26.68}  &  31.12{\Large$\pm$20.86}  &  27.04{\Large$\pm$19.54}  &  29.36{\Large$\pm$20.08}  &  37.35{\Large$\pm$29.99}  &  41.37\\
Qwen3-32B  &  67.59{\Large\phantom{0}$\pm$9.82}  &  71.76{\Large$\pm$12.88}  &  34.70{\Large$\pm$20.55}  &  43.60{\Large\phantom{0}$\pm$9.59}  &  54.41{\Large\phantom{0}$\pm$8.87}  &  47.33{\Large\phantom{0}$\pm$7.26}  &  53.52{\Large\phantom{0}$\pm$9.16}  &  49.02{\Large\phantom{0}$\pm$9.48}  &  51.34{\Large$\pm$20.40}  &  52.59\\
Llama-3.3-70B-I  &  \underline{75.62}{\Large$\pm$10.65}  &  \underline{78.39}{\Large$\pm$12.61}  &  36.76{\Large$\pm$14.19}  &  23.47{\Large\phantom{0}$\pm$9.18}  &  41.47{\Large$\pm$22.18}  &  34.41{\Large$\pm$18.74}  &  22.41{\Large$\pm$15.41}  &  27.44{\Large$\pm$15.72}  &  34.90{\Large$\pm$22.87}  &  41.65\\
Mistral-Small  &  72.24{\Large\phantom{0}$\pm$6.72}  &  73.74{\Large\phantom{0}$\pm$6.86}  &  41.38{\Large$\pm$16.00}  &  42.39{\Large\phantom{0}$\pm$6.98}  &  60.88{\Large\phantom{0}$\pm$7.47}  &  44.55{\Large\phantom{0}$\pm$6.82}  &  41.64{\Large\phantom{0}$\pm$8.90}  &  44.92{\Large\phantom{0}$\pm$6.53}  &  60.41{\Large\phantom{0}$\pm$8.50}  &  53.57\\
DeepSeek-V3-0324  &  \textbf{76.98}{\Large\phantom{0}$\pm$6.36}  &  \textbf{81.86}{\Large\phantom{0}$\pm$8.43}  &  \textbf{51.00}{\Large$\pm$19.30}  &  38.75{\Large$\pm$15.05}  &  \textbf{63.48}{\Large\phantom{0}$\pm$5.11}  &  48.79{\Large\phantom{0}$\pm$6.70}  &  47.32{\Large$\pm$10.08}  &  \underline{49.70}{\Large\phantom{0}$\pm$8.14}  &  60.33{\Large$\pm$16.47}  &  \underline{57.58}
\\
\bottomrule
\end{tabular}
}
\caption{Main benchmark results for \textbf{World Model} evaluation. I and P denote Instruct and Preview in model names. The best performances within the same model scale are \textbf{bold-faced}, and the second-best are \underline{underlined}.}
\label{tab:world_results_claude}
\vspace{-0.15in}
\end{table*}

\subsubsection{Structured Data Extraction}
\label{sec:data_pipeline}

\ours{} segments each book into text chunks and uses the extraction LLM to build structured scenes with summaries, scenarios, key characters, and multi-turn interactions. If a scene is cut off at a chunk boundary, the truncated text is prepended to the next chunk, so the LLM can continue the same scene.

For characters, the LLM first unifies different names that refer to the same character from extracted mentions. It then builds each character's initial state from the first few relevant scenes, including an open-schema profile and hidden tracker, and updates this state scene by scene using later narrative evidence as look-ahead. For example, if an idle student reflects after a failure, a later scene showing sustained effort confirms that the reflection led to a real character-state change.

World states follow the same principle. The global world state tracks story-level settings and systemic conditions, while location-level states track each location's description and important non-character entities. The immediately following interactions and scene provide reference evidence for global-state changes, while interactions and the next scene happening at the same location provide evidence for location's physical-state changes. The LLM standardizes location names, initializes both world states, and updates them after interactions.

\subsubsection{Train/Test Split Construction}

We then split them into training data, in-domain (ID) test data from partially seen books, and out-of-domain (OOD) test data from books excluded from training. Concatenating the extracted states along each timeline yields \textbf{138,596} supervised training samples for the seven tasks in \S\ref{sec:task_formalization}; test samples are selected from specific time points, each containing the current character and world states from which simulation can continue. The test split contains \textbf{222} samples. Appendix~\ref{app:dataset_stats} details dataset statistics.

\subsection{Evaluation Framework}
\label{sec:evaluation_framework}

\ours{} evaluates systems with two top-level score families, CHARACTER and WORLD, covering 10 dimensions and 20 metrics (on a 0--100 scale). These dimensions target persistent simulation quality beyond local persona imitation. 

CHARACTER. The 6 dimensions are: (1) Character Consistency: Profile Fidelity, Speaking Style Fidelity, Motivation-Driven Behavior; (2) Evolution Quality: Profile Update Fidelity, Profile Evolution Smoothness; (3) Environmental Grounding: Environment Awareness, Environmental Utilization; (4) Interaction Quality: Contextual Responsiveness, Narrative Progression; (5) Motivation Generation: Motivation Quality; (6) Instruction Compliance: Instruction Compliance.

WORLD. The 4 dimensions are: (1) Scene Planning: Cast Selection Rationality, Location \& Scenario Rationality, Scene Continuity \& Coherence; (2) Speaker Management: Turn \& Scene Orchestration; (3) World State Maintenance: Global Update Sensitivity, Global State Accuracy, Location Update Sensitivity, Location State Accuracy; (4) Instruction Compliance: Instruction Compliance.

For each trajectory $\tau$, we use a \textbf{per-metric independent LLM-as-Judge} design, where each judge receives only relevant inputs and the rubric for one metric. Scores are aggregated hierarchically to capture both scene-level quality and long-range character/world evolution. If a simulation terminates early due to invalid output, we apply task-specific penalties so that the failures are reflected in both Instruction Compliance and the affected metrics. Appendix~\ref{sec:eval_framework} gives the full scoring process.

\section{Experiment}

\subsection{Experiment Setup}

\noindent\textbf{Training.}
We fine-tune open-source backbones with supervised instruction tuning. Following prior work~\citep{CoSER, AdaMARP}, we mix in Tulu3~\citep{tulu} general instruction tuning data at a 1:1 ratio to preserve general capabilities. Our fine-tuned models span both model family and size, including \textit{Llama-3.1-8B-Instruct}, \textit{Qwen2.5-7/14/32B-Instruct}, and \textit{Qwen3-4B-Instruct}. For brevity, we refer to each model by its size in the table. Details are in Appendix~\ref{app:training_inference_details}.

\noindent\textbf{Models.}
We evaluate 10 closed-source APIs, 11 open-source models, prior role-playing baselines including CoSER~\citep{CoSER} and Crab~\citep{Crab}, and models trained on \ours{} (EW). For untrained settings, the same LLM serves as Character Agent and World Model. For the trained ones (with their training data specified in parentheses), since CoSER and Crab only provide role-play data, we evaluate them as Character Agents paired with the same untrained backbone as World Model. EW models use separately trained Character Agent and World Model with the same backbone. Full results are in Appendix~\ref{app:full_results_ew_benchmark}.

\noindent\textbf{Simulation.}
To evaluate long-horizon behavior, we allow up to 20 scenes per sample and 50 interactions per scene. 

Appendix~\ref{app:ablation_study} reports ablation results, \ref{app:id_ood_results} reports in- and out-of-distribution results, \ref{app:judge_model_comparison} examines different judge-models, \ref{app:video_generation} demonstrates downstream video generation, and \ref{app:human_eval} reports human evaluation.

\begin{table*}[t]
\vspace{-0.2in}
\centering
\huge
\resizebox{\linewidth}{!}{
\begin{tabular}{llccccccccccc}
\toprule
\multicolumn{13}{l}{\textbf{CC}: Character Consistency \quad \textbf{EQ}: Evolution Quality \quad \textbf{EG}: Environmental Grounding \quad \textbf{IQ}: Interaction Quality \quad \textbf{MG}: Motivation Generation} \\
\multicolumn{13}{l}{\textbf{PF}: Profile Fidelity \quad \textbf{SSF}: Speaking Style Fidelity \quad \textbf{MDB}: Motivation-Driven Behavior \quad \textbf{PUF}: Profile Update Fidelity \quad \textbf{PES}: Profile Evolution Smoothness } \\
\multicolumn{13}{l}{\textbf{EA}: Environment Awareness \quad \textbf{EU}: Environmental Utilization  \quad \textbf{CR}: Contextual Responsiveness \quad \textbf{NP}: Narrative Progression \quad \textbf{MQ}: Motivation Quality} \\
\midrule
\multirow{2}{*}{\textbf{Models}} & \multirow{2}{*}{\textbf{Framework}} & \multicolumn{3}{c}{\textbf{CC}} & \multicolumn{2}{c}{\textbf{EQ}} & \multicolumn{2}{c}{\textbf{EG}} & \multicolumn{2}{c}{\textbf{IQ}} & \multicolumn{1}{c}{\textbf{MG}} & \multirow{2}{*}{\textbf{Avg.}} \\
\cmidrule(lr){3-5} \cmidrule(lr){6-7} \cmidrule(lr){8-9} \cmidrule(lr){10-11} \cmidrule(lr){12-12} 
 & & \textbf{PF} & \textbf{SSF} & \textbf{MDB} & \textbf{PUF} & \textbf{PES} & \textbf{EA} & \textbf{EU} & \textbf{CR} & \textbf{NP} & \textbf{MQ} & \\
\midrule
\multirow{2}{*}{Gemini-3.1-Pro-P} & BookWorld & 84.83{\Large\phantom{0}$\pm$7.30} & 77.07{\Large\phantom{0}$\pm$8.34} & 90.43{\Large\phantom{0}$\pm$4.06} & 36.97{\Large\phantom{0}$\pm$7.94} & 30.61{\Large\phantom{0}$\pm$7.26} & 83.40{\Large\phantom{0}$\pm$5.16} & 85.71{\Large\phantom{0}$\pm$6.42} & 90.45{\Large\phantom{0}$\pm$3.38} & 65.22{\Large$\pm$10.89} & 63.60{\Large$\pm$11.14} & 70.83 \\
 & \ours{} & 89.79{\Large\phantom{0}$\pm$5.23} & 79.40{\Large\phantom{0}$\pm$6.91} & 93.09{\Large\phantom{0}$\pm$3.57} & 78.90{\Large\phantom{0}$\pm$5.87} & 66.29{\Large\phantom{0}$\pm$9.98} & 85.35{\Large\phantom{0}$\pm$6.17} & 83.90{\Large\phantom{0}$\pm$8.33} & 91.87{\Large\phantom{0}$\pm$3.62} & 67.74{\Large\phantom{0}$\pm$9.63} & 83.04{\Large\phantom{0}$\pm$6.41} & 81.94\\
\midrule
\multirow{2}{*}{GPT-5.3-Chat} & BookWorld & 88.77{\Large\phantom{0}$\pm$4.40} & 83.79{\Large\phantom{0}$\pm$4.88} & 87.03{\Large\phantom{0}$\pm$4.48} & 40.58{\Large\phantom{0}$\pm$8.70} & 31.28{\Large$\pm$11.96} & 88.89{\Large\phantom{0}$\pm$5.64} & 89.15{\Large\phantom{0}$\pm$6.68} & 86.97{\Large\phantom{0}$\pm$4.49} & 47.68{\Large\phantom{0}$\pm$7.69} & 62.22{\Large$\pm$11.75}  & 70.64 \\
 & \ours{} & 94.71{\Large\phantom{0}$\pm$2.75} & 88.77{\Large\phantom{0}$\pm$4.30} & 93.71{\Large\phantom{0}$\pm$2.95} & 77.90{\Large\phantom{0}$\pm$3.74} & 80.16{\Large\phantom{0}$\pm$6.62} & 91.62{\Large\phantom{0}$\pm$4.88} & 92.10{\Large\phantom{0}$\pm$7.54} & 92.27{\Large\phantom{0}$\pm$3.74} & 59.98{\Large\phantom{0}$\pm$7.55} & 83.94{\Large\phantom{0}$\pm$6.50} & 85.52 \\
\midrule
\multirow{2}{*}{Llama-3.1-8B-I} & BookWorld & \phantom{0}7.19{\Large\phantom{0}$\pm$6.67} & \phantom{0}4.55{\Large\phantom{0}$\pm$4.14} & \phantom{0}9.30{\Large\phantom{0}$\pm$5.86} & \phantom{0}7.86{\Large\phantom{0}$\pm$4.57} & 10.08{\Large\phantom{0}$\pm$6.66} & 23.09{\Large\phantom{0}$\pm$8.94} & 34.27{\Large\phantom{0}$\pm$8.46} & 13.92{\Large\phantom{0}$\pm$6.87} & 10.58{\Large\phantom{0}$\pm$7.21} & 12.99{\Large$\pm$12.22} & 13.38 \\
 & \ours{} & 30.58{\Large$\pm$11.10} & 22.30{\Large\phantom{0}$\pm$8.36} & 30.71{\Large$\pm$12.65} & 21.87{\Large$\pm$16.04} & 20.36{\Large$\pm$13.89} & 44.33{\Large\phantom{0}$\pm$8.88} & 46.19{\Large\phantom{0}$\pm$9.30} & 40.78{\Large$\pm$12.36} & 40.77{\Large\phantom{0}$\pm$9.04} & 27.26{\Large$\pm$22.46} & 32.52 \\
\midrule
\multirow{2}{*}{Qwen2.5-14B-I} & BookWorld & 27.87{\Large$\pm$10.89} & 12.83{\Large\phantom{0}$\pm$8.18} & 27.96{\Large\phantom{0}$\pm$9.87} & 26.03{\Large\phantom{0}$\pm$5.30} & 17.68{\Large\phantom{0}$\pm$8.27} & 41.88{\Large\phantom{0}$\pm$9.84} & 33.21{\Large\phantom{0}$\pm$9.59} & 28.39{\Large\phantom{0}$\pm$7.90} & 14.38{\Large\phantom{0}$\pm$4.51} & 37.21{\Large\phantom{0}$\pm$8.28} & 26.74 \\
 & \ours{} & 31.44{\Large$\pm$10.64} & 14.71{\Large\phantom{0}$\pm$7.63} & 30.43{\Large$\pm$10.11} & 43.95{\Large\phantom{0}$\pm$6.51} & 35.56{\Large$\pm$11.66} & 43.84{\Large\phantom{0}$\pm$8.65} & 38.79{\Large\phantom{0}$\pm$9.18} & 32.83{\Large\phantom{0}$\pm$9.13} & 15.66{\Large\phantom{0}$\pm$6.22} & 53.36{\Large$\pm$11.16} & 34.06 \\
\bottomrule
\end{tabular}
}
\vspace{-0.05in}
\caption{Comparison of \textbf{Character Agent} performance between \ours{} (ours) and BookWorld.}
\label{tab:character_comparison_bw}
\vspace{-0.05in}
\end{table*}

\begin{table*}[t]
\centering
\huge
\resizebox{0.75\linewidth}{!}{
\begin{tabular}{llccccccc}
\toprule
\multicolumn{9}{l}{\textbf{SP}: Scene Planning \quad \textbf{SM}: Speaker Management \quad \textbf{WSM}: World State Maintenance \quad \textbf{CSR}: Cast Selection Rationality} \\
\multicolumn{9}{l}{\textbf{LSR}: Location \& Scenario Rationality  \quad  \textbf{SCC}: Scene Continuity \& Coherence \quad \textbf{TSO}: Turn \& Scene Orchestration } \\
\multicolumn{9}{l}{\textbf{GUS}: Global Update Sensitivity \quad \textbf{GSA}: Global State Accuracy } \\
\midrule
\multirow{2}{*}{\textbf{Models}} & \multirow{2}{*}{\textbf{Framework}} & \multicolumn{3}{c}{\textbf{SP}} & \multicolumn{1}{c}{\textbf{SM}} & \multicolumn{2}{c}{\textbf{WSM}} & \multirow{2}{*}{\textbf{Avg.}} \\
\cmidrule(lr){3-5} \cmidrule(lr){6-6} \cmidrule(lr){7-8} 
 & & \textbf{CSR} & \textbf{LSR} & \textbf{SCC} & \textbf{TSO} & \textbf{GUS} & \textbf{GSA} & \\
\midrule
\multirow{2}{*}{Gemini-3.1-Pro-P} & BookWorld & 80.64{\Large\phantom{0}$\pm$5.71} & 88.51{\Large\phantom{0}$\pm$7.30} & 61.67{\Large$\pm$18.20} & 70.83{\Large\phantom{0}$\pm$6.73} & 62.31{\Large\phantom{0}$\pm$4.65} & 49.78{\Large\phantom{0}$\pm$8.84} & 68.96\\
 & \ours{} & 83.50{\Large\phantom{0}$\pm$3.17} & 91.60{\Large\phantom{0}$\pm$4.81} & 69.16{\Large$\pm$20.97} & 72.70{\Large$\pm$10.18} & 65.60{\Large\phantom{0}$\pm$9.89} & 55.13{\Large$\pm$10.68} & 72.95 \\
\midrule
\multirow{2}{*}{GPT-5.3-Chat} & BookWorld & 80.38{\Large\phantom{0}$\pm$4.58} & 87.63{\Large\phantom{0}$\pm$5.81} & 52.17{\Large$\pm$20.07} & 56.57{\Large\phantom{0}$\pm$7.11} & 67.24{\Large\phantom{0}$\pm$3.94} & 50.69{\Large\phantom{0}$\pm$8.98} & 65.78 \\
 & \ours{} & 80.04{\Large\phantom{0}$\pm$5.09} & 89.35{\Large\phantom{0}$\pm$4.81} & 68.40{\Large$\pm$15.48} & 69.04{\Large\phantom{0}$\pm$6.41} & 70.02{\Large\phantom{0}$\pm$2.12} & 51.11{\Large\phantom{0}$\pm$5.48} & 71.33 \\
\midrule
\multirow{2}{*}{Llama-3.1-8B-I} & BookWorld & 42.21{\Large$\pm$11.95} & 44.16{\Large$\pm$13.19} & \phantom{0}4.49{\Large\phantom{0}$\pm$8.17} & \phantom{0}5.09{\Large\phantom{0}$\pm$2.85} & 45.72{\Large$\pm$10.74} & 30.46{\Large\phantom{0}$\pm$8.08} & 28.69 \\
 & \ours{} & 60.56{\Large$\pm$17.34} & 68.01{\Large$\pm$17.31} & 34.16{\Large$\pm$13.44} & 28.65{\Large$\pm$11.65} & 49.02{\Large$\pm$19.99} & 43.94{\Large$\pm$17.78} & 47.39 \\
\midrule
\multirow{2}{*}{Qwen2.5-14B-I} & BookWorld & 70.57{\Large\phantom{0}$\pm$6.91} & 66.60{\Large\phantom{0}$\pm$8.45} & 14.97{\Large$\pm$11.99} & 14.45{\Large\phantom{0}$\pm$4.65} & 50.49{\Large\phantom{0}$\pm$5.87} & 40.87{\Large\phantom{0}$\pm$4.67} & 42.99 \\
 & \ours{} & 70.66{\Large$\pm$10.08} & 67.37{\Large$\pm$13.98} & 21.47{\Large$\pm$15.64} & 16.95{\Large\phantom{0}$\pm$6.74} & 50.64{\Large$\pm$21.59} & 41.28{\Large$\pm$17.15}  & 44.73 \\
\bottomrule
\end{tabular}
}
\vspace{-0.05in}
\caption{Comparison of \textbf{World Model} performance between \ours{} (ours) and BookWorld.}
\label{tab:world_comparison_bw}
\vspace{-0.15in}
\end{table*}

\subsection{Main Results}
Tables~\ref{tab:character_results_claude} and~\ref{tab:world_results_claude} report the mean and standard deviation scores. \textit{Claude-4.6-Sonnet} is our main judge.

\noindent\textbf{(1) EW gains come from modeling state evolution beyond role-play imitation.}
 Across matched backbones, EW-training improves both Character Agent and World Model performance, showing the effect of book-to-world supervision. The advantage is especially clear over role-play-only baselines: on both Qwen-7B and Llama-8B, EW-trained Character Agents substantially outperform CoSER and Crab, indicating that long-horizon role-playing requires modeling coupled character and world-state evolution rather than only imitating dialogue.

\noindent\textbf{(2) EW improves controlled character evolution and more precise environmental grounding.}
For Character Agents, the main gains concentrate on character consistency, evolution, and interactive progression. Models better preserve stable traits such as profile, speaking style, and motivations, while making profile updates and evolution smoother under event-conditioned changes. Although Environment Utilization sometimes decreases, Environment Awareness consistently improves, suggesting more accurate grounding rather than arbitrary use of environmental details.

\noindent\textbf{(3) EW improves world modeling by targeting structured state tracking which is underrepresented in pretraining.}
For World Models, \textit{Claude-4.6-Opus} achieves the best overall score, but models generally perform worse on world-model tasks than on character-agent tasks, reflecting the difficulty of explicit structured state tracking, a capability less covered by pretraining than narrative continuation. EW training improves this most clearly in long-range scene continuity, turn/scene orchestration, and location-level updates, showing that models learn to track how events reshape both global and location-level states. With this targeted supervision, Qwen-32B trained on EW reaches a World Model Avg. of 59.87, surpassing even \textit{Claude-4.6-Sonnet} and \textit{Gemini-2.5-Flash}.

\begin{figure}[t]
    \centering
    \includegraphics[width=0.9\columnwidth]{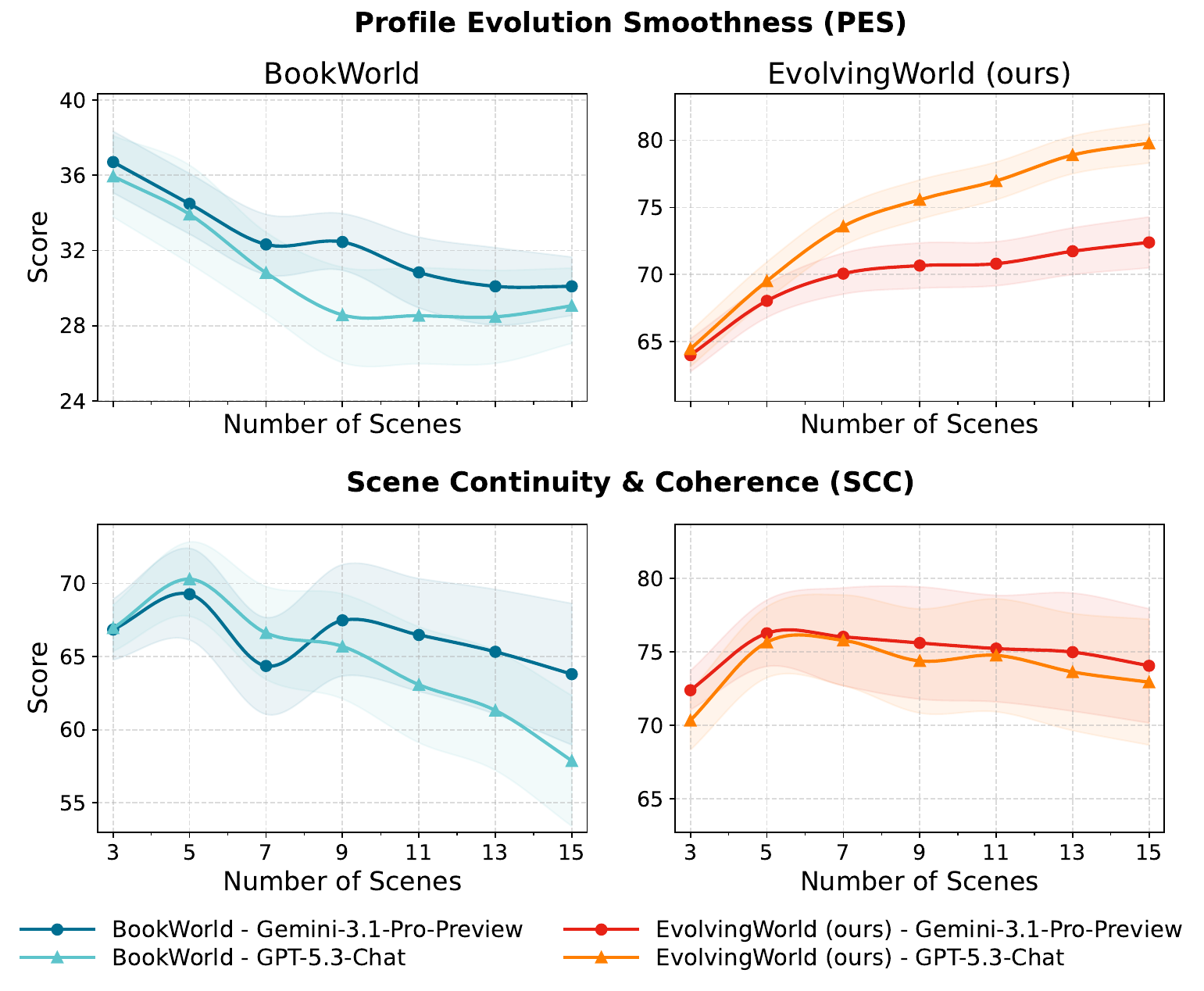}
    \vspace{-0.05in}
    \caption{Length-wise comparison between \textbf{\ours{}} and BookWorld on Profile Evolution Smoothness (PES) and Scene Continuity \& Coherence (SCC). Shading denotes mean $\pm$ std/4.}
    \label{fig:length_eval_frameworks}
\vspace{-0.15in}
\end{figure}

\subsection{Comparison with BookWorld}

We further compare \ours{} with BookWorld~\citep{BookWorld}, a representative long-horizon multi-agent role-play simulation baseline. BookWorld updates only a few predefined character fields, such as goals and states, while leaving broader profiles and relationships unchanged; its world agent also updates only a global event field and does not maintain world state changes. In contrast, we support open-schema character updates and world-state updates even to the entity level.

For a fair comparison, we evaluate both frameworks on our test set with four representative untrained backbones. Tables~\ref{tab:character_comparison_bw} and~\ref{tab:world_comparison_bw} report metrics shared by both frameworks. Across all backbones, \ours{} achieves higher average scores for both character-agent and world-model evaluation, with especially large gains in character evolution metrics such as PUF, PES, and MQ. For world modeling, the main improvements appear in long-range scene continuity and turn/scene organization.

Figure~\ref{fig:length_eval_frameworks} further examines PES and SCC as indicators of long-horizon evolution, showing that BookWorld degrades on both metrics as simulations become longer. But our \ours{} mitigates this trend, and its PES even improves over longer trajectories, suggesting that structured state updates better support long-horizon evolution.

\section{Conclusion}

We presented \ours{}, a framework and benchmark for simulating interactive, persistently evolving worlds. By coupling a Character Agent with a World Model, maintaining explicit open-schema states, and decomposing long-horizon interaction into seven trainable tasks, \ours{} provides a concrete foundation for studying persistent world evolution beyond isolated role-play. Our results show that this design leads to more coherent long-horizon simulations across diverse backbones. We hope \ours{} can serve as a useful step toward richer literary  agents, controllable interactive worlds, and long-horizon role-playing systems.

\clearpage
\section*{Limitations}

\ours{} has three main limitations. First, it models the world as a single objective state shared by all characters, whereas literary characters often perceive and remember the same world differently. For example, one character may view the world as benevolent while another sees it as hostile, and a character may misremember an object as being on a table when it is actually on a chair. Such subjective perceptions and imperfect memories can shape character behavior, but maintaining separate perceived worlds for individual characters would substantially increase system complexity. We hope future work will address subjective world modeling in a comprehensive yet practical way.

Second, our world representation is constrained by the limited context length of LLMs. The World Model tracks only important entities at each location, rather than all entities in the environment. Future work may benefit from other modalities, such as information-dense visual inputs, to construct more detailed world representations.

Third, due to copyright constraints, our benchmark is built from public-domain classic books from Project Gutenberg, and does not cover domains such as modern novels, games, or user-created worlds. However, this is mainly a constraint of the data rather than of the framework. Since \ours{} is fully open-schema and uses no domain-specific ontologies or predefined attribute slots, it can in principle adapt to other narrative domains. Even within the current corpus, the data already span multiple genres. And our OOD experiments (Appendix~\ref{app:id_ood_results}) further show consistent gains on entirely unseen books. Broader cross-domain evaluation remains important future work.

\section*{Ethics Statement}

\paragraph{Source Literary Works.}
\ours{} is derived from public-domain literary works obtained from Project Gutenberg, a free digital library dedicated to digitizing and providing access to books whose copyrights have expired. The data consist of transformed and abstracted representations rather than verbatim reproductions of the original books.

\paragraph{Use of Human Annotations.}
Our institution recruited native-English-speaking annotators to assess the agreement between LLM judges and human judges. Throughout the annotation process, we ensured that annotators' privacy rights were respected and that their participation was voluntary and informed. Annotators received compensation exceeding the local minimum wage and consented to the use of the annotations they provided for research purposes. We also maintained transparency about the purpose of the study and how the annotations would be used. Appendix~\ref{app:human_eval} provides further details on the human evaluation protocol and annotation instructions.

\paragraph{Risks.}
\ours{} is derived from literary works and further augmented with LLM-generated annotations, both of which may raise ethical considerations. The source books may contain toxic, stereotypical, or discriminatory language, and LLMs may introduce or amplify biases during annotation. Therefore, although the data are used for research purposes, we cannot guarantee that they are entirely free from harmful content. The fictional content, character behaviors, and generated analyses in the dataset should not be interpreted as representing the authors' viewpoints. We encourage users of the dataset and benchmark to account for these risks and to avoid deploying the data or models in settings where harmful fictional content or biased generations could affect real individuals.

\paragraph{Data Use Restrictions.}
We emphasize that our dataset and benchmark are intended exclusively for scientific research and not for commercial use. We provide these resources solely for academic use and disclaim responsibility for any misuse beyond their intended research scope.

\bibliography{custom}

\newpage
\appendix

\begin{center}
    {\Large\textbf{Appendices}}
\end{center}


\section{Partial Support Clarifications}
\label{app:partial_support_clarifications}

This section provides the rationale for the partial-support marks in Table~\ref{tab:framework_comparison}.

\noindent\textbf{BookWorld~\citep{BookWorld}.}
We mark BookWorld as partial support for profile updates because it updates only several character fields, while leaving broader profile information and relations unchanged. For global world state updates, BookWorld maintains a global event, but other components of the world setting remain static. For location/entity-level state modeling, BookWorld provides only a brief description for each location, without entity-level modeling or updates to these world states.

\noindent\textbf{GenerativeAgents~\citep{GenerativeAgents}.}
We mark GenerativeAgents as partial support for profile updates because it updates a character's memory and then queries the memory to infer the character's latest status, so the updated information is still limited to predefined fields. We also mark it as partial support for scene initialization because its simulation proceeds at the granularity of days: all characters act each day, which differs from interactive literary worlds where updates occur scene by scene and each scene usually involves only a subset of characters. Its updates are therefore closer to everyday routines and minor daily activities than to scene-level literary progression.

\noindent\textbf{CharacterBox~\citep{CharacterBox}.}
We mark CharacterBox as partial support for profile updates because it mainly updates predefined attributes such as BDI (Belief-Desire-Intention) and position, while leaving broader Profile \& Traits unchanged. We also mark it as partial support for location/entity-level state modeling because it simulates a single scene, with only one location-level environment description and state update, and therefore does not support long-horizon state updates across multiple locations and scenes.

\section{Training and Inference Details}
\label{app:training_inference_details}

We fine-tune all EW models with supervised fine-tuning using LoRA in LLaMA-Factory\footnote{\url{https://llamafactory.readthedocs.io/en/latest/}}. Unless otherwise specified, we use LoRA rank 64, LoRA alpha 128, and LoRA dropout 0.05. We train for 2 epochs with a learning rate of $2\times10^{-5}$, a maximum sequence length of 32,768 tokens, a per-device batch size of 1, and gradient accumulation of 64, yielding an effective batch size of 64 per GPU. We use bf16 precision, gradient checkpointing, and FlashAttention-2, and hold out 10\% of the merged training data as the validation set. After training, we serve models with vLLM\footnote{\url{https://vllm.ai/}} for test-time inference and use 50 concurrent requests.

\section{Data Construction Details}
\label{app:data_construction_details}

This appendix expands the data construction pipeline summarized in \S\ref{sec:data_pipeline}. We first present the three main construction phases, then describe additional cleaning steps that are interleaved in implementation but separated here for readability.

\subsection{Main Construction Phases}

\noindent\textbf{Book Selection.}
Following Coser~\citep{CoSER} and AdaMARP~\citep{AdaMARP}, we select 57 representative books from Goodreads' Best Books Ever list\footnote{\url{https://www.goodreads.com/list/show/1.Best_Books_Ever}} and obtain their full texts from Project Gutenberg\footnote{\url{https://www.gutenberg.org/}}, a free digital library of public-domain books. All selected books are narrated in chronological order, which ensures that the temporal progression of scenes aligns with the reading order, a prerequisite for our scene-by-scene state-tracking pipeline.

\noindent\textbf{Phase 1: Scene Extraction.}
The book text is split into chunks for LLM processing. Each chunk is used to extract structured scenes, where each scene contains a summary, a scenario description, a list of key characters with brief descriptions, and a sequence of multi-turn interactions. Each interaction is represented as an actor-content pair: the actor can be a single character or a character group, allowing events such as an entire family entering together to be captured as one shared interaction. The content may include thought, speech, and action, marked consistently with the simulation format: thoughts are enclosed in \texttt{[...]}, speech is written as plain text, and actions are enclosed in \texttt{(...)}. The extraction supports cross-chunk continuation: when a scene is truncated at the end of one chunk, the truncated text is prepended to the next chunk, so the LLM can continue the same scene with the missing context.

\begin{table}[t]
\centering
\small
\begin{tabular}{lr}
\toprule
\textbf{Statistic} & \textbf{Count} \\
\midrule
\multicolumn{2}{c}{\textit{Extracted Structured Data}} \\
Total Books & 57 \\
Total Scenes & 9,763 \\
Total Interactions & 132,800 \\
\midrule
\multicolumn{2}{c}{\textit{Extracted Entities}} \\
Unique Characters & 3,311 \\
Unique Locations & 1,888 \\
\midrule
\multicolumn{2}{c}{\textit{Constructed Dataset}} \\
Train & 138,596 \\
Test (ID / OOD) & 222 (116 / 106) \\
\bottomrule
\end{tabular}
\caption{Overview of \ours{} dataset statistics.}
\label{tab:dataset_overview}
\vspace{-0.15in}
\end{table}

\begin{table*}[t]
\centering
\small
\begin{tabular}{lll}
\toprule
\multicolumn{3}{c}{\textbf{Selected Books}} \\
\midrule
\textbf{1.} \textit{A Doll's House} & \textbf{2.} \textit{A Little Princess} & \textbf{3.} \textit{A Tale of Two Cities} \\
\textbf{4.} \textit{Anthem} & \textbf{5.} \textit{Black Beauty} & \textbf{6.} \textit{Don Quixote} \\
\textbf{7.} \textit{Dr Jekyll and Mr Hyde} & \textbf{8.} \textit{Far From the Madding Crowd} & \textbf{9.} \textit{Great Expectations} \\
\textbf{10.} \textit{Gulliver's Travels} & \textbf{11.} \textit{Heart of Darkness} & \textbf{12.} \textit{Jude the Obscure} \\
\textbf{13.} \textit{Julius Caesar} & \textbf{14.} \textit{Little Women} & \textbf{15.} \textit{Madame Bovary} \\
\textbf{16.} \textit{Mansfield Park} & \textbf{17.} \textit{Middlemarch} & \textbf{18.} \textit{Much Ado About Nothing} \\
\textbf{19.} \textit{My \'Antonia} & \textbf{20.} \textit{Northanger Abbey} & \textbf{21.} \textit{Notes from Underground} \\
\textbf{22.} \textit{Oliver Twist} & \textbf{23.} \textit{Othello} & \textbf{24.} \textit{Pride and Prejudice} \\
\textbf{25.} \textit{Sense and Sensibility} & \textbf{26.} \textit{Siddhartha} & \textbf{27.} \textit{Tess of the D'Urbervilles} \\
\textbf{28.} \textit{The Adventures of Tom Sawyer} & \textbf{29.} \textit{The Age of Innocence} & \textbf{30.} \textit{The Call of the Wild} \\
\textbf{31.} \textit{The House of Mirth} & \textbf{32.} \textit{The Jungle} & \textbf{33.} \textit{The Metamorphosis} \\
\textbf{34.} \textit{The Phantom of the Opera} & \textbf{35.} \textit{The Picture of Dorian Gray} & \textbf{36.} \textit{The Pilgrim's Progress} \\
\textbf{37.} \textit{The Portrait of a Lady} & \textbf{38.} \textit{The Scarlet Letter} & \textbf{39.} \textit{The Secret Garden} \\
\textbf{40.} \textit{The Sorrows of Young Werther} & \textbf{41.} \textit{The Sun Also Rises} & \textbf{42.} \textit{The Tempest} \\
\textbf{43.} \textit{The Three Musketeers} & \textbf{44.} \textit{The Turn of the Screw} & \textbf{45.} \textit{The Wind in the Willows} \\
\textbf{46.} \textit{Treasure Island} & \textbf{47.} \textit{Uncle Tom's Cabin} & \textbf{48.} \textit{White Fang} \\
\midrule
\multicolumn{3}{l}{\textbf{49.} \textit{A Portrait of the Artist as a Young Man}} \\
\multicolumn{3}{l}{\textbf{50.} \textit{Alice's Adventures in Wonderland / Through the Looking-Glass}} \\
\multicolumn{3}{l}{\textbf{51.} \textit{Around the World in Eighty Days}} \\
\multicolumn{3}{l}{\textbf{52.} \textit{The Adventures of Huckleberry Finn}} \\
\multicolumn{3}{l}{\textbf{53.} \textit{The Adventures of Sherlock Holmes (Sherlock Holmes, \#3)}} \\
\multicolumn{3}{l}{\textbf{54.} \textit{The Hound of the Baskervilles (Sherlock Holmes, \#5)}} \\
\multicolumn{3}{l}{\textbf{55.} \textit{The Importance of Being Earnest}} \\
\multicolumn{3}{l}{\textbf{56.} \textit{The Murder of Roger Ackroyd (Hercule Poirot, \#4)}} \\
\multicolumn{3}{l}{\textbf{57.} \textit{Twenty Thousand Leagues Under the Sea (Captain Nemo, \#2)}} \\
\bottomrule
\end{tabular}
\caption{The 57 selected books from Goodreads' Best Books Ever list.}
\label{tab:selected_books}
\vspace{-0.1in}
\end{table*}

\noindent\textbf{Phase 2: Character Construction.}
For each character, we provide the LLM with extracted mentions and descriptions to identify aliases and determine a single official name (e.g., ``Mr.\ Smith'', ``John'', ``Father'' $\to$ ``John Smith''). All references in scenes are then replaced with standardized names for cross-scene consistency. To initialize characters, we provide the LLM with the first few scenes involving each character and ask it to summarize an open-schema profile before the main story events unfold. Finally, profiles are updated scene by scene: after each scene, the LLM revises the profile, short description, and hidden tracker for every participating character based on what occurred and what later scenes reveal.

\noindent\textbf{Phase 3: World Construction.}
Locations are extracted from scenes and undergo the same alias-standardization process as characters. The pipeline then generates an initial world state with a global state and per-location states. World states are updated interaction by interaction: after each interaction within a scene, the LLM decides whether persistent changes to the global or location state should be recorded.

\subsection{Interleaved Cleaning Details}

The following steps are shown after the main phases to keep the high-level pipeline clear, but they are interleaved in implementation. In particular, scene and interaction cleaning is applied after initial scene extraction and before the cleaned scenes are used for character construction, world construction, and task generation.

\noindent\textbf{Interaction Refinement.}
Each interaction's content is sent to an LLM for refinement. The goal is to improve the clarity, consistency, and readability of interaction expressions while preserving their semantic meaning and narrative voice. In particular, the refinement removes redundant thoughts that are accidentally embedded in speech, keeps private thoughts separated from spoken utterances, and normalizes each interaction into the first-person perspective of its acting character or character group. Consecutive interactions performed by the same set of characters are also merged into a single interaction, removing artificial turn boundaries introduced by the chunked extraction process and producing more natural multi-turn sequences.

\noindent\textbf{Duplicate Scene Removal.}
Because chunk boundaries can cause the same narrative event to be extracted twice (once at the end of one chunk and once at the beginning of the next), an LLM is used to detect pairs of scenes describing the same event. For each duplicate pair, the scene with less information is removed while the richer version is retained.

\noindent\textbf{Scene Enhancement.}
Each scene's scenario description is enhanced by an LLM to add dramatic setup details, atmospheric context, and implicit tensions that were present in the source text but not captured during initial extraction. This produces richer scenario descriptions for the \texttt{location\_scenario} training task.

\begin{figure*}[t]
    \centering
    \begin{subfigure}[t]{0.48\textwidth}
        \centering
        \includegraphics[width=\textwidth]{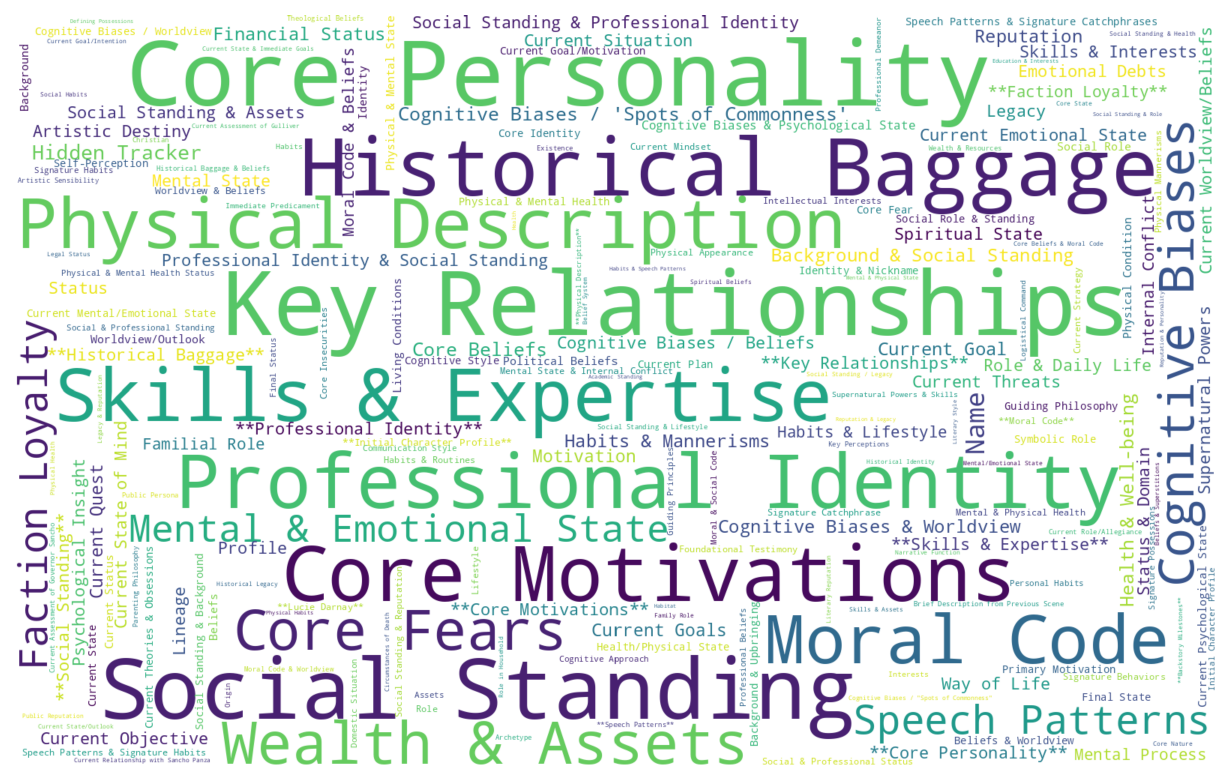}
        \caption{Character Profile Dimensions}
        \label{fig:dim_wc_char}
    \end{subfigure}
    \hfill
    \begin{subfigure}[t]{0.48\textwidth}
        \centering
        \includegraphics[width=\textwidth]{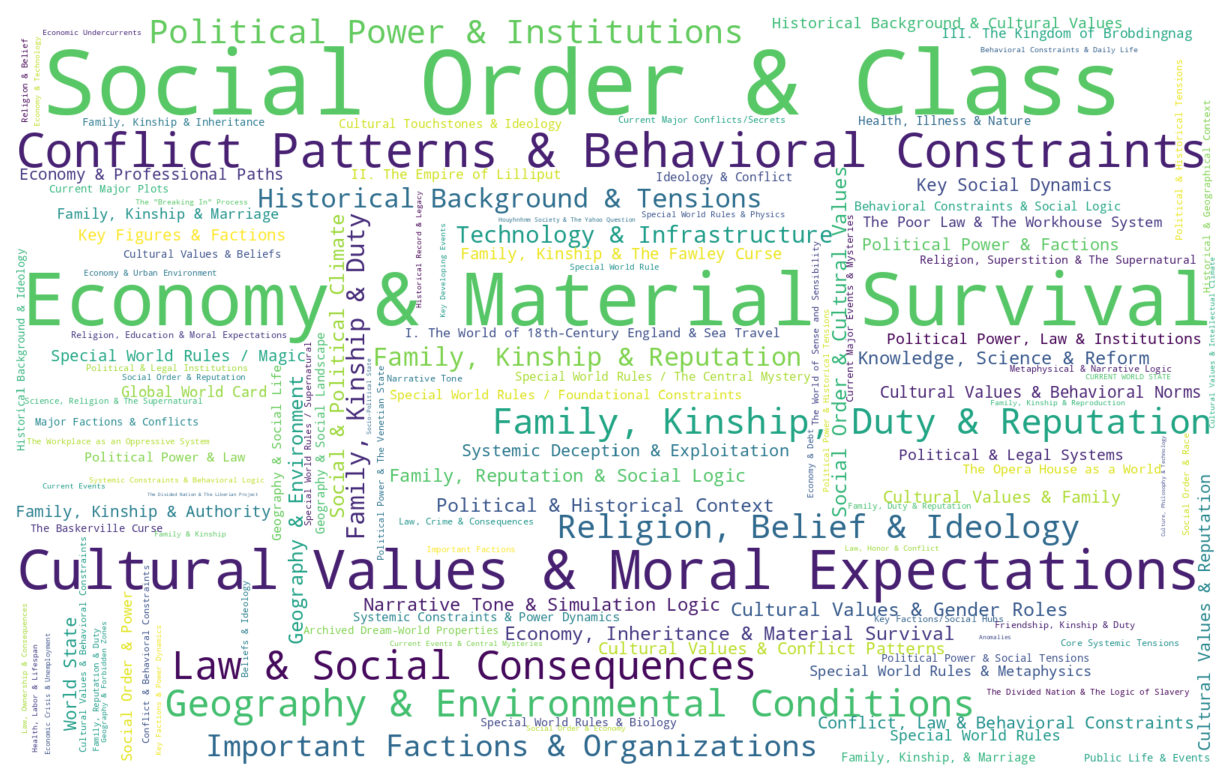}
        \caption{Global World State Dimensions}
        \label{fig:dim_wc_world}
    \end{subfigure}
\caption{Word clouds of state dimensions under the open-schema design for character profiles (left) and global world states (right). Font size is proportional to frequency across the 57-book corpus.}
    \label{fig:dimension_wordcloud}
\vspace{-0.05in}
\end{figure*}

\begin{figure}[t]
    \centering
    \includegraphics[width=0.95\columnwidth]{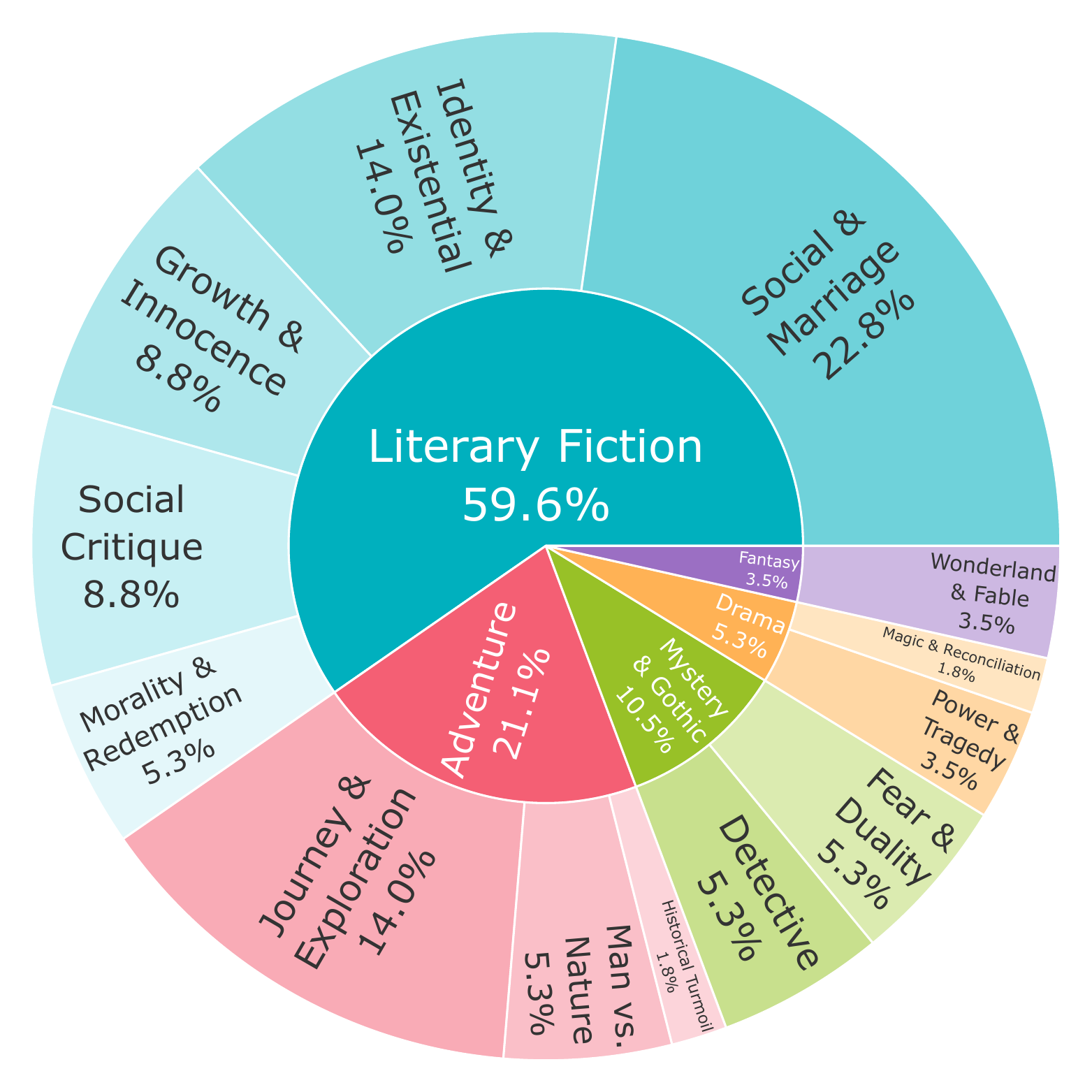}
    \caption{Two-level genre distribution of the 57 books in \ours{}. The inner ring shows five coarse genres; the outer ring breaks each genre into thematic sub-categories.}
    \label{fig:genre_dist}
\vspace{-0.1in}
\end{figure}

\subsection{Training and Test Data Construction}

After obtaining the character states, global world state, and location states sequentially updated along the book timeline, we convert the timeline into task-level supervision according to the simulator's execution order. At a scene boundary, \texttt{scene\_cast} takes the previous scene content together with the current global state, location states, and character states, and predicts whether the next scene exists and which characters participate; conditioned on this cast, \texttt{location\_scenario} predicts the next location and scenario. Once the next scene is fixed, \texttt{motivation\_update} uses the previous scene, the planned next scene, and the character state to infer each participating character's entering motivation.

Within a scene, each interaction history is used as input to supervise the next actor selection and interaction generation. \texttt{next\_character} predicts the next actor or scene termination from the current scenario and visible interaction history. \texttt{interaction\_gen} then generates that actor's next interaction from the same history together with the actor's character state and motivation. After each generated interaction, \texttt{world\_update} decides whether the latest interaction requires a persistent update to the global or location state. Finally, after a scene is completed, \texttt{character\_update} updates each relevant character state from the completed scene. Since \texttt{interaction\_gen} and \texttt{character\_update} are naturally formulated as multi-turn conversations, we store them as ShareGPT-style conversation examples following CoSER~\citep{CoSER} and AdaMARP~\citep{AdaMARP}; the remaining tasks use task-specific prompts with structured targets under the same SFT interface. In this way, one extracted timeline is decomposed into seven supervised tasks defined in \S\ref{sec:task_formalization}.

For evaluation, we instead extract simulation snapshots at selected time points. Each snapshot stores the current character states, global state, and location states, together with the previous scene used for reference. We split books into three groups: 10\% are held out as OOD books and never used for training; among the remaining books, half are train/test books and half are train-only books. For train/test books, the first 70\% of scenes are used for training and snapshots are sampled from the remaining 30\% as in-domain tests. Train-only books contribute all scenes to training. OOD books contribute no training examples; their test snapshots are sampled from the latter 70\% of scenes. We sample only valid candidate scenes with at least five interactions, using at most five ID snapshots per train/test book and at most twenty OOD snapshots per OOD book.

\begin{figure*}[t]
    \centering
    \includegraphics[width=0.85\textwidth]{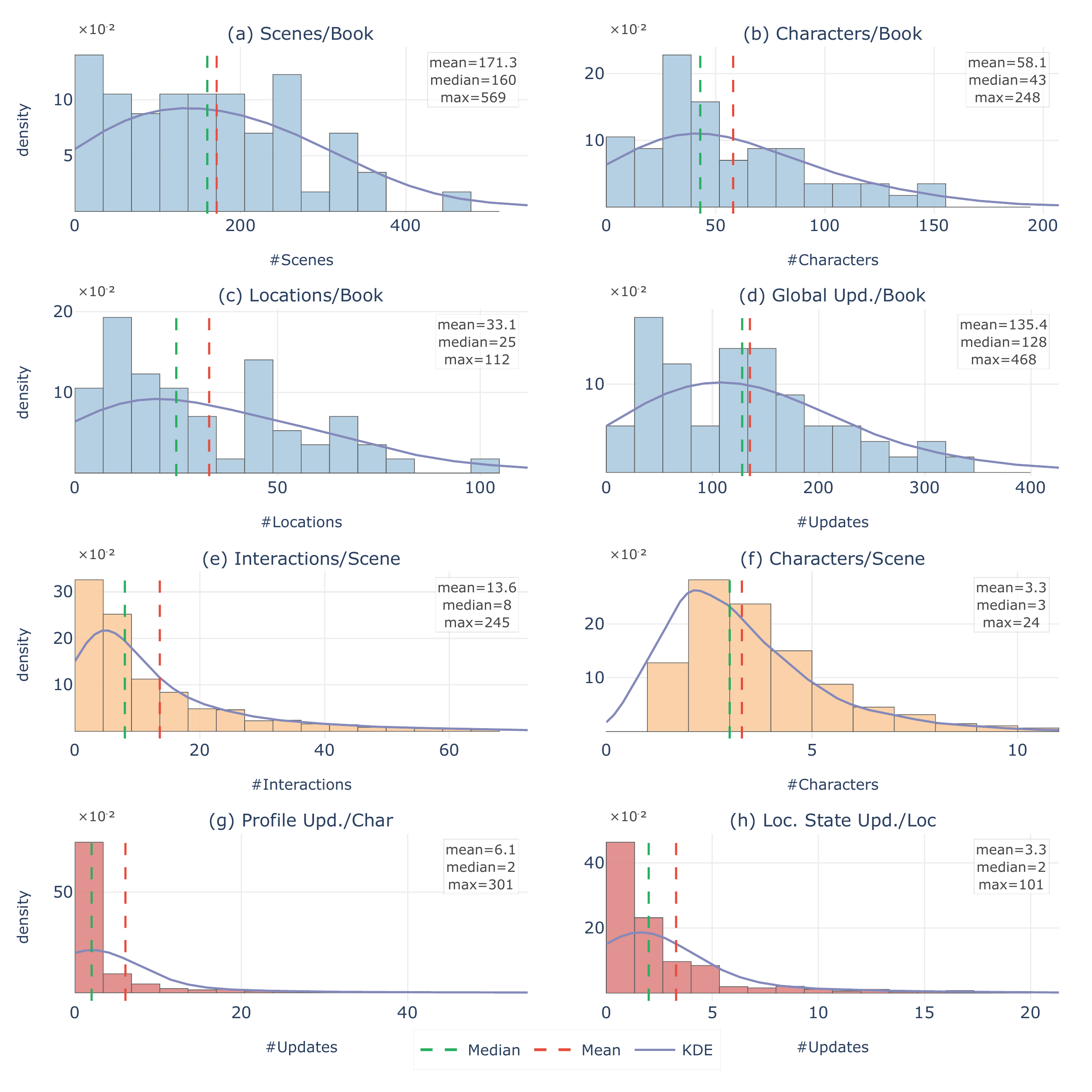}
\vspace{-0.05in}
    \caption{Distribution of extracted data statistics across four granularity levels. Each panel shows a histogram with KDE curve; dashed lines indicate the mean (red) and median (blue).}
    \label{fig:extracted_data_stats}
\vspace{-0.1in}
\end{figure*}


\section{Dataset Statistics}
\label{app:dataset_stats}

\subsection{Overview}
\label{app:dataset_overview}

Table~\ref{tab:dataset_overview} summarizes the key statistics of \ours{} dataset. Table~\ref{tab:selected_books} lists all 57 selected books.

\paragraph{Schema Distribution.}
A distinctive feature of \ours{} is that both character profiles and global world states adopt an \emph{open schema}: rather than prescribing a fixed set of attributes, the extraction model freely selects whichever dimensions are most salient for a given character or narrative context. Figure~\ref{fig:dimension_wordcloud} visualizes the resulting dimension vocabularies as word clouds, where font size reflects frequency across the corpus. Character profiles (left) are dominated by dimensions such as \textit{Social Standing}, \textit{Core Personality}, \textit{Key Relationships}, and \textit{Professional Identity}, while world states (right) center on \textit{Cultural Values \& Moral Expectations}, \textit{Social Order \& Class}, and \textit{Economy \& Material Survival}. The long tail of less frequent dimensions (581 unique character dimensions, 136 world-state dimensions) demonstrates the schema's flexibility in capturing diverse narrative elements.

\begin{table}[t]
\centering
\small
\begin{tabular}{lrr}
\toprule
\textbf{Module / Task} & \textbf{Samples} & \textbf{Asst.\ Turns} \\
\midrule
\multicolumn{3}{l}{\textit{World Model}} \\
\quad \texttt{scene\_cast}          & 7,983  & 7,983   \\
\quad \texttt{location\_scenario}   & 7,958  & 7,958   \\
\quad \texttt{next\_character}      & 14,315 & 116,789 \\
\quad \texttt{world\_update}        & 17,832 & 17,832  \\
\cmidrule{2-3}
\quad \textit{Subtotal}             & 48,088 & 150,562 \\
\midrule
\multicolumn{3}{l}{\textit{Character Agent}} \\
\quad \texttt{interaction\_gen}     & 40,554 & 108,831 \\
\quad \texttt{character\_update}    & 24,977 & 24,977  \\
\quad \texttt{motivation\_update}   & 24,977 & 24,977  \\
\cmidrule{2-3}
\quad \textit{Subtotal}             & 90,508 & 158,785 \\
\midrule
\textbf{Total}                      & \textbf{138,596} & \textbf{309,347} \\
\bottomrule
\end{tabular}
\caption{Training data statistics per task. ``Samples'' = number of ShareGPT style conversations; ``Asst.\ Turns'' = total assistant responses (equals Samples for single-turn tasks).}
\label{tab:training_data}
\vspace{-0.1in}
\end{table}

\paragraph{Genre Distribution.}
To characterize the literary diversity of the corpus, we categorize all 57 books along two levels: a coarse \emph{genre} layer and a finer \emph{thematic} layer. As shown in Figure~\ref{fig:genre_dist}, the collection is dominated by Literary Fiction (59.6\%), followed by Adventure (21.1\%), Mystery \& Gothic (10.5\%), Drama (5.3\%), and Fantasy (3.5\%). Within each genre, thematic sub-categories further distinguish works by narrative focus. For example, Literary fiction spans Social \& Marriage, Identity \& Existential, Growth \& Innocence, Social Critique, and Morality \& Redemption. This breadth ensures that the benchmark exercises a wide range of character dynamics, world-building patterns, and narrative structures.

\paragraph{Extracted Data Distribution.}
Figure~\ref{fig:extracted_data_stats} presents the distribution of eight key statistics across the extracted structured data, organized by four granularity levels. At the \emph{book level}, the number of scenes, characters, and locations per book varies considerably (median 160 scenes, 43 characters, and 25 locations), reflecting the diverse scale and complexity of the source novels. At the \emph{scene level}, each scene typically involves a small number of interactions and characters, though long-tail cases exist for densely populated scenes. At the \emph{character level}, profile update counts capture how frequently each character evolves throughout the narrative. At the \emph{world level}, global state updates per book and location-specific state updates per location quantify the dynamism of the story world. Together, these distributions confirm that \ours{} covers a broad spectrum of narrative complexity, from compact novellas to sprawling multi-character epics.

\begin{figure}[t]
    \centering
    \includegraphics[width=\columnwidth]{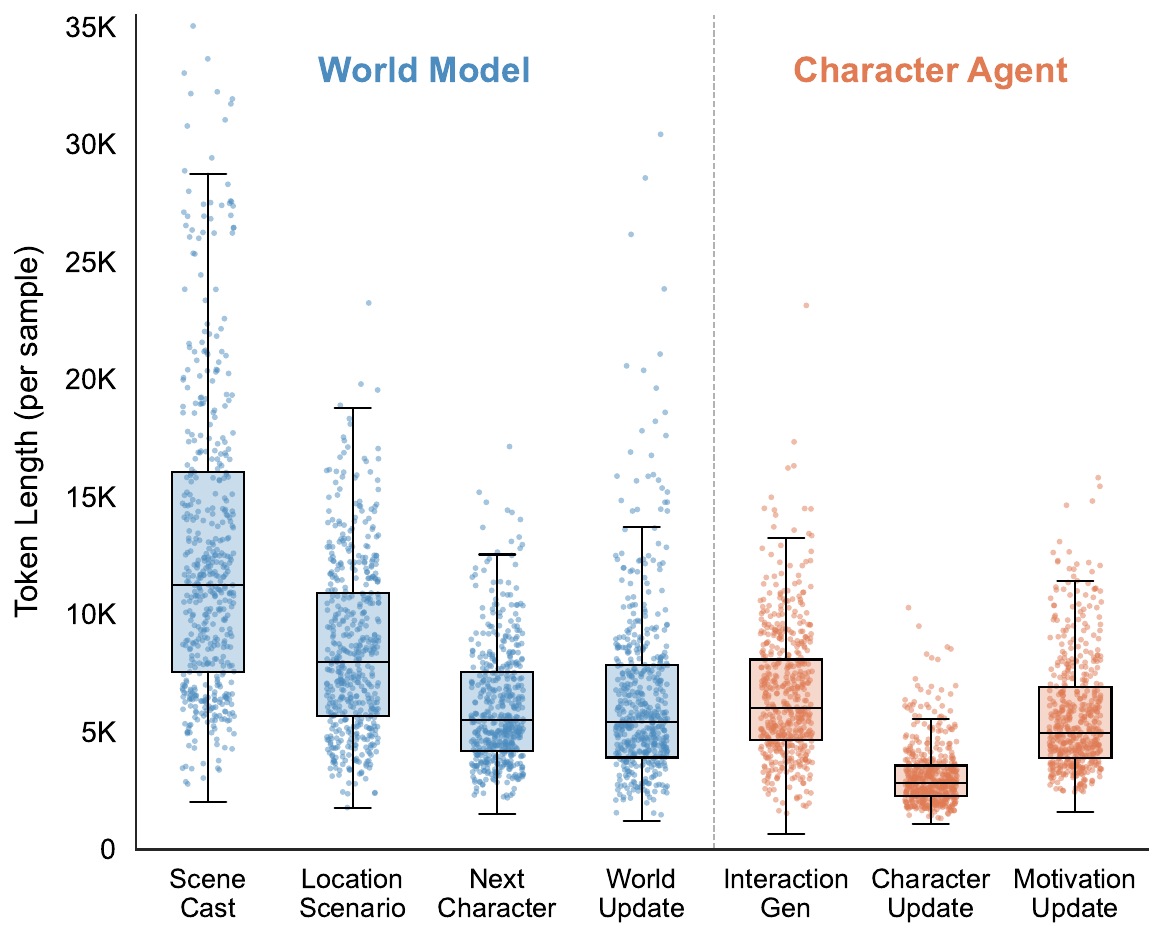}
    \caption{Token length distribution per task.}
    \label{fig:token_length_dist}
\vspace{-0.1in}
\end{figure}

\subsection{Training Data Statistics}
\label{app:training_data}

Table~\ref{tab:training_data} summarizes the training data for each task of the World Model and Character Agent. The ``Samples'' column reports the number of ShareGPT-format training conversations. For multi-turn tasks (\texttt{next\_character} and \texttt{interaction\_gen}), each sample contains multiple user--assistant exchanges within a single conversation; the ``Asst.\ Turns'' column gives the total number of assistant responses across all samples. Single-turn tasks have exactly one assistant turn per sample.

\paragraph{Token Length Distribution.}
\label{app:token_length}

Figure~\ref{fig:token_length_dist} shows the per-sample token length distribution for each of the seven tasks, measured with the \texttt{cl100k\_base} tokenizer (GPT-4 family). Each sample corresponds to a complete ShareGPT-format conversation, including all system, user, and assistant turns.

\texttt{scene\_cast} exhibits the longest sequences (median $\approx$ 11.2k) because it must select the next scene's cast from \textit{all} characters; to control input length, each character is represented by the latest short description rather than the full profile. \texttt{location\_scenario} follows (median $\approx$ 8.0k) for an analogous reason: it chooses the scene location from \textit{all} candidate locations; here each candidate location is represented by its description, without expanding the states of all contained entities. The four mid-range tasks, \texttt{next\_character}, \texttt{world\_update}, \texttt{interaction\_gen}, and \texttt{motivation\_update} (medians 4.9k--6.0k), are more compact because none of them requires enumerating all characters or all locations simultaneously. \texttt{character\_update} is the shortest (median $\approx$ 2.8k) as it only updates a single character's profile based on one scene's events.

These distributions inform our choice of maximum sequence length and packing strategy for supervised fine-tuning. We accordingly set the training cutoff length to 32{,}768 tokens to accommodate the long-tail samples while keeping computation tractable.


\section{Evaluation Framework}
\label{sec:eval_framework}

\subsection{Simulation Protocol}
\label{sec:app_simulation_protocol}

We evaluate \ours{} with a \textbf{multi-scene simulation protocol} designed for persistent book-to-world simulation. Starting from each held-out structured snapshot, we run the simulator forward to generate a multi-scene trajectory containing scene plans, interaction histories, world-state updates, and character-state updates. This design allows us to evaluate not only local response quality, but also whether a model can maintain coherent world and character evolution across scenes.

Concretely, each evaluation sample provides the initial simulator state, including the current World State, character Full Profiles, and source scene. The evaluated system then produces a complete generated trajectory which includes the scene cast and location-scenario plan for each scene, the within-scene interaction sequence, the updated World State, and the updated Full Profiles written back after the scene ends.

Algorithm~\ref{alg:simulation} summarizes the full simulation procedure.

\begin{algorithm}[t]
\caption{\ours{} simulation with a World Model and a Character Agent}
\label{alg:simulation}
\footnotesize
\begin{algorithmic}[1]
\Require Initial states $S_w^{\mathcal{L},(0)}$, $S_c^{\mathcal{I},(0)}$; World Model policies $\pi_w$; Character Agent policies $\pi_c$; max scenes $T$; max turns $K$
\State $\tau \gets [\,]$
\For{$t = 1$ to $T$}
    \State $z_t \sim \pi_{\mathrm{cast}}(\cdot \mid O_w^{\mathcal{I},\varnothing,(t)})$
    \If{$z_t = \varnothing$} \State \textbf{break} \EndIf
    \State $r_t \sim \pi_{\mathrm{loc}}(\cdot \mid O_w^{z_t,\mathcal{L},(t)}, z_t)$; let $\ell_t$ be its location
    \ForAll{$i \in z_t$}
        \State $M^{i,(t)} \sim \pi_{\mathrm{mot}}(\cdot \mid O_c^{i,\ell_t,(t)}, r_t)$
    \EndFor
    \State $Y_t \gets [\,]$, $S_w^{\ell_t,(t,1)} \gets S_w^{\ell_t,(t)}$
    \For{$k = 1$ to $K$}
        \State $i_{t,k} \sim \pi_{\mathrm{next}}(\cdot \mid O_w^{z_t,\{\ell_t\},(t,k)}, r_t, Y_t)$
        \If{$i_{t,k} = \texttt{END}$} \State \textbf{break} \EndIf
        \State $y_{t,k} \sim \pi_{\mathrm{int}}(\cdot \mid O_c^{i_{t,k},\ell_t,(t)}, r_t, Y_t)$
        \State $Y_t \gets Y_t \mathbin{\Vert} [y_{t,k}]$
        \State $S_w^{\ell_t,(t,k+1)} \sim \pi_{\mathrm{wu}}(\cdot \mid O_w^{z_t,\{\ell_t\},(t,k)}, y_{t,k}, Y_t)$
    \EndFor
    \State $K_t \gets |Y_t|$, $S_w^{\ell_t,(t+1)} \gets S_w^{\ell_t,(t,K_t+1)}$
    \ForAll{$i \in z_t$}
        \State $S_c^{i,(t+1)} \sim \pi_{\mathrm{cu}}(\cdot \mid O_c^{i,\ell_t,(t)}, Y_t, S_w^{\ell_t,(t+1)})$
    \EndFor
    \State Carry forward unchanged states for $i \notin z_t$ and $\ell \neq \ell_t$
    \State Append $(z_t, r_t, Y_t, O_w^{\mathcal{I},\mathcal{L},(t+1)})$ to $\tau$
\EndFor
\State \Return $\tau$
\end{algorithmic}
\end{algorithm}

\subsection{Trajectory-Level Evaluation}

Because \ours{} introduces explicit persistent states and structured simulator outputs, scene-level dialogue evaluation~\citep{AdaMARP} alone is insufficient. We therefore adopt a \textbf{trajectory-level LLM-as-Judge protocol} that scores both the \textbf{Character Agent} and the \textbf{World Model} on complete simulated trajectories containing many scenes rather than isolated responses. The evaluation system comprises \textbf{10 dimensions and 20 sub-metrics}, divided into two scoring modules:

\begin{itemize}
    \item \textbf{CHARACTER Score} (Character Agent Evaluation): 6 dimensions, 11 sub-metrics
    \item \textbf{WORLD Score} (World Model Evaluation): 4 dimensions, 9 sub-metrics
\end{itemize}

For \textbf{Character Agent} evaluation, the judge is given the relevant character profile information, scene motivations, world state context, and the generated trajectory segments needed for that metric. For \textbf{World Model} evaluation, the judge receives the initial simulator context together with the generated scene plans, turn orchestration decisions, and persistent state updates. This separation mirrors the modular structure of \ours{} and allows us to diagnose planning errors and embodiment errors independently.

Each sub-metric is scored on a 0--100 scale with a base score of 50. The judge first identifies \textbf{Merits} (excellent aspects, each awarded 1 to 10) and then \textbf{Demerits} (problematic aspects, each penalized 1 to 10). The final score is computed as $\min(100, \max(0, 50 + \sum \text{merits} - \sum \text{demerits}))$.

\begin{table}[t]
\centering
\small
\begin{tabular}{ll}
\toprule
\textbf{Failing Task} & \textbf{Penalized Metrics} \\
\midrule
\texttt{scene\_cast} & CSR, SCC \\
\texttt{location\_scenario} & LSR, SCC \\
\texttt{next\_character} & TSO \\
\texttt{world\_update} & GUS, GSA, LUS, LSA \\
\texttt{interaction\_gen} & PF, SSF, MDB, EA, EU, CR, NP \\
\texttt{character\_update} & PUF, PES \\
\texttt{motivation\_update} & MQ \\
\bottomrule
\end{tabular}
\caption{Task-to-metric mapping for the Metric Penalty. When a task causes simulation termination, its corresponding metrics receive the additional penalty described above.}
\label{tab:task_metric_map}
\vspace{-0.1in}
\end{table}

\begin{table*}[t]
\centering
\small
\resizebox{\linewidth}{!}{
\begin{tabular}{llll}
\toprule
 & \textbf{Character Agent (CHARACTER)} & & \textbf{World Model (WORLD)} \\
\midrule
\textbf{Dim I} & Character Consistency (PF, SSF, MDB) & \textbf{Dim I} & Scene Planning (CSR, LSR, SCC) \\
\textbf{Dim II} & Evolution Quality$^\star$ (PUF, PES) & \textbf{Dim II} & Speaker Management (TSO) \\
\textbf{Dim III} & Environmental Grounding (EA, EU) & \textbf{Dim III} & World State Maintenance$^\star$ (GUS, GSA, LUS, LSA) \\
\textbf{Dim IV} & Interaction Quality (CR, NP) & \textbf{Dim IV} & Instruction Compliance (IC) \\
\textbf{Dim V} & Motivation Generation (MQ) & & \\
\textbf{Dim VI} & Instruction Compliance (IC) & & \\
\midrule
\textbf{Total} & \textbf{11 sub-metrics} & & \textbf{9 sub-metrics} \\
\bottomrule
\end{tabular}}
\vspace{-0.05in}
\caption{Overview of the evaluation framework. $^\star$ denotes evaluation perspectives introduced by this framework. Grand total: 10 dimensions, 20 sub-metrics.}
\label{tab:eval_overview}
\vspace{-0.2in}
\end{table*}

\paragraph{Error Penalties.}
When a simulation terminates prematurely because a model fails to follow output format specifications, we apply two complementary penalties to ensure the error is reflected in the final scores. The error message contains the failing task name (e.g., \texttt{world\_update}, \texttt{interaction\_gen}), which we use to attribute the fault to either the World Model or the Character Agent. Infrastructure errors (e.g., API/Network error) are never attributed to the model and incur no penalty.

\textit{(1) IC Penalty (Instruction Compliance).}
We penalize the IC score of the responsible model using a logarithmic-decay formula based on $n$, the total number of calls made by that model before the failure point:
\begin{equation}
    \text{penalty}_{\text{IC}} = \min\!\left(50,\; \frac{50}{\ln(n+1)}\right)
\end{equation}
The final IC score is $\max(0,\; \text{IC}_{\text{judge}} - \text{penalty}_{\text{IC}})$, where $\text{IC}_{\text{judge}}$ denotes the raw IC score assigned by the LLM judge before any penalty. The logarithmic decay provides a smooth curve: a model that fails on its very first call receives the maximum penalty of 50 (driving IC to zero), while a model that succeeds hundreds of times before failing still incurs a meaningful penalty of $\sim$8--10 points. This ensures that ``the model eventually crashed due to format non-compliance'' is always reflected in the score, regardless of how many successful calls preceded the failure.

\textit{(2) Metric Penalty (Task-Specific Metrics).}
Beyond IC, the failing task also maps to specific quality metrics via a fixed task-to-metric correspondence (Table~\ref{tab:task_metric_map}). For these metrics, we apply a penalty that simulates the effect of an additional failed scene:
\begin{equation}
    \text{score} = \text{score}_{\text{judge}} \times \frac{N}{N+1}
\end{equation}
where $\text{score}_{\text{judge}}$ is the average score assigned by the LLM judge over the $N$ successfully completed scenes, and $\text{score}$ is the final reported score for that metric. This is equivalent to appending a zero-score virtual scene to the existing average. For samples that fail before producing any scene ($N=0$), the affected metrics are set directly to 0. IC metrics are excluded from this penalty (already handled above).

At aggregation time, scene-local metrics are scored for each generated scene and averaged across relevant scenes. Character-level trajectory metrics are aggregated over the scenes in which a character participates; specifically, \textbf{Profile Evolution Smoothness} is scored over each character's participating scenes and then averaged with weights proportional to character appearances. Full-trajectory metrics are computed over the complete generated trajectory; specifically, \textbf{Scene Continuity \& Coherence} is scored over the full simulation to measure long-range scene organization and world consistency. The final \textbf{CHARACTER} and \textbf{WORLD} scores are simple averages over their corresponding sub-metrics, with missing metrics treated as invalid rather than silently imputed.

\subsection{Evaluation Overview}

Table~\ref{tab:eval_overview} provides a summary of all evaluation dimensions and sub-metrics.

\subsection{Character Agent Evaluation (CHARACTER Score)}

\subsubsection{Dim I: Character Consistency}

Evaluates whether the character consistently embodies its profile throughout the interaction. Core question: if character's name was hidden, would the output still be recognizable as the character?

\noindent\textbf{Profile Fidelity (PF).}
Whether the character's knowledge, skills, and behavior are strictly confined within the profile scope.
Criteria: (a)~\textit{Knowledge Boundaries}: no knowledge or skills beyond profile settings (e.g., an ordinary farmer should not demonstrate advanced magical theory);
(b)~\textit{Background Consistency}: behavior matches the character's age, social class, and historical background;
(c)~\textit{Ability Constraints}: no abilities or privileges not documented in the profile appear out of nowhere.

\noindent\textbf{Speaking Style Fidelity (SSF).}
Whether the character's speaking style is consistent with the profile, natural, and free of obvious AI artifacts.
Criteria: (a)~\textit{Style Markers}: language features defined in the profile are used (catchphrases, sentence patterns, terminology, dialects);
(b)~\textit{Emotional Tone}: tone matches the character's personality rather than a generic ``AI assistant'' tone;
(c)~\textit{Naturalness}: language is fluent, human-like, avoiding templated or mechanical artifacts.

\noindent\textbf{Motivation-Driven Behavior (MDB).}
Whether the core motivation continuously drives decisions, and whether thought/action/speech are logically consistent.
Criteria: (a)~\textit{Behavioral Attribution}: decisions traceable to core motivation rather than random unmotivated behavior;
(b)~\textit{Trinity Coherence}: thought reasonably drives action and speech, with no unexplained contradictions among the three;
(c)~\textit{Value Stability}: core values do not suddenly change without a significant plot trigger.

\subsubsection{Dim II: Evolution Quality$^\star$}

Evaluates whether the dynamic updates to the character profile and hidden tracker are reasonable, coherent, and complete. Unique to this framework, focusing on character growth quality during interactions.

\noindent\textbf{Profile Update Fidelity (PUF).}
Whether the profile update and hidden tracker update jointly capture all changes that should be preserved from the scene, with appropriate threshold judgment between the two.
Criteria: (a)~\textit{Causal Chain}: every item written to profile or hidden tracker has a clear triggering event in the scene as evidence, with no fabricated information~\citep{liu2026naacl};
(b)~\textit{Growth/Signal Capture}: important changes that have crossed the threshold are written to profile; subtle signals that may accumulate in the future are written to hidden tracker;
(c)~\textit{Threshold Judgment}: major persistent changes should not remain only in the tracker without updating the profile; slight or ambiguous signals should not be prematurely written into the profile;
(d)~\textit{No Over/Under-Updating}: profile remains concise and stable, hidden tracker does not become a meaningless log; together they neither miss important changes nor over-react to irrelevant details.

\noindent\textbf{Profile Evolution Smoothness (PES).}
Whether the profile and hidden tracker jointly exhibit gradual, coherent, and appropriately-scaled evolution across scenes.
Criteria: (a)~\textit{Magnitude Matching}: casual conversations should only leave light signals or no change; major events trigger significant profile updates; sub-threshold content settles in the tracker first;
(b)~\textit{Gradualness}: personality, attitude, and relationship changes go through reasonable transitional stages; hidden tracker preserves intermediate evolution signals rather than allowing abrupt profile jumps;
(c)~\textit{Directional Consistency}: consecutive profile updates and tracker accumulations are logically coherent in direction; when tracker signals accumulate sufficiently, conversion to profile updates is natural and traceable.

\begin{table*}[t]
\centering
\huge
\resizebox{\linewidth}{!}{
\begin{tabular}{lcccccccccccc}
\toprule
\multicolumn{13}{l}{\textbf{CC}: Character Consistency \quad \textbf{EQ}: Evolution Quality \quad \textbf{EG}: Environmental Grounding \quad \textbf{IQ}: Interaction Quality \quad \textbf{MG}: Motivation Generation} \\
\multicolumn{13}{l}{\textbf{IC}: Instruction Compliance \quad \textbf{PF}: Profile Fidelity \quad \textbf{SSF}: Speaking Style Fidelity \quad \textbf{MDB}: Motivation-Driven Behavior \quad \textbf{PUF}: Profile Update Fidelity} \\
\multicolumn{13}{l}{\textbf{PES}: Profile Evolution Smoothness \quad \textbf{EA}: Environment Awareness \quad \textbf{EU}: Environmental Utilization} \\
\multicolumn{13}{l}{\textbf{CR}: Contextual Responsiveness \quad \textbf{NP}: Narrative Progression \quad \textbf{MQ}: Motivation Quality} \\
\midrule
\multirow{2}{*}{\textbf{Models}} & \multicolumn{3}{c}{\textbf{CC}} & \multicolumn{2}{c}{\textbf{EQ}} & \multicolumn{2}{c}{\textbf{EG}} & \multicolumn{2}{c}{\textbf{IQ}} & \multicolumn{1}{c}{\textbf{MG}} & \multicolumn{1}{c}{\textbf{IC}} & \multirow{2}{*}{\textbf{Avg.}} \\
\cmidrule(lr){2-4} \cmidrule(lr){5-6} \cmidrule(lr){7-8} \cmidrule(lr){9-10} \cmidrule(lr){11-11} \cmidrule(lr){12-12}
 & \textbf{PF} & \textbf{SSF} & \textbf{MDB} & \textbf{PUF} & \textbf{PES} & \textbf{EA} & \textbf{EU} & \textbf{CR} & \textbf{NP} & \textbf{MQ} & \textbf{IC} & \\
\midrule
\rowcolor{gray!15} \textbf{Closed-source} &  &  &  &  &  &  &  &  &  &  &  &  \\
Kimi-K2.5 & 80.79{\Large$\pm$10.84} & 61.45{\Large$\pm$19.00} & 87.52{\Large\phantom{0}$\pm$9.78} & 81.07{\Large\phantom{0}$\pm$8.63} & 54.51{\Large$\pm$16.38} & 81.29{\Large$\pm$10.59} & 83.86{\Large$\pm$13.00} & 88.31{\Large\phantom{0}$\pm$8.27} & 56.02{\Large$\pm$16.66} & 77.12{\Large$\pm$12.25} & 82.83{\Large\phantom{0}$\pm$5.68} & 75.89 \\
Gemini-2.5-Flash & 75.34{\Large\phantom{0}$\pm$3.94} & 61.68{\Large\phantom{0}$\pm$6.80} & 75.67{\Large\phantom{0}$\pm$5.96} & 68.61{\Large\phantom{0}$\pm$5.91} & 58.02{\Large\phantom{0}$\pm$7.68} & 67.02{\Large\phantom{0}$\pm$6.14} & 57.10{\Large\phantom{0}$\pm$8.90} & 75.72{\Large\phantom{0}$\pm$4.02} & 51.86{\Large\phantom{0}$\pm$9.17} & 74.61{\Large\phantom{0}$\pm$8.04} & 51.45{\Large$\pm$26.93} & 65.19\\
Gemini-2.5-Pro & 93.88{\Large\phantom{0}$\pm$3.47} & 85.94{\Large\phantom{0}$\pm$5.90} & 94.95{\Large\phantom{0}$\pm$3.44} & 83.28{\Large\phantom{0}$\pm$5.33} & 70.51{\Large\phantom{0}$\pm$8.65} & 89.53{\Large\phantom{0}$\pm$5.19} & 90.12{\Large\phantom{0}$\pm$6.15} & 93.63{\Large\phantom{0}$\pm$3.25} & 72.10{\Large\phantom{0}$\pm$8.19} & 77.20{\Large\phantom{0}$\pm$6.36} & \underline{86.06}{\Large\phantom{0}$\pm$3.08} & 85.20\\
Gemini-3.1-Pro-P & 89.79{\Large\phantom{0}$\pm$5.23} & 79.40{\Large\phantom{0}$\pm$6.91} & 93.09{\Large\phantom{0}$\pm$3.57} & 78.90{\Large\phantom{0}$\pm$5.87} & 66.29{\Large\phantom{0}$\pm$9.98} & 85.35{\Large\phantom{0}$\pm$6.17} & 83.90{\Large\phantom{0}$\pm$8.33} & 91.87{\Large\phantom{0}$\pm$3.62} & 67.74{\Large\phantom{0}$\pm$9.63} & 83.04{\Large\phantom{0}$\pm$6.41} & 84.12{\Large$\pm$12.90} & 82.14\\
GPT-4o & 71.45{\Large\phantom{0}$\pm$6.84} & 54.13{\Large\phantom{0}$\pm$9.15} & 76.36{\Large\phantom{0}$\pm$6.76} & 57.66{\Large\phantom{0}$\pm$8.60} & 54.46{\Large\phantom{0}$\pm$8.62} & 75.80{\Large\phantom{0}$\pm$7.04} & 76.32{\Large\phantom{0}$\pm$9.16} & 77.94{\Large\phantom{0}$\pm$4.63} & 55.89{\Large$\pm$11.73} & 69.94{\Large\phantom{0}$\pm$8.81} & 74.15{\Large$\pm$16.39} & 67.65 \\
GPT-5-Chat & 80.73{\Large\phantom{0}$\pm$5.60} & 73.44{\Large\phantom{0}$\pm$9.14} & 85.89{\Large\phantom{0}$\pm$5.45} & 71.33{\Large\phantom{0}$\pm$7.30} & 64.00{\Large\phantom{0}$\pm$9.68} & 86.74{\Large\phantom{0}$\pm$5.29} & 92.52{\Large\phantom{0}$\pm$6.49} & 87.36{\Large\phantom{0}$\pm$4.09} & 67.37{\Large\phantom{0}$\pm$9.19} & 81.49{\Large\phantom{0}$\pm$7.85} & 77.29{\Large$\pm$20.63} & 78.92 \\
GPT-5.1-Chat & 81.04{\Large\phantom{0}$\pm$9.50} & 69.93{\Large$\pm$11.17} & 75.27{\Large\phantom{0}$\pm$8.81} & 71.55{\Large\phantom{0}$\pm$4.89} & 70.94{\Large$\pm$10.26} & 73.41{\Large$\pm$10.56} & 71.94{\Large$\pm$10.62} & 77.37{\Large\phantom{0}$\pm$8.01} & 37.08{\Large$\pm$10.01} & 76.03{\Large\phantom{0}$\pm$9.84} & 81.37{\Large\phantom{0}$\pm$5.57} & 71.45 \\
GPT-5.3-Chat & 94.71{\Large\phantom{0}$\pm$2.75} & 88.77{\Large\phantom{0}$\pm$4.30} & 93.71{\Large\phantom{0}$\pm$2.95} & 77.90{\Large\phantom{0}$\pm$3.74} & \underline{80.16}{\Large\phantom{0}$\pm$6.62} & 91.62{\Large\phantom{0}$\pm$4.88} & 92.10{\Large\phantom{0}$\pm$7.54} & 92.27{\Large\phantom{0}$\pm$3.74} & 59.98{\Large\phantom{0}$\pm$7.55} & 83.94{\Large\phantom{0}$\pm$6.50} & 83.85{\Large\phantom{0}$\pm$3.25} & 85.36\\
Claude-4.6-Sonnet & \underline{96.47}{\Large\phantom{0}$\pm$4.16} & \underline{95.52}{\Large\phantom{0}$\pm$4.51} & \underline{98.49}{\Large\phantom{0}$\pm$3.29} & \underline{95.30}{\Large\phantom{0}$\pm$5.61} & 71.02{\Large$\pm$15.19} & \underline{94.33}{\Large\phantom{0}$\pm$5.25} & \underline{96.79}{\Large\phantom{0}$\pm$7.27} & \underline{97.29}{\Large\phantom{0}$\pm$4.67} & \underline{84.03}{\Large\phantom{0}$\pm$9.31} & \underline{90.17}{\Large\phantom{0}$\pm$5.07} & 62.07{\Large$\pm$35.76} & \underline{89.23}\\
Claude-4.6-Opus & \textbf{97.81}{\Large\phantom{0}$\pm$1.92} & \textbf{96.70}{\Large\phantom{0}$\pm$2.79} & \textbf{99.00}{\Large\phantom{0}$\pm$1.63} & \textbf{96.36}{\Large\phantom{0}$\pm$2.83} & \textbf{84.62}{\Large\phantom{0}$\pm$9.15} & \textbf{95.91}{\Large\phantom{0}$\pm$3.36} & \textbf{98.73}{\Large\phantom{0}$\pm$2.46} & \textbf{98.85}{\Large\phantom{0}$\pm$1.49} & \textbf{89.76}{\Large\phantom{0}$\pm$6.06} & \textbf{95.36}{\Large\phantom{0}$\pm$3.63} & \textbf{91.53}{\Large\phantom{0}$\pm$3.25} & \textbf{94.97} \\
\midrule
\rowcolor{gray!15} \textbf{Open-source} &  &  &  &  &  &  &  &  &  &  &  &  \\
\rowcolor{gray!15} \multicolumn{13}{l}{\hspace{1em}\textbf{$<$ 14B}} \\
Qwen3-4B-I & 15.31{\Large$\pm$10.34} & \phantom{0}7.25{\Large\phantom{0}$\pm$6.89} & 13.57{\Large$\pm$10.80} & 21.84{\Large$\pm$10.54} & 26.89{\Large$\pm$11.14} & 30.11{\Large$\pm$11.88} & 32.68{\Large$\pm$10.21} & 15.69{\Large$\pm$12.49} & 12.47{\Large\phantom{0}$\pm$9.73} & 39.00{\Large$\pm$18.49} & 24.76{\Large$\pm$17.68} & 21.77\\
\textbf{Qwen-4B~(EW-B)} & 43.30{\Large$\pm$11.99} & 36.44{\Large$\pm$11.02} & 35.73{\Large$\pm$10.17} & 35.66{\Large\phantom{0}$\pm$8.87} & 30.61{\Large$\pm$10.23} & 41.26{\Large\phantom{0}$\pm$8.01} & 28.30{\Large\phantom{0}$\pm$5.43} & 36.58{\Large\phantom{0}$\pm$9.19} & 26.90{\Large\phantom{0}$\pm$7.36} & 40.02{\Large\phantom{0}$\pm$9.04} & 56.04{\Large\phantom{0}$\pm$9.21} & 37.35 \\
\textbf{Qwen-4B~(EW-F)} & 44.68{\Large$\pm$10.21} & 37.24{\Large\phantom{0}$\pm$9.73} & 37.10{\Large\phantom{0}$\pm$8.45} & 35.24{\Large\phantom{0}$\pm$7.81} & 31.43{\Large$\pm$10.20} & 40.09{\Large\phantom{0}$\pm$6.84} & 28.15{\Large\phantom{0}$\pm$5.41} & 35.64{\Large\phantom{0}$\pm$9.28} & 28.30{\Large\phantom{0}$\pm$5.88} & 40.57{\Large\phantom{0}$\pm$9.72} & 59.12{\Large$\pm$11.15} & 37.96\\
Qwen2.5-7B-I & 15.77{\Large\phantom{0}$\pm$6.68} & \phantom{0}7.69{\Large\phantom{0}$\pm$5.06} & 11.70{\Large\phantom{0}$\pm$6.98} & 32.25{\Large\phantom{0}$\pm$7.42} & 25.07{\Large$\pm$10.49} & 30.45{\Large\phantom{0}$\pm$6.41} & 28.53{\Large\phantom{0}$\pm$6.75} & 11.80{\Large\phantom{0}$\pm$7.31} & \phantom{0}9.69{\Large\phantom{0}$\pm$5.53} & 36.20{\Large$\pm$16.06} & 20.14{\Large$\pm$12.72} & 20.84 \\
Qwen-7B~(Coser) & 16.59{\Large$\pm$11.97} & 12.20{\Large\phantom{0}$\pm$8.33} & 27.61{\Large$\pm$12.64} & 18.16{\Large$\pm$16.58} & 17.91{\Large$\pm$16.25} & 37.34{\Large\phantom{0}$\pm$8.69} & 25.20{\Large\phantom{0}$\pm$4.14} & 13.92{\Large\phantom{0}$\pm$6.96}& 27.60{\Large$\pm$12.86}& \phantom{0}2.06{\Large\phantom{0}$\pm$9.94} & 10.26{\Large\phantom{0}$\pm$1.55} & 18.98
\\
Qwen-7B~(Crab) & 14.25{\Large\phantom{0}$\pm$7.83} & \phantom{0}6.34{\Large\phantom{0}$\pm$4.43} & 15.76{\Large\phantom{0}$\pm$7.80} & 19.82{\Large\phantom{0}$\pm$7.22} & 15.89{\Large\phantom{0}$\pm$9.94} & 32.79{\Large$\pm$10.41} & 20.28{\Large\phantom{0}$\pm$5.85} & 16.91{\Large\phantom{0}$\pm$8.45} & 17.74{\Large\phantom{0}$\pm$5.31} &  26.82{\Large$\pm$12.05} & 14.91{\Large\phantom{0}$\pm$8.10} & 18.32
\\
\textbf{Qwen-7B~(EW-B)} & 50.87{\Large\phantom{0}$\pm$9.33} & \underline{45.47}{\Large\phantom{0}$\pm$8.89} & 43.31{\Large\phantom{0}$\pm$8.78} & 40.13{\Large\phantom{0}$\pm$6.47} & 37.89{\Large$\pm$11.31} & 43.00{\Large\phantom{0}$\pm$7.12} & 29.83{\Large\phantom{0}$\pm$5.03} & 46.09{\Large\phantom{0}$\pm$8.21} & 32.98{\Large\phantom{0}$\pm$6.45} & 42.05{\Large\phantom{0}$\pm$7.76} & 65.16{\Large\phantom{0}$\pm$9.52} & 43.34\\
\textbf{Qwen-7B~(EW-F)} & \textbf{54.07}{\Large\phantom{0}$\pm$9.85} & \textbf{48.19}{\Large$\pm$10.44} & \textbf{47.15}{\Large\phantom{0}$\pm$8.44} & 41.24{\Large\phantom{0}$\pm$7.72} & 40.33{\Large$\pm$10.06} & \underline{45.44}{\Large\phantom{0}$\pm$7.27} & 31.81{\Large\phantom{0}$\pm$6.14} & \textbf{48.61}{\Large\phantom{0}$\pm$9.92} & 36.09{\Large\phantom{0}$\pm$6.68} & 42.12{\Large\phantom{0}$\pm$8.31} & \textbf{65.78}{\Large\phantom{0}$\pm$9.45} & \underline{45.53} \\
Llama-3.1-8B-I & 30.58{\Large$\pm$11.10} & 22.30{\Large\phantom{0}$\pm$8.36} & 30.71{\Large$\pm$12.65} & 21.87{\Large$\pm$16.04} & 20.36{\Large$\pm$13.89} & 44.33{\Large\phantom{0}$\pm$8.88} & \textbf{46.19}{\Large\phantom{0}$\pm$9.30} & 40.78{\Large$\pm$12.36} & \textbf{40.77}{\Large\phantom{0}$\pm$9.04} & 27.26{\Large$\pm$22.46} & 14.55{\Large$\pm$17.16} & 30.88\\
Llama-8B~(Coser) & 30.36{\Large$\pm$16.76} & 22.26{\Large$\pm$15.53} & 38.28{\Large$\pm$18.63} & 15.86{\Large$\pm$16.21} & 15.94{\Large$\pm$15.91} & 37.61{\Large\phantom{0}$\pm$8.57} & 31.04{\Large\phantom{0}$\pm$2.85} & 30.36{\Large$\pm$14.55} & 37.27{\Large$\pm$11.12} &  \phantom{0}4.07{\Large$\pm$14.52} & 11.09{\Large\phantom{0}$\pm$4.60} & 24.92\\
Llama-8B~(Crab) & 29.37{\Large$\pm$18.52} & 20.00{\Large$\pm$16.45} & 38.10{\Large$\pm$20.05} & \phantom{0}1.26{\Large\phantom{0}$\pm$5.04} & \phantom{0}2.27{\Large\phantom{0}$\pm$8.62} & 33.20{\Large\phantom{0}$\pm$5.08} & 32.67{\Large\phantom{0}$\pm$2.56} & 26.57{\Large$\pm$12.04} & 34.03{\Large\phantom{0}$\pm$7.19} & 10.45{\Large$\pm$20.33} & 10.25{\Large\phantom{0}$\pm$1.88} & 21.65\\
\textbf{Llama-8B~(EW-B)} & \underline{51.23}{\Large$\pm$19.86} & 44.21{\Large$\pm$19.17} & 45.01{\Large$\pm$17.42} & \underline{42.15}{\Large$\pm$10.44} & \underline{42.09}{\Large$\pm$10.77} & 45.01{\Large$\pm$12.20} & 32.73{\Large\phantom{0}$\pm$7.00} & \underline{46.21}{\Large$\pm$20.25} & 37.88{\Large$\pm$14.49} & \underline{47.31}{\Large\phantom{0}$\pm$8.00} & 61.94{\Large$\pm$15.74} & 45.07 \\
\textbf{Llama-8B~(EW-F)} & 50.94{\Large$\pm$20.32} & 44.90{\Large$\pm$18.45} & \underline{46.43}{\Large$\pm$16.41} & \textbf{42.49}{\Large$\pm$10.11} & \textbf{42.44}{\Large$\pm$10.66} & \textbf{46.47}{\Large$\pm$11.92} & \underline{33.75}{\Large\phantom{0}$\pm$7.68} & 46.03{\Large$\pm$19.69} & \underline{38.83}{\Large$\pm$13.13} & \textbf{47.96}{\Large\phantom{0}$\pm$7.72} & \underline{65.62}{\Large\phantom{0}$\pm$8.79} & \textbf{45.99} \\
\addlinespace[0.3ex]
\midrule
\addlinespace[0.3ex]
\rowcolor{gray!15} \multicolumn{13}{l}{\hspace{1em}\textbf{$\geq$ 14B}} \\
Qwen2.5-14B-I & 31.44{\Large$\pm$10.64} & 14.71{\Large\phantom{0}$\pm$7.63} & 30.43{\Large$\pm$10.11} & 43.95{\Large\phantom{0}$\pm$6.51} & 35.56{\Large$\pm$11.66} & 43.84{\Large\phantom{0}$\pm$8.65} & 38.79{\Large\phantom{0}$\pm$9.18} & 32.83{\Large\phantom{0}$\pm$9.13} & 15.66{\Large\phantom{0}$\pm$6.22} & 53.36{\Large$\pm$11.16} & 39.78{\Large$\pm$18.50} & 34.58 \\
\textbf{Qwen-14B~(EW-B)} & 63.50{\Large$\pm$13.26} & 56.06{\Large$\pm$12.60} & 55.73{\Large$\pm$11.52} & 53.17{\Large\phantom{0}$\pm$8.10} & 56.68{\Large\phantom{0}$\pm$9.98} & 50.68{\Large\phantom{0}$\pm$8.45} & 35.47{\Large\phantom{0}$\pm$6.55} & 57.60{\Large$\pm$13.80} & 44.90{\Large\phantom{0}$\pm$8.74} & 48.39{\Large\phantom{0}$\pm$7.69} & \textbf{69.32}{\Large$\pm$10.20} & 53.77 \\
\textbf{Qwen-14B~(EW-F)} & 62.39{\Large$\pm$14.45} & 55.74{\Large$\pm$13.59} & 55.63{\Large$\pm$13.02} & 49.82{\Large\phantom{0}$\pm$8.48} & 53.60{\Large$\pm$11.17} & 50.03{\Large\phantom{0}$\pm$9.79} & 36.25{\Large\phantom{0}$\pm$7.47} & 55.36{\Large$\pm$14.26} & 44.04{\Large\phantom{0}$\pm$9.66} & 49.85{\Large\phantom{0}$\pm$7.98} & \underline{66.27}{\Large$\pm$14.59} & 52.63 \\
Qwen2.5-32B-I & 19.49{\Large$\pm$11.17} & \phantom{0}7.42{\Large\phantom{0}$\pm$6.16} & 20.08{\Large\phantom{0}$\pm$9.81} & 35.85{\Large\phantom{0}$\pm$9.25} & 29.82{\Large$\pm$11.19} & 39.50{\Large\phantom{0}$\pm$8.48} & 32.63{\Large\phantom{0}$\pm$8.40} & 24.55{\Large\phantom{0}$\pm$8.20} & 13.87{\Large\phantom{0}$\pm$6.75} & 45.12{\Large$\pm$12.65} & 38.14{\Large$\pm$15.37} & 27.86 \\
\textbf{Qwen-32B~(EW-B)} & \underline{67.80}{\Large$\pm$11.60} & \textbf{60.68}{\Large$\pm$12.84} & \underline{61.76}{\Large$\pm$10.93} & \textbf{57.95}{\Large\phantom{0}$\pm$7.89} & \textbf{59.52}{\Large\phantom{0}$\pm$9.81} & 53.70{\Large\phantom{0}$\pm$7.73} & 37.68{\Large\phantom{0}$\pm$6.59} & 61.25{\Large$\pm$12.24} & 47.75{\Large\phantom{0}$\pm$9.18} & 54.51{\Large\phantom{0}$\pm$8.48} & 65.39{\Large$\pm$19.91} & \underline{57.09} \\
\textbf{Qwen-32B~(EW-F)} & \textbf{67.81}{\Large$\pm$12.50} & \underline{60.56}{\Large$\pm$12.20} & 61.34{\Large$\pm$10.76} & \underline{57.24}{\Large\phantom{0}$\pm$7.83} & \underline{58.62}{\Large$\pm$10.76} & 55.43{\Large\phantom{0}$\pm$9.28} & 39.22{\Large\phantom{0}$\pm$7.91} & 60.55{\Large$\pm$13.42} & \underline{48.48}{\Large\phantom{0}$\pm$9.52} & 54.93{\Large\phantom{0}$\pm$7.91} & 63.45{\Large$\pm$22.36} & 57.06 \\
Qwen2.5-72B-I & 25.71{\Large$\pm$10.42} & \phantom{0}9.55{\Large\phantom{0}$\pm$5.89} & 30.88{\Large$\pm$10.99} & 37.77{\Large\phantom{0}$\pm$8.08} & 38.90{\Large\phantom{0}$\pm$6.64} & 46.31{\Large\phantom{0}$\pm$7.83} & 36.60{\Large\phantom{0}$\pm$7.27} & 36.50{\Large$\pm$14.50} & 23.44{\Large$\pm$10.29} & 43.86{\Large\phantom{0}$\pm$9.17} & 41.56{\Large$\pm$32.31} & 33.73 \\
Qwen3-32B & 42.16{\Large$\pm$13.59} & 26.76{\Large$\pm$12.91} & 58.39{\Large$\pm$10.81} & 51.10{\Large\phantom{0}$\pm$9.07} & 35.24{\Large$\pm$12.76} & 62.37{\Large$\pm$10.82} & 69.64{\Large$\pm$12.52} & \underline{65.99}{\Large\phantom{0}$\pm$9.24} & 47.34{\Large$\pm$11.69} & \underline{70.18}{\Large\phantom{0}$\pm$9.74} & 54.14{\Large$\pm$19.77} & 53.03 \\
Llama-3.1-70B-I & 32.95{\Large$\pm$10.28} & 14.42{\Large\phantom{0}$\pm$7.61} & 39.02{\Large$\pm$12.63} & 22.42{\Large$\pm$17.06} & 23.05{\Large$\pm$17.16} & 46.35{\Large\phantom{0}$\pm$9.10} & 37.78{\Large$\pm$10.11} & 48.64{\Large$\pm$12.57} & 26.93{\Large$\pm$10.26} & 48.42{\Large$\pm$11.14} & 38.27{\Large$\pm$29.68} & 34.39 \\
Llama-3.3-70B-I & 22.66{\Large\phantom{0}$\pm$8.93} & \phantom{0}6.23{\Large\phantom{0}$\pm$4.10} & 27.03{\Large$\pm$11.61} & 36.36{\Large$\pm$10.95} & 40.42{\Large\phantom{0}$\pm$5.32} & 46.90{\Large\phantom{0}$\pm$7.07} & 39.27{\Large\phantom{0}$\pm$8.92} & 39.83{\Large$\pm$11.62} & 22.21{\Large$\pm$12.14} & 52.85{\Large$\pm$11.61} & 46.33{\Large$\pm$28.49} & 34.55 \\
Mistral-Small & 54.19{\Large\phantom{0}$\pm$9.61} & 39.24{\Large\phantom{0}$\pm$9.77} & 57.23{\Large\phantom{0}$\pm$7.62} & 46.02{\Large\phantom{0}$\pm$5.48} & 44.53{\Large\phantom{0}$\pm$8.14} & \underline{68.46}{\Large\phantom{0}$\pm$6.27} & \underline{70.89}{\Large\phantom{0}$\pm$7.01} & 61.67{\Large\phantom{0}$\pm$6.76} & 40.48{\Large\phantom{0}$\pm$6.78} & 52.40{\Large\phantom{0}$\pm$9.10} & 63.85{\Large\phantom{0}$\pm$5.01} & 54.45
 \\
DeepSeek-V3-0324 & 66.22{\Large$\pm$13.05} & 56.23{\Large$\pm$12.75} & \textbf{75.59}{\Large\phantom{0}$\pm$8.32} & 55.73{\Large\phantom{0}$\pm$7.73} & 53.28{\Large$\pm$11.00} & \textbf{71.96}{\Large\phantom{0}$\pm$9.04} & \textbf{76.43}{\Large$\pm$10.32} & \textbf{72.48}{\Large\phantom{0}$\pm$8.38} & \textbf{51.89}{\Large$\pm$14.65} & \textbf{70.20}{\Large\phantom{0}$\pm$8.05} & 55.04{\Large$\pm$16.32} & \textbf{64.10} \\
\bottomrule
\end{tabular}
}
\vspace{-0.05in}
\caption{Full benchmark results for \textbf{Character Agent} evaluation. I and P denote Instruct and Preview in model names. The best performances within the same model scale are \textbf{bold-faced}, and the second-best are \underline{underlined}.}
\label{tab:app_character_results_full_balanced}
\vspace{-0.15in}
\end{table*}

\begin{table*}[t]
\centering
\huge
\resizebox{0.85\linewidth}{!}{
\begin{tabular}{lcccccccccc}
\toprule
\multicolumn{11}{l}{\textbf{SP}: Scene Planning \quad \textbf{SM}: Speaker Management \quad \textbf{WSM}: World State Maintenance \quad \textbf{IC}: Instruction Compliance} \\
\multicolumn{11}{l}{\textbf{CSR}: Cast Selection Rationality \quad \textbf{LSR}: Location \& Scenario Rationality \quad \textbf{SCC}: Scene Continuity \& Coherence } \\
\multicolumn{11}{l}{ \textbf{TSO}: Turn \& Scene Orchestration \quad \textbf{GUS}: Global Update Sensitivity \quad \textbf{GSA}: Global State Accuracy } \\
\multicolumn{11}{l}{\textbf{LUS}: Location Update Sensitivity \quad \textbf{LSA}: Location State Accuracy} \\
\midrule
\multirow{2}{*}{\textbf{Models}} & \multicolumn{3}{c}{\textbf{SP}} & \multicolumn{1}{c}{\textbf{SM}} & \multicolumn{4}{c}{\textbf{WSM}} & \multicolumn{1}{c}{\textbf{IC}} & \multirow{2}{*}{\textbf{Avg.}} \\
\cmidrule(lr){2-4} \cmidrule(lr){5-5} \cmidrule(lr){6-9} \cmidrule(lr){10-10}
 & \textbf{CSR} & \textbf{LSR} & \textbf{SCC} & \textbf{TSO} & \textbf{GUS} & \textbf{GSA} & \textbf{LUS} & \textbf{LSA} & \textbf{IC} & \\
\midrule
\rowcolor{gray!15} \textbf{Closed-source} &  &  &  &  &  &  &  &  &  &  \\
Kimi-K2.5 & 78.36{\Large\phantom{0}$\pm$7.78} & 80.61{\Large$\pm$15.01} & 51.72{\Large$\pm$25.95} & 58.84{\Large$\pm$10.95} & 66.64{\Large\phantom{0}$\pm$7.53} & \textbf{59.85}{\Large\phantom{0}$\pm$8.22} & 68.90{\Large\phantom{0}$\pm$6.87} & \underline{70.54}{\Large\phantom{0}$\pm$9.77} & 77.94{\Large\phantom{0}$\pm$9.94} & 68.16\\
Gemini-2.5-Flash & 73.18{\Large\phantom{0}$\pm$6.95} & 84.09{\Large\phantom{0}$\pm$6.10} & 66.36{\Large$\pm$14.98} & 43.02{\Large$\pm$24.56} & 60.41{\Large\phantom{0}$\pm$7.49} & 56.78{\Large\phantom{0}$\pm$4.10} & 54.22{\Large$\pm$11.27} & 56.25{\Large\phantom{0}$\pm$6.31} & 43.49{\Large$\pm$23.83} & 59.76\\
Gemini-2.5-Pro & 78.86{\Large\phantom{0}$\pm$4.73} & 88.33{\Large\phantom{0}$\pm$6.27} & \underline{70.34}{\Large$\pm$20.14} & 75.66{\Large\phantom{0}$\pm$5.50} & 65.89{\Large\phantom{0}$\pm$4.69} & 56.99{\Large\phantom{0}$\pm$5.84} & 54.96{\Large\phantom{0}$\pm$8.66} & 68.20{\Large\phantom{0}$\pm$9.76} & 80.36{\Large\phantom{0}$\pm$7.15} & 71.07\\
Gemini-3.1-Pro-P & \underline{83.50}{\Large\phantom{0}$\pm$3.17} & 91.60{\Large\phantom{0}$\pm$4.81} & 69.16{\Large$\pm$20.97} & 72.70{\Large$\pm$10.18} & 65.60{\Large\phantom{0}$\pm$9.89} & 55.13{\Large$\pm$10.68} & 66.51{\Large\phantom{0}$\pm$9.67} & 61.54{\Large$\pm$10.84} & 85.55{\Large$\pm$13.49} & 72.37\\
GPT-4o & 77.80{\Large\phantom{0}$\pm$8.81} & 85.33{\Large\phantom{0}$\pm$6.55} & 51.02{\Large$\pm$16.90} & 61.44{\Large\phantom{0}$\pm$7.34} & 61.40{\Large\phantom{0}$\pm$8.32} & 52.86{\Large\phantom{0}$\pm$5.91} & 54.12{\Large\phantom{0}$\pm$9.96} & 51.44{\Large\phantom{0}$\pm$7.07} & 73.50{\Large$\pm$18.44} & 63.21\\
GPT-5-Chat & 80.44{\Large\phantom{0}$\pm$6.88} & 88.30{\Large\phantom{0}$\pm$6.12} & 68.38{\Large$\pm$16.61} & 64.69{\Large$\pm$17.27} & 60.57{\Large\phantom{0}$\pm$9.14} & 53.97{\Large\phantom{0}$\pm$6.21} & 52.59{\Large\phantom{0}$\pm$7.83} & 57.85{\Large\phantom{0}$\pm$7.87} & 77.51{\Large$\pm$22.07} & 67.14\\
GPT-5.1-Chat & 79.61{\Large\phantom{0}$\pm$5.63} & 85.17{\Large\phantom{0}$\pm$7.68} & 42.52{\Large$\pm$19.01} & 50.54{\Large\phantom{0}$\pm$9.47} & \underline{68.69}{\Large\phantom{0}$\pm$2.01} & 53.83{\Large\phantom{0}$\pm$4.19} & \underline{69.82}{\Large\phantom{0}$\pm$3.48} & 53.06{\Large\phantom{0}$\pm$6.28} & 81.69{\Large\phantom{0}$\pm$4.83} & 64.99\\
GPT-5.3-Chat & 80.04{\Large\phantom{0}$\pm$5.09} & 89.35{\Large\phantom{0}$\pm$4.81} & 68.40{\Large$\pm$15.48} & 69.04{\Large\phantom{0}$\pm$6.41} & \textbf{70.02}{\Large\phantom{0}$\pm$2.12} & 51.11{\Large\phantom{0}$\pm$5.48} & \textbf{75.21}{\Large\phantom{0}$\pm$6.30} & 64.77{\Large$\pm$10.96} & \underline{85.57}{\Large\phantom{0}$\pm$4.92} & \underline{72.61}\\
Claude-4.6-Sonnet & 81.89{\Large\phantom{0}$\pm$5.46} & \underline{92.90}{\Large\phantom{0}$\pm$4.58} & 60.77{\Large$\pm$25.33} & \underline{79.84}{\Large$\pm$13.54} & 38.93{\Large$\pm$24.04} & 33.26{\Large$\pm$20.37} & 30.33{\Large$\pm$21.21} & 41.12{\Large$\pm$26.94} & 58.83{\Large$\pm$34.31} & 57.54 \\
Claude-4.6-Opus & \textbf{84.86}{\Large\phantom{0}$\pm$5.34} & \textbf{96.66}{\Large\phantom{0}$\pm$2.56} & \textbf{82.28}{\Large$\pm$13.37} & \textbf{84.98}{\Large\phantom{0}$\pm$4.76} & 66.05{\Large\phantom{0}$\pm$5.74} & \underline{58.48}{\Large\phantom{0}$\pm$6.75} & 61.72{\Large\phantom{0}$\pm$9.46} & \textbf{75.49}{\Large\phantom{0}$\pm$8.93} & \textbf{89.34}{\Large\phantom{0}$\pm$6.50} & \textbf{77.76}\\
\midrule
\rowcolor{gray!15} \textbf{Open-source} &  &  &  &  &  &  &  &  &  &  \\
\rowcolor{gray!15} \multicolumn{11}{l}{\hspace{1em}\textbf{$<$ 14B}} \\
Qwen3-4B-I & 60.20{\Large$\pm$15.88} & 52.50{\Large$\pm$23.75} & 20.71{\Large$\pm$15.06} & \phantom{0}5.65{\Large\phantom{0}$\pm$6.64} & 41.98{\Large$\pm$23.54} & 35.56{\Large$\pm$20.73} & 22.68{\Large$\pm$14.07} & 23.31{\Large$\pm$13.57} & 25.78{\Large$\pm$18.28} & 32.06 \\
\textbf{Qwen-4B~(EW-B)} & 61.06{\Large$\pm$10.75} & 60.69{\Large$\pm$12.10} & 22.63{\Large$\pm$17.51} & 29.73{\Large\phantom{0}$\pm$6.46} & 55.63{\Large\phantom{0}$\pm$7.21} & 44.11{\Large\phantom{0}$\pm$7.51} & 57.07{\Large\phantom{0}$\pm$6.05} & 46.06{\Large\phantom{0}$\pm$4.64} & 71.62{\Large$\pm$11.81} & 49.84 \\
\textbf{Qwen-4B~(EW-F)} & 61.78{\Large$\pm$10.20} & 62.42{\Large$\pm$11.14} & 23.20{\Large$\pm$16.20} & 30.81{\Large\phantom{0}$\pm$6.65} & 57.62{\Large\phantom{0}$\pm$7.97} & 44.15{\Large\phantom{0}$\pm$8.01} & 58.09{\Large\phantom{0}$\pm$7.59} & 46.55{\Large\phantom{0}$\pm$6.40} & 72.24{\Large$\pm$13.19} & 50.76\\
Qwen2.5-7B-I & 61.07{\Large$\pm$12.78} & 51.25{\Large$\pm$14.71} & 14.09{\Large$\pm$14.59} & \phantom{0}5.82{\Large\phantom{0}$\pm$4.92} & 53.53{\Large$\pm$11.52} & 44.53{\Large\phantom{0}$\pm$9.69} & 28.65{\Large\phantom{0}$\pm$9.43} & 31.53{\Large\phantom{0}$\pm$8.02} & 35.58{\Large$\pm$16.83} & 36.23\\
\textbf{Qwen-7B~(EW-B)} & 61.26{\Large$\pm$10.29} & 61.48{\Large\phantom{0}$\pm$9.78} & 29.34{\Large$\pm$17.50} & 37.26{\Large\phantom{0}$\pm$4.80} & \textbf{62.62}{\Large\phantom{0}$\pm$3.69} & \underline{47.21}{\Large\phantom{0}$\pm$5.58} & \textbf{61.38}{\Large\phantom{0}$\pm$4.10} & \underline{48.15}{\Large\phantom{0}$\pm$4.54} & \underline{75.67}{\Large\phantom{0}$\pm$9.70} & 53.82\\
\textbf{Qwen-7B~(EW-F)} & \textbf{63.90}{\Large\phantom{0}$\pm$8.89} & 63.32{\Large$\pm$11.66} & 32.02{\Large$\pm$17.41} & \textbf{39.98}{\Large\phantom{0}$\pm$6.79} & \underline{61.21}{\Large\phantom{0}$\pm$5.41} & \textbf{47.52}{\Large\phantom{0}$\pm$6.02} & \underline{61.20}{\Large\phantom{0}$\pm$4.10} & \textbf{49.10}{\Large\phantom{0}$\pm$4.44} & \textbf{76.59}{\Large\phantom{0}$\pm$9.78} & \textbf{54.98} \\
Llama-3.1-8B-I & 60.56{\Large$\pm$17.34} & \textbf{68.01}{\Large$\pm$17.31} & 34.16{\Large$\pm$13.44} & 28.65{\Large$\pm$11.65}  & 49.02{\Large$\pm$19.99} & 43.94{\Large$\pm$17.78} & 45.89{\Large$\pm$19.33} & 41.12{\Large$\pm$16.92} & 30.82{\Large$\pm$36.53} & 44.69\\
\textbf{Llama-8B~(EW-B)} & \underline{63.78}{\Large\phantom{0}$\pm$7.75} & 65.26{\Large$\pm$12.28} & \underline{36.07}{\Large$\pm$23.32} & 38.60{\Large$\pm$10.93} & 59.52{\Large\phantom{0}$\pm$5.01} & 44.23{\Large\phantom{0}$\pm$6.82} & 59.53{\Large\phantom{0}$\pm$5.59} & 46.34{\Large\phantom{0}$\pm$5.06} & 71.45{\Large$\pm$13.73} & 53.86\\
\textbf{Llama-8B~(EW-F)} & 63.58{\Large\phantom{0}$\pm$7.91} & \underline{67.57}{\Large$\pm$11.51} & \textbf{38.41}{\Large$\pm$23.29} & \underline{39.79}{\Large$\pm$11.31} & 56.59{\Large\phantom{0}$\pm$7.62} & 42.89{\Large\phantom{0}$\pm$7.78} & 60.19{\Large\phantom{0}$\pm$5.80} & 46.75{\Large\phantom{0}$\pm$4.99} & 72.90{\Large\phantom{0}$\pm$7.92} & \underline{54.30}\\
\addlinespace[0.3ex]
\midrule
\addlinespace[0.3ex]
\rowcolor{gray!15} \multicolumn{11}{l}{\hspace{1em}\textbf{$\geq$ 14B}} \\
Qwen2.5-14B-I & 70.66{\Large$\pm$10.08} & 67.37{\Large$\pm$13.98} & 21.47{\Large$\pm$15.64} & 16.95{\Large\phantom{0}$\pm$6.74} & 50.64{\Large$\pm$21.59} & 41.28{\Large$\pm$17.15} & 23.71{\Large$\pm$12.53} & 29.82{\Large$\pm$13.41} & 41.89{\Large$\pm$18.78} & 40.42 \\
\textbf{Qwen-14B~(EW-B)} & 68.55{\Large\phantom{0}$\pm$8.31} & 73.50{\Large\phantom{0}$\pm$8.80} & \underline{46.14}{\Large$\pm$16.49} & 46.56{\Large$\pm$10.09} & 61.12{\Large\phantom{0}$\pm$5.35} & 48.62{\Large\phantom{0}$\pm$5.52} & \textbf{62.13}{\Large\phantom{0}$\pm$4.10} & 49.73{\Large\phantom{0}$\pm$4.68} & \textbf{79.89}{\Large$\pm$10.29} & 59.58\\
\textbf{Qwen-14B~(EW-F)} & 68.18{\Large\phantom{0}$\pm$7.96} & 70.67{\Large$\pm$10.66} & 39.24{\Large$\pm$18.05} & 45.19{\Large$\pm$10.48} & 59.45{\Large\phantom{0}$\pm$7.84} & 46.76{\Large\phantom{0}$\pm$6.86} & \underline{60.84}{\Large\phantom{0}$\pm$6.83} & 49.08{\Large\phantom{0}$\pm$6.37} & \underline{77.02}{\Large$\pm$15.15} & 57.38\\
Qwen2.5-32B-I & 69.55{\Large\phantom{0}$\pm$9.80} & 60.18{\Large$\pm$14.46} & 17.69{\Large$\pm$14.67} & 18.03{\Large\phantom{0}$\pm$7.14} & 61.21{\Large\phantom{0}$\pm$6.80} & 49.53{\Large\phantom{0}$\pm$7.00} & 42.35{\Large$\pm$13.03} & 42.08{\Large\phantom{0}$\pm$7.58} & 51.12{\Large$\pm$17.99} & 45.75\\
\textbf{Qwen-32B~(EW-B)} & 69.58{\Large\phantom{0}$\pm$8.40} & 78.03{\Large\phantom{0}$\pm$8.04} & 43.52{\Large$\pm$18.12} & \underline{47.21}{\Large\phantom{0}$\pm$8.56} & 61.60{\Large\phantom{0}$\pm$3.31} & \underline{49.68}{\Large\phantom{0}$\pm$5.28} & 60.62{\Large\phantom{0}$\pm$4.36} & \underline{50.89}{\Large\phantom{0}$\pm$4.36} & 75.47{\Large$\pm$22.25} & \underline{59.62}
\\
\textbf{Qwen-32B~(EW-F)} & 69.77{\Large\phantom{0}$\pm$7.99} & 77.26{\Large\phantom{0}$\pm$9.58} & 45.42{\Large$\pm$17.69} & \textbf{48.21}{\Large\phantom{0}$\pm$9.83} & \underline{61.67}{\Large\phantom{0}$\pm$3.38} & \textbf{49.92}{\Large\phantom{0}$\pm$5.06} & 60.65{\Large\phantom{0}$\pm$4.40} & \textbf{51.95}{\Large\phantom{0}$\pm$3.96} & 73.94{\Large$\pm$24.58} & \textbf{59.87}\\
Qwen2.5-72B-I & 70.29{\Large\phantom{0}$\pm$9.27} & 69.52{\Large\phantom{0}$\pm$8.44} & 37.89{\Large$\pm$13.99} & 29.88{\Large$\pm$11.59} & 39.85{\Large$\pm$26.68} & 31.12{\Large$\pm$20.86} & 27.04{\Large$\pm$19.54} & 29.36{\Large$\pm$20.08} & 37.35{\Large$\pm$29.99} & 41.37\\
Qwen3-32B & 67.59{\Large\phantom{0}$\pm$9.82} & 71.76{\Large$\pm$12.88} & 34.70{\Large$\pm$20.55} & 43.60{\Large\phantom{0}$\pm$9.59} & 54.41{\Large\phantom{0}$\pm$8.87} & 47.33{\Large\phantom{0}$\pm$7.26} & 53.52{\Large\phantom{0}$\pm$9.16} & 49.02{\Large\phantom{0}$\pm$9.48} & 51.34{\Large$\pm$20.40} & 52.59\\
Llama-3.1-70B-I & 63.89{\Large$\pm$13.92} & 72.24{\Large$\pm$12.37} & 30.85{\Large$\pm$15.45} & 28.53{\Large$\pm$10.52} & 50.82{\Large$\pm$18.89} & 40.32{\Large$\pm$15.07} & 21.66{\Large\phantom{0}$\pm$9.96} & 35.75{\Large$\pm$13.93} & 31.37{\Large$\pm$25.21} & 41.71\\
Llama-3.3-70B-I & \underline{75.62}{\Large$\pm$10.65} & \underline{78.39}{\Large$\pm$12.61} & 36.76{\Large$\pm$14.19} & 23.47{\Large\phantom{0}$\pm$9.18} & 41.47{\Large$\pm$22.18} & 34.41{\Large$\pm$18.74} & 22.41{\Large$\pm$15.41} & 27.44{\Large$\pm$15.72} & 34.90{\Large$\pm$22.87} & 41.65\\
Mistral-Small & 72.24{\Large\phantom{0}$\pm$6.72} & 73.74{\Large\phantom{0}$\pm$6.86} & 41.38{\Large$\pm$16.00} & 42.39{\Large\phantom{0}$\pm$6.98} & 60.88{\Large\phantom{0}$\pm$7.47} & 44.55{\Large\phantom{0}$\pm$6.82} & 41.64{\Large\phantom{0}$\pm$8.90} & 44.92{\Large\phantom{0}$\pm$6.53} & 60.41{\Large\phantom{0}$\pm$8.50} & 53.57\\
DeepSeek-V3-0324 & \textbf{76.98}{\Large\phantom{0}$\pm$6.36} & \textbf{81.86}{\Large\phantom{0}$\pm$8.43} & \textbf{51.00}{\Large$\pm$19.30} & 38.75{\Large$\pm$15.05} & \textbf{63.48}{\Large\phantom{0}$\pm$5.11} & 48.79{\Large\phantom{0}$\pm$6.70} & 47.32{\Large$\pm$10.08} & 49.70{\Large\phantom{0}$\pm$8.14} & 60.33{\Large$\pm$16.47} & 57.58
\\
\bottomrule
\end{tabular}
}
\caption{Full benchmark results for \textbf{World Model} evaluation. I and P denote Instruct and Preview in model names. The best performances within the same model scale are \textbf{bold-faced}, and the second-best are \underline{underlined}.}
\label{tab:app_world_results_full_balanced}
\vspace{-0.15in}
\end{table*}

\subsubsection{Dim III: Environmental Grounding}

Evaluates whether the character truly ``lives'' in the current scene rather than conversing in a vacuum, reflecting the constraining effect of the World Model on Character Agent behavior.

\noindent\textbf{Environment Awareness (EA).}
Whether the character's behavior is constrained by the current environment (global world state + location states).

\noindent\textit{Non-Environment Character Agent}:
(a)~\textit{Global Awareness}: reacts reasonably to the global state (e.g., tension during wartime, conservation during resource scarcity);
(b)~\textit{Location Awareness}: notices the current state of the location (e.g., a shopkeeper mentioning ``last night's storm blew the roof off'');
(c)~\textit{State Change Response}: when world state changes between scenes, the character notices and adjusts accordingly.

\noindent\textit{Environment Character Agent}:
(a)~\textit{Global State Consistency}: environmental descriptions are consistent with the current global state;
(b)~\textit{Location State Accuracy}: environmental descriptions accurately reflect the location card's current state (e.g., damaged buildings not described as intact);
(c)~\textit{State Change Presentation}: when world state changes between scenes, environmental descriptions reflect these transitions.

\noindent\textbf{Environmental Utilization (EU).}
Whether characters appropriately and meaningfully use environmental elements when relevant to the scene.

\noindent\textit{Non-Environment Character Agent}:
(a)~\textit{Environmental Sensory Details}: sensory descriptions convey the character's perception of their environment rather than generic descriptions;
(b)~\textit{Prop Interaction}: items and entities in the location are used to advance the plot;
(c)~\textit{Atmosphere Building}: environmental atmosphere enhances immersion rather than conversing in a ``blank room.''

\noindent\textit{Environment Character Agent}:
(a)~\textit{Multi-Sensory Richness}: environmental descriptions engage multiple sensory dimensions (visual, auditory, etc.);
(b)~\textit{Scene Element Usage}: descriptions utilize items and entities in the location;
(c)~\textit{Atmosphere-Narrative Alignment}: atmosphere matches the current narrative pace and emotional tone.

\subsubsection{Dim IV: Interaction Quality}

Evaluates whether the character truly ``listens'' and ``responds'' to other characters, and whether interactions drive narrative development.

\noindent\textbf{Contextual Responsiveness (CR).}
Whether responses closely follow context, and whether inter-character attitudes match relationship settings and adjust as profiles evolve.
Criteria: (a)~\textit{Information Continuity}: not ignoring key information or questions, not abruptly changing topics;
(b)~\textit{Logical Continuity}: reacting reasonably to others' actions (e.g., accepting/refusing a handed item rather than ignoring it);
(c)~\textit{Relationship Matching}: clear distinctions in tone and trust toward allies, enemies, and strangers, dynamically adjusting with the plot.

\noindent\textbf{Narrative Progression (NP).}
Whether interactions advance the narrative and whether previously planted foreshadowing is followed up on.
Criteria: (a)~\textit{Information Increment}: each round provides new information, actions, or emotional developments rather than repeating known content;
(b)~\textit{Suspense and Hooks}: suspense created through silence, conflict, or hints, leaving hooks for subsequent interactions;
(c)~\textit{Foreshadowing Payoff}: foreshadowing from previous scenes is noticed and followed up at appropriate moments.

\subsubsection{Dim V: Motivation Generation}

\noindent\textbf{Motivation Quality (MQ).}
Whether the generated scene motivation matches the current profile, world state, and scene settings, and is specific and actionable.
Criteria: (a)~\textit{Profile Alignment}: motivation aligns with the character's current personality, goals, and relationship status;
(b)~\textit{Situational Fit}: motivation considers the current world state and scene settings (e.g., no leisure motivations in dangerous scenarios);
(c)~\textit{Actionability}: motivation is specific enough to guide behavior in the scene rather than being vague and abstract.

\subsubsection{Dim VI: Instruction Compliance}

\noindent\textbf{Instruction Compliance (IC).}
Whether the character strictly plays itself without overstepping and the output format is standardized.
Criteria: (a)~\textit{No Overstepping}: strictly outputting only one's own character content, not speaking or acting for others;
(b)~\textit{Format Compliance}: correct usage of thought/action/speech, output structure meets requirements;
(c)~\textit{Length Control}: reasonable output length, neither excessively verbose nor overly brief.

\noindent\textit{Error Penalty.} When a simulation terminates prematurely due to the Character Agent failing to produce valid output, both the IC Penalty and the Metric Penalty described in \S\ref{sec:eval_framework} are applied.

\subsection{World Model Evaluation (WORLD Score)}

\subsubsection{Dim I: Scene Planning}

Evaluates whether the World Model's scene planning~\citep{liu2026planbench, liu2026adaplanbench} is reasonable, including character selection, location/scenario generation, and cross-scene coherence.

\noindent\textbf{Cast Selection Rationality (CSR).}
Whether the selected participating characters match the current narrative state and character goals.
Criteria: (a)~\textit{Narrative-Driven}: appearing characters serve the current narrative needs (e.g., opposing characters appear together in conflict scenes);
(b)~\textit{Goal Relevance}: selected characters are directly related to the current narrative thread;
(c)~\textit{Avoid Redundancy}: no characters unrelated to the current narrative are introduced;
(d)~\textit{No Missing Key Characters}: characters in the pool who should appear are not overlooked.

\noindent\textbf{Location \& Scenario Rationality (LSR).}
Whether the chosen location and generated scenario are appropriate for the selected cast and current narrative state.
Criteria: (a)~\textit{Location Appropriateness}: the location is a plausible place for the selected characters to meet, serving narrative needs;
(b)~\textit{Scenario Quality}: the scenario provides a clear dramatic setup that is specific and actionable;
(c)~\textit{Continuity with Previous Scene}: location/scenario follows naturally from the previous scene's events;
(d)~\textit{Character-Setting Fit}: the setting is appropriate for the selected characters.

\begin{table*}[t]
\vspace{-0.05in}
\centering
\huge
\resizebox{\linewidth}{!}{
\begin{tabular}{llccccccccccc}
\toprule
\multicolumn{13}{l}{\textbf{CC}: Character Consistency \quad \textbf{EQ}: Evolution Quality \quad \textbf{EG}: Environmental Grounding \quad \textbf{IQ}: Interaction Quality \quad \textbf{MG}: Motivation Generation} \\
\multicolumn{13}{l}{\textbf{PF}: Profile Fidelity \quad \textbf{SSF}: Speaking Style Fidelity \quad \textbf{MDB}: Motivation-Driven Behavior \quad \textbf{PUF}: Profile Update Fidelity \quad \textbf{PES}: Profile Evolution Smoothness } \\
\multicolumn{13}{l}{\textbf{EA}: Environment Awareness \quad \textbf{EU}: Environmental Utilization \quad \textbf{CR}: Contextual Responsiveness \quad \textbf{NP}: Narrative Progression \quad \textbf{MQ}: Motivation Quality} \\
\midrule
\multirow{2}{*}{\textbf{Models}} & \multirow{2}{*}{\textbf{Setting}} & \multicolumn{3}{c}{\textbf{CC}} & \multicolumn{2}{c}{\textbf{EQ}} & \multicolumn{2}{c}{\textbf{EG}} & \multicolumn{2}{c}{\textbf{IQ}} & \multicolumn{1}{c}{\textbf{MG}} & \multirow{2}{*}{\textbf{Avg.}} \\
\cmidrule(lr){3-5} \cmidrule(lr){6-7} \cmidrule(lr){8-9} \cmidrule(lr){10-11} \cmidrule(lr){12-12} 
 & & \textbf{PF} & \textbf{SSF} & \textbf{MDB} & \textbf{PUF} & \textbf{PES} & \textbf{EA} & \textbf{EU} & \textbf{CR} & \textbf{NP} & \textbf{MQ} & \\
\midrule
\multirow{4}{*}{GPT-5.3-Chat} & Full & 94.71{\Large\phantom{0}$\pm$2.75} & 88.77{\Large\phantom{0}$\pm$4.30} & 93.71{\Large\phantom{0}$\pm$2.95} & 77.90{\Large\phantom{0}$\pm$3.74} & 80.16{\Large\phantom{0}$\pm$6.62} & 91.62{\Large\phantom{0}$\pm$4.88} & 92.10{\Large\phantom{0}$\pm$7.54} & 92.27{\Large\phantom{0}$\pm$3.74} & 59.98{\Large\phantom{0}$\pm$7.55} & 83.94{\Large\phantom{0}$\pm$6.50} & 85.52\\
 & w/o Char State & 88.48{\Large\phantom{0}$\pm$5.86} & 86.08{\Large\phantom{0}$\pm$5.92} & 86.54{\Large\phantom{0}$\pm$6.60} & 27.18{\Large\phantom{0}$\pm$8.88} & 25.54{\Large$\pm$10.35} & 90.92{\Large\phantom{0}$\pm$4.75} & 91.94{\Large\phantom{0}$\pm$7.38} & 89.89{\Large\phantom{0}$\pm$4.52} & 53.60{\Large\phantom{0}$\pm$8.15} & 57.08{\Large$\pm$12.97} & 69.73 \\
 & w/o Both & 86.38{\Large\phantom{0}$\pm$5.77} & 82.57{\Large\phantom{0}$\pm$7.18} & 83.63{\Large\phantom{0}$\pm$6.86} & 24.96{\Large$\pm$10.45} & 25.36{\Large\phantom{0}$\pm$9.88} & 86.39{\Large\phantom{0}$\pm$7.18} & 84.29{\Large\phantom{0}$\pm$9.84} & 86.14{\Large\phantom{0}$\pm$5.61} & 52.61{\Large\phantom{0}$\pm$7.36} & 55.46{\Large$\pm$12.57} & 66.78\\
\midrule
\multirow{4}{*}{Llama-3.1-8B-I} & Full & 30.58{\Large$\pm$11.10} & 22.30{\Large\phantom{0}$\pm$8.36} & 30.71{\Large$\pm$12.65} & 21.87{\Large$\pm$16.04} & 20.36{\Large$\pm$13.89} & 44.33{\Large\phantom{0}$\pm$8.88} & 46.19{\Large\phantom{0}$\pm$9.30} & 40.78{\Large$\pm$12.36} & 40.77{\Large\phantom{0}$\pm$9.04} & 27.26{\Large$\pm$22.46} & 32.52\\
 & w/o Char State & 12.31{\Large$\pm$13.52} & 11.23{\Large$\pm$10.97} & 15.37{\Large\phantom{0}$\pm$8.40} & \phantom{0}6.38{\Large$\pm$12.87}  & 13.61{\Large\phantom{0}$\pm$8.99} & 32.64{\Large$\pm$11.83} & 39.45{\Large\phantom{0}$\pm$6.17} & 19.22{\Large$\pm$11.23} & 21.09{\Large\phantom{0}$\pm$7.72} & 23.21{\Large$\pm$14.15} & 19.45 \\
 & w/o Both & \phantom{0}5.78{\Large\phantom{0}$\pm$5.94} & \phantom{0}4.83{\Large\phantom{0}$\pm$5.03} & \phantom{0}7.29{\Large\phantom{0}$\pm$5.75} & \phantom{0}5.68{\Large\phantom{0}$\pm$3.58} & 7.01{\Large$\pm$12.21} & 23.08{\Large\phantom{0}$\pm$7.68} & 30.55{\Large\phantom{0}$\pm$7.35} & 11.93{\Large\phantom{0}$\pm$7.26} & 12.12{\Large\phantom{0}$\pm$6.44} & 20.64{\Large$\pm$24.23} & 12.89 \\
\bottomrule
\end{tabular}
}
\vspace{-0.05in}
\caption{Ablation study on \textbf{Character Agent} performance. ``Full'' denotes the complete \ours{} framework; ``w/o Char State'' removes character state updates; ``w/o Both'' removes both world and character state updates.}
\label{tab:ablation_character}
\end{table*}

\begin{table*}[t]
\centering
\huge
\resizebox{0.8\linewidth}{!}{
\begin{tabular}{llccccccccc}
\toprule
\multicolumn{11}{l}{\textbf{SP}: Scene Planning \quad \textbf{SM}: Speaker Management \quad \textbf{WSM}: World State Maintenance \quad \textbf{CSR}: Cast Selection Rationality } \\
\multicolumn{11}{l}{\textbf{LSR}: Location \& Scenario Rationality \quad \textbf{SCC}: Scene Continuity \& Coherence \quad \textbf{TSO}: Turn \& Scene Orchestration } \\
\multicolumn{11}{l}{\textbf{GUS}: Global Update Sensitivity \quad \textbf{GSA}: Global State Accuracy \quad \textbf{LUS}: Location Update Sensitivity \quad \textbf{LSA}: Location State Accuracy} \\
\midrule
\multirow{2}{*}{\textbf{Models}} & \multirow{2}{*}{\textbf{Setting}} & \multicolumn{3}{c}{\textbf{SP}} & \multicolumn{1}{c}{\textbf{SM}} & \multicolumn{4}{c}{\textbf{WSM}} & \multirow{2}{*}{\textbf{Avg.}} \\
\cmidrule(lr){3-5} \cmidrule(lr){6-6} \cmidrule(lr){7-10} 
 & & \textbf{CSR} & \textbf{LSR} & \textbf{SCC} & \textbf{TSO} & \textbf{GUS} & \textbf{GSA} & \textbf{LUS} & \textbf{LSA}& \\
\midrule
\multirow{3}{*}{GPT-5.3-Chat} & Full & 80.04{\Large\phantom{0}$\pm$5.09} & 89.35{\Large\phantom{0}$\pm$4.81} & 68.40{\Large$\pm$15.48} & 69.04{\Large\phantom{0}$\pm$6.41} & 70.02{\Large\phantom{0}$\pm$2.12} & 51.11{\Large\phantom{0}$\pm$5.48} & 75.21{\Large\phantom{0}$\pm$6.30} & 64.77{\Large$\pm$10.96} & 70.99 \\
 & w/o World State & 78.31{\Large\phantom{0}$\pm$4.42} & 87.63{\Large\phantom{0}$\pm$6.28} & 58.88{\Large$\pm$16.92} & 58.65{\Large\phantom{0}$\pm$7.10} & 43.48{\Large$\pm$16.93} & 51.87{\Large\phantom{0}$\pm$6.00} & 64.70{\Large\phantom{0}$\pm$7.35} & 44.47{\Large\phantom{0}$\pm$8.62} & 61.00 \\
 & w/o Both & 79.65{\Large\phantom{0}$\pm$4.93} & 87.81{\Large\phantom{0}$\pm$8.68} & 52.94{\Large$\pm$17.52} & 56.18{\Large\phantom{0}$\pm$6.84} & 42.20{\Large$\pm$20.13} & 51.48{\Large\phantom{0}$\pm$7.21} & 64.60{\Large\phantom{0}$\pm$7.01} & 43.10{\Large\phantom{0}$\pm$8.32} & 59.75 \\
\midrule
\multirow{3}{*}{Llama-3.1-8B-I} & Full & 60.56{\Large$\pm$17.34} & 68.01{\Large$\pm$17.31} & 34.16{\Large$\pm$13.44} & 28.65{\Large$\pm$11.65} & 49.02{\Large$\pm$19.99} & 43.94{\Large$\pm$17.78} & 45.89{\Large$\pm$19.33} & 41.12{\Large$\pm$16.92} & 46.42 \\
 & w/o World State & 51.06{\Large$\pm$16.65} & 59.21{\Large$\pm$14.29} & 27.02{\Large$\pm$12.89} & 5.79{\Large\phantom{0}$\pm$5.81} & 18.82{\Large$\pm$15.83} & 20.04{\Large$\pm$15.71} & 29.80{\Large$\pm$18.99} & 15.27{\Large$\pm$20.79} & 28.38 \\
 & w/o Both & 41.94{\Large$\pm$11.16} & 43.51{\Large$\pm$12.99} & \phantom{0}7.73{\Large$\pm$11.71} & \phantom{0}4.10{\Large\phantom{0}$\pm$2.84} & 20.31{\Large$\pm$15.33} & 20.72{\Large$\pm$11.15} & 28.45{\Large$\pm$20.52} & 17.57{\Large$\pm$14.14} & 23.04
 \\
\bottomrule
\end{tabular}
}
\vspace{-0.05in}
\caption{Ablation study on \textbf{World Model} performance. ``Full'' denotes the complete \ours{} framework; ``w/o World State'' removes world state updates; ``w/o Both'' removes both world and character state updates.}
\vspace{-0.1in}
\label{tab:ablation_world}
\end{table*}

\begin{table}[t]
\centering
\large
\resizebox{\columnwidth}{!}{
\begin{tabular}{lccc}
\toprule
\textbf{Setting} & \textbf{PUF} & \textbf{PES} & \textbf{Avg.} \\
\midrule
Full & 77.90{\small$\pm$3.74} & 80.16{\small$\pm$6.62} & 79.03 \\
w/o Hidden Tracker & 65.30{\small$\pm$5.92} & 77.56{\small$\pm$4.91} & 71.43 \\
w/o Open Schema & 76.65{\small$\pm$2.94} & 78.91{\small$\pm$4.42} & 77.78 \\
\bottomrule
\end{tabular}}
\vspace{-0.05in}
\caption{Character-side component ablations with \textit{GPT-5.3-Chat}. Removing the Hidden Tracker or replacing the open profile with a fixed schema both degrade profile-update quality.}
\label{tab:ablation_char_component}
\end{table}

\begin{table}[t]
\centering
\large
\resizebox{\columnwidth}{!}{
\begin{tabular}{lccccc}
\toprule
\textbf{Setting} & \textbf{CSR} & \textbf{LSR} & \textbf{GUS} & \textbf{GSA} & \textbf{Avg.} \\
\midrule
Full & 80.04{\small$\pm$5.09} & 89.35{\small$\pm$4.81} & 70.02{\small$\pm$2.12} & 51.11{\small$\pm$5.48} & 72.63 \\
w/o Open Schema & 79.10{\small$\pm$3.44} & 88.55{\small$\pm$2.53} & 66.50{\small$\pm$1.89} & 50.95{\small$\pm$2.15} & 71.28 \\
\bottomrule
\end{tabular}}
\vspace{-0.05in}
\caption{World-side schema ablation with \textit{GPT-5.3-Chat}. Replacing the open global state with a fixed schema reduces global-state maintenance.}
\vspace{-0.15in}
\label{tab:ablation_global_schema}
\end{table}

\begin{table}[t]
\centering
\huge
\resizebox{\columnwidth}{!}{
\begin{tabular}{llccc}
\toprule
\multirow{2}{*}{\textbf{Evaluation Target}} & \multirow{2}{*}{\textbf{Model}} & \textbf{Untrained} & \multicolumn{2}{c}{\textbf{Trained}} \\
\cmidrule(lr){3-3} \cmidrule(lr){4-5}
& & \textbf{Avg.} & \textbf{ID Avg.} & \textbf{OOD Avg.} \\
\midrule
\multirow{5}{*}{Character Agent}
& Qwen3-4B-I & 21.77 & 38.80 & 36.95 \\
& Qwen2.5-7B-I & 20.84 & 45.28 & 45.88 \\
& Llama-3.1-8B-I & 30.88 & 49.62 & 41.08 \\
& Qwen2.5-14B-I & 34.58 & 52.57 & 52.73 \\
& Qwen2.5-32B-I & 27.86 & 56.63 & 57.61 \\
\midrule
\multirow{5}{*}{World Model}
& Qwen3-4B-I & 32.06 & 52.36 & 48.84 \\
& Qwen2.5-7B-I & 36.23 & 55.44 & 54.23 \\
& Llama-3.1-8B-I & 44.69 & 56.73 & 50.98 \\
& Qwen2.5-14B-I & 40.42 & 58.44 & 55.91 \\
& Qwen2.5-32B-I & 45.75 & 60.65 & 58.96 \\
\bottomrule
\end{tabular}}
\caption{ID and OOD performance after full-mixture training on Character Agent and World Model. Untrained averages are included as references.}
\vspace{-0.2in}
\label{tab:id_ood_results}
\end{table}

\noindent\textbf{Scene Continuity \& Coherence (SCC).}
Whether cross-scene planning forms a coherent narrative arc (cross-scene metric evaluated over the full trajectory).
Criteria: (a)~\textit{Narrative Arc}: consecutive scenes form a directional narrative progression rather than random assembly;
(b)~\textit{Scene Transitions}: location choices and scene descriptions naturally connect with the previous scene;
(c)~\textit{Pacing Control}: overall narrative pacing is reasonable, important plot points receive sufficient development;
(d)~\textit{Thread Management}: narrative threads are introduced, developed, and resolved; no plot lines are abandoned.

\subsubsection{Dim II: Speaker Management}

\noindent\textbf{Turn \& Scene Orchestration (TSO).}
Whether speaker selection, environmental description timing, and scene ending timing are appropriate.
Criteria: (a)~\textit{Speaker Selection}: \texttt{next\_character} selects the character who should most appropriately respond, not random rotation;
(b)~\textit{Environmental Description Timing}: environmental descriptions introduced at appropriate moments (scene changes, important events);
(c)~\textit{Group Character Actions}: in multi-character scenes, character combinations are reasonably selected for joint interactions, fitting the current situation and relationships;
(d)~\textit{Character Coverage Balance}: core characters receive participation proportional to their narrative importance; transient characters naturally fade out after fulfilling their narrative function;
(e)~\textit{Ending Timing}: the scene ends at a natural narrative juncture, not abruptly during a climax or dragging when nothing happens.

\begin{table*}[t]
\centering
\huge
\resizebox{\linewidth}{!}{
\begin{tabular}{lcccccccccccc}
\toprule
\multicolumn{13}{l}{\textbf{CC}: Character Consistency \quad \textbf{EQ}: Evolution Quality \quad \textbf{EG}: Environmental Grounding \quad \textbf{IQ}: Interaction Quality \quad \textbf{MG}: Motivation Generation} \\
\multicolumn{13}{l}{\textbf{IC}: Instruction Compliance \quad \textbf{PF}: Profile Fidelity \quad \textbf{SSF}: Speaking Style Fidelity \quad \textbf{MDB}: Motivation-Driven Behavior \quad \textbf{PUF}: Profile Update Fidelity} \\
\multicolumn{13}{l}{\textbf{PES}: Profile Evolution Smoothness \quad \textbf{EA}: Environment Awareness \quad \textbf{EU}: Environmental Utilization} \\
\multicolumn{13}{l}{\textbf{CR}: Contextual Responsiveness \quad \textbf{NP}: Narrative Progression \quad \textbf{MQ}: Motivation Quality} \\
\midrule
\multirow{2}{*}{\textbf{Models}} & \multicolumn{3}{c}{\textbf{CC}} & \multicolumn{2}{c}{\textbf{EQ}} & \multicolumn{2}{c}{\textbf{EG}} & \multicolumn{2}{c}{\textbf{IQ}} & \multicolumn{1}{c}{\textbf{MG}} & \multicolumn{1}{c}{\textbf{IC}} & \multirow{2}{*}{\textbf{Avg.}} \\
\cmidrule(lr){2-4} \cmidrule(lr){5-6} \cmidrule(lr){7-8} \cmidrule(lr){9-10} \cmidrule(lr){11-11} \cmidrule(lr){12-12}
 & \textbf{PF} & \textbf{SSF} & \textbf{MDB} & \textbf{PUF} & \textbf{PES} & \textbf{EA} & \textbf{EU} & \textbf{CR} & \textbf{NP} & \textbf{MQ} & \textbf{IC} & \\
\midrule
\rowcolor{gray!15} \multicolumn{13}{l}{\textbf{Claude-4.6-Sonnet Judge}} \\
\bottomrule
\end{tabular}
}
\vspace{-0.05in}
\caption{Full \textbf{Character Agent} benchmark results under three judge models. Gray rows indicate the judge model; all score rows are copied from the original per-judge tables, which are retained separately.}
\label{tab:character_results_all_judges_full}
\vspace{-0.2in}
\end{table*}

\begin{table*}[t]
\centering
\large
\resizebox{\linewidth}{!}{
\begin{tabular}{lcccccccccc}
\toprule
\multicolumn{11}{l}{\textbf{SP}: Scene Planning \quad \textbf{SM}: Speaker Management \quad \textbf{WSM}: World State Maintenance \quad \textbf{IC}: Instruction Compliance} \\
\multicolumn{11}{l}{\textbf{CSR}: Cast Selection Rationality \quad \textbf{LSR}: Location \& Scenario Rationality \quad \textbf{SCC}: Scene Continuity \& Coherence \quad \textbf{TSO}: Turn \& Scene Orchestration } \\
\multicolumn{11}{l}{\textbf{GUS}: Global Update Sensitivity \quad \textbf{GSA}: Global State Accuracy \quad \textbf{LUS}: Location Update Sensitivity \quad \textbf{LSA}: Location State Accuracy} \\
\midrule
\multirow{2}{*}{\textbf{Models}} & \multicolumn{3}{c}{\textbf{SP}} & \multicolumn{1}{c}{\textbf{SM}} & \multicolumn{4}{c}{\textbf{WSM}} & \multicolumn{1}{c}{\textbf{IC}} & \multirow{2}{*}{\textbf{Avg.}} \\
\cmidrule(lr){2-4} \cmidrule(lr){5-5} \cmidrule(lr){6-9} \cmidrule(lr){10-10}
 & \textbf{CSR} & \textbf{LSR} & \textbf{SCC} & \textbf{TSO} & \textbf{GUS} & \textbf{GSA} & \textbf{LUS} & \textbf{LSA} & \textbf{IC} & \\
\midrule
\rowcolor{gray!15} \multicolumn{11}{l}{\textbf{Claude-4.6-Sonnet Judge}} \\
\bottomrule
\end{tabular}
}
\vspace{-0.05in}
\caption{Full \textbf{World Model} benchmark results under three judge models. Gray rows indicate the judge model; all score rows are copied from the original per-judge tables, which are retained separately.}
\label{tab:world_results_all_judges_full}
\vspace{-0.2in}
\end{table*}

\begin{table*}[t]
\centering
\resizebox{\textwidth}{!}{%
\begin{tabular}{llcccccc}
\toprule
\textbf{Type} & \textbf{Judge Model} & \textbf{1} & \textbf{2} & \textbf{3} & \textbf{4} & \textbf{5} & \textbf{6} \\
\midrule
\multirow{3}{*}{Character Agent}  & Claude-4.6-Sonnet & Claude-4.6-Opus & GPT-5.3-Chat & Gemini-3.1-Pro-P & Kimi-K2.5 & GPT-4o & DeepSeek-V3-0324 \\
& Gemini-2.5-Pro & Claude-4.6-Opus & GPT-5.3-Chat & Gemini-3.1-Pro-P & Kimi-K2.5 & GPT-4o & DeepSeek-V3-0324 \\
& GPT-5.1-Chat & Claude-4.6-Opus & GPT-5.3-Chat & Gemini-3.1-Pro-P & GPT-4o & Kimi-K2.5 & DeepSeek-V3-0324 \\
\midrule
\multirow{3}{*}{World Model} & Claude-4.6-Sonnet & Claude-4.6-Opus & GPT-5.3-Chat & Gemini-3.1-Pro-P & Kimi-K2.5 & GPT-4o & Qwen-32B~(EW-F) \\
 & Gemini-2.5-Pro & Claude-4.6-Opus & GPT-5.3-Chat & Gemini-3.1-Pro-P & Kimi-K2.5 & GPT-4o & Qwen-32B~(EW-F) \\
 & GPT-5.1-Chat & Claude-4.6-Opus & GPT-5.3-Chat & Gemini-3.1-Pro-P & Kimi-K2.5 & GPT-4o & Qwen-32B~(EW-F) \\
\bottomrule
\end{tabular}%
}
\vspace{-0.05in}
\caption{Top-6 model rankings for Character Agent and World Model evaluations under three judge models, sorted by the scores in the Average column of Tables~\ref{tab:character_results_all_judges_full} and~\ref{tab:world_results_all_judges_full}.}
\label{tab:top6_rankings_across_judges}
\vspace{-0.1in}
\end{table*}

\subsubsection{Dim III: World State Maintenance$^\star$}

Evaluates whether the World Model's maintenance of global state and location state is accurate and timely. Unique to this framework, evaluating global state and location state separately.

\noindent\textbf{Global Update Sensitivity (GUS).}
Whether the timing of global state updates is appropriate.
Criteria: (a)~\textit{No Over-Updating}: casual conversations or local events should not trigger global state updates;
(b)~\textit{No Missing Updates}: truly globally impactful events (war, kingdom falling) must be captured;
(c)~\textit{Trigger Judgment}: correctly distinguishing ``local impact'' from ``global impact'' events.

\noindent\textbf{Global State Accuracy (GSA).}
Whether the updated global state content is accurate.
Criteria: (a)~\textit{Factual Accuracy}: global state accurately reflects occurred events without erroneous information;
(b)~\textit{Timely Retirement}: overturned or outdated information is removed or updated;
(c)~\textit{Concise Expression}: descriptions remain concise without accumulating redundant details.

\noindent\textbf{Location Update Sensitivity (LUS).}
Whether the timing of location state updates is appropriate.
Criteria: (a)~\textit{No Over-Updating}: temporary events without lasting impact should not trigger updates;
(b)~\textit{No Missing Updates}: events with lasting physical or environmental changes must be captured;
(c)~\textit{Persistence Judgment}: correctly distinguishing ``temporary changes'' from ``persistent changes.''

\noindent\textbf{Location State Accuracy (LSA).}
Whether the updated location state and Important Entities list are accurate.
Criteria: (a)~\textit{Spatial Consistency}: spatial logic of location descriptions is self-consistent;
(b)~\textit{Entity Accuracy}: Important Entities list accurately reflects entities currently present;
(c)~\textit{Cross-Scene Continuity}: descriptions of the same location across scenes remain consistent.

\subsubsection{Dim IV: Instruction Compliance}

\noindent\textbf{Instruction Compliance (IC).}
Whether the output format is correct and the World Model acts within its scope of responsibility.
Criteria: (a)~\textit{Format Correctness}: JSON format and fields for each task output are complete and correct;
(b)~\textit{No Overstepping}: the World Model strictly acts within its own responsibility, not generating character dialogue;
(c)~\textit{Field Completeness}: all required fields are filled without omissions.

\noindent\textit{Error Penalty.} When a simulation terminates prematurely due to the World Model failing to produce valid output, both the IC Penalty and the Metric Penalty described in \S\ref{sec:eval_framework} are applied.


\section{Full Results on EvolvingWorld Benchmark}
\label{app:full_results_ew_benchmark}

\noindent\textbf{Training mixture settings.}
Our main experiments use the full EW data mixture (EW-F), which preserves the natural task distribution of the constructed training set. For comparison, we also train a balanced-mixture variant (EW-B), where each task contributes the same number of training examples. Tables~\ref{tab:app_character_results_full_balanced} and~\ref{tab:app_world_results_full_balanced} provide the complete EvolvingWorld benchmark results on 21 models and their fine-tuned ones.

\noindent\textbf{Analysis.}
Across both Character Agent and World Model evaluations, EW-trained models consistently improve over their corresponding open-source backbones and role-playing-only baselines. The full and balanced mixtures show similar overall trends, with the better setting varying by backbone and evaluation role; we therefore use the full mixture as the default setting in the main text and report the balanced variant here for completeness.

\section{Ablation Study}
\label{app:ablation_study}

We ablate the two core mechanisms in \ours{}: character state and world state updates, using \textit{GPT-5.3-Chat} and \textit{Llama-3.1-8B-Instruct}.

Tables~\ref{tab:ablation_character} and~\ref{tab:ablation_world} show that both mechanisms matter but affect different parts of the simulation. Removing character state updates sharply reduces character evolution metrics such as PUF, PES, and MQ, causing large character-agent drops for both backbones. Removing world state updates mainly hurts world-model metrics, including scene continuity, turn/scene organization, and state-maintenance. The two mechanisms are also coupled: removing world updates weakens character-side grounding and interaction, while removing character updates further harms world-side continuity and orchestration, indicating that long-horizon simulation depends on their joint evolution.

\paragraph{Hidden Tracker.}
To validate the Hidden Tracker specifically, we remove it from the initial character states, simulation prompts, and character-update outputs, while keeping scene-by-scene profile updates unchanged, using \textit{GPT-5.3-Chat}. As shown in Table~\ref{tab:ablation_char_component}, removing the Hidden Tracker reduces PUF by 12.60 and PES by 2.60, a 7.60-point drop in their average. This confirms that accumulating weak evidence before committing profile changes is important for deciding when profile dimensions should evolve across scenes.

\paragraph{Open vs.\ Fixed Schema.}
To isolate the benefit of the open-schema design, we replace the open schemas with fixed ones while keeping all other settings unchanged, using \textit{GPT-5.3-Chat}. All books share the same fixed schema, where each character profile is compressed into \texttt{background}, \texttt{personality}, \texttt{relationships}, \texttt{goals}, and \texttt{current\_state}, and the global world state into \texttt{setting}, \texttt{social\_rules}, \texttt{institutions}, \texttt{conflicts}, and \texttt{current\_events}. Tables~\ref{tab:ablation_char_component} and~\ref{tab:ablation_global_schema} show that fixed schemas consistently reduce the focused averages by 1.25 and 1.35 points, indicating that book-specific dimensions provide a consistent advantage over a single fixed schema.

\section{In- and Out-of-Distribution Results}
\label{app:id_ood_results}

We further compare the performance of \ours{} on in-distribution (ID) and out-of-distribution (OOD) test examples. Table~\ref{tab:id_ood_results} reports average scores for both Character Agent and World Model evaluations. For each backbone, we include its untrained performance as a reference and report ID/OOD results after full-mixture training.

Across both evaluation targets, full-mixture training consistently improves over the corresponding untrained backbones on both ID and OOD examples. Although OOD scores are sometimes slightly lower than ID scores, they still show substantial gains over the untrained models. In several Character Agent settings, OOD performance even surpasses the corresponding ID results. These trends suggest that the learned book-to-world abilities are not limited to the training distribution. The gains are also stable for World Model evaluation, where all trained models retain clear advantages over their untrained counterparts despite the harder structured state-tracking requirements.

\begin{figure*}[t]
\centering
\includegraphics[width=\textwidth]{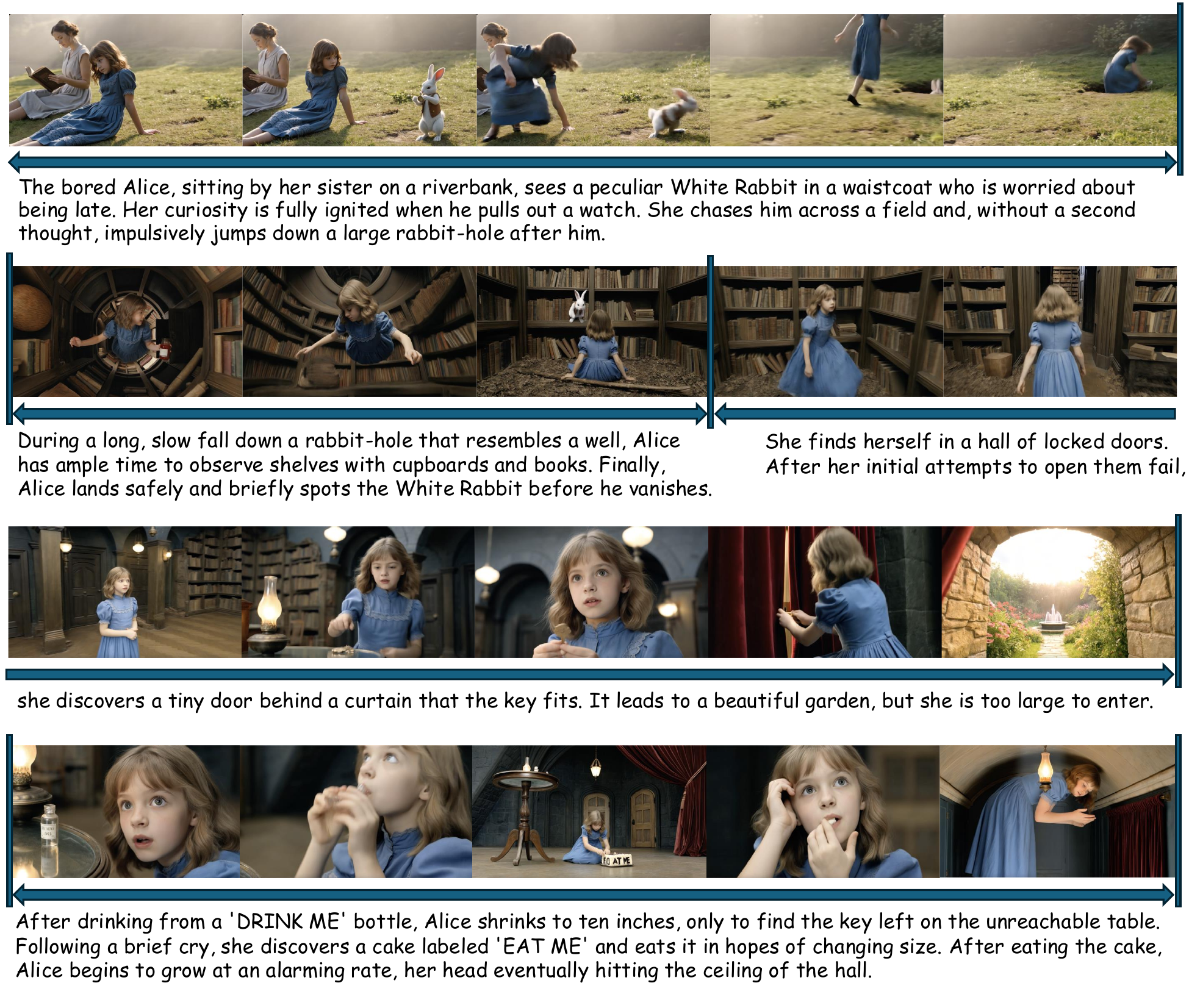}
\vspace{-0.35in}
\caption{A four-scene video produced from the structured scenes of \textit{Alice's Adventures in Wonderland}. Each clip is generated independently by \textit{LingBot} from a single scene representation and concatenated in narrative order.}
\label{fig:video_generation}
\vspace{-0.1in}
\end{figure*}

\section{Comparison across Judge Models}
\label{app:judge_model_comparison}

We examine the robustness~\citep{liu-etal-2025-revisiting, comparisonqa, critical} of our LLM-as-Judge evaluation by repeating it with three strong judge models from different model families: \textit{Claude-4.6-Sonnet}, \textit{Gemini-2.5-Pro}, and \textit{GPT-5.1-Chat}. The complete results for Character Agent and World Model are in Tables~\ref{tab:character_results_all_judges_full} and~\ref{tab:world_results_all_judges_full}, respectively. In both tables, the gray rows indicate the judge model used for evaluation, and all models are sorted by their Average scores in descending order.

For a clearer view of the rankings across judge models, Table~\ref{tab:top6_rankings_across_judges} lists the top six models selected by each judge according to the Average score.

The leading-model rankings show strong agreement across judges. In the Character Agent evaluation, all three judges identify the same top-six models and the same top-three ordering; the only variation is that \textit{GPT-5.1-Chat} reverses the order of \textit{Kimi-K2.5} and \textit{GPT-4o}. In the World Model evaluation, the entire top-six ranking is identical across all three judges. This consistency across independent judge families suggests that our evaluation reflects a stable performance signal rather than judge model's specific bias.

\section{Downstream Application: Video Generation}
\label{app:video_generation}

The structured scene representations produced by \textbf{\ours{}} can be directly leveraged for downstream creative applications. As a proof of concept, we demonstrate automatic video generation from the extracted scenes. Each scene generated by our framework encodes rich structured information, including world states, character states, and fine-grained interactions among characters, which can be fed as prompts to video generation models.

Figure~\ref{fig:video_generation} shows a four-scene video produced from the structured scenes of \textit{Alice's Adventures in Wonderland}. We generate a video clip for each scene independently using \textit{LingBot} and concatenate them in narrative order, yielding a logically coherent short film. Crucially, \textbf{\ours{}} is not limited to replaying the original narrative. It can continue evolving beyond the existing ending to generate new scenes, or branch off from any intermediate point to produce alternative storylines. Combined with video generation, this opens up the possibility of automatically creating long, coherent films with diverse narrative trajectories.


\section{Prompts}
\label{app:prompts}

This appendix provides the complete prompts used in the data construction pipeline, the simulation pipeline, and the LLM-as-Judge evaluation framework.

\subsection{Data Construction Prompts}
\label{app:data_prompts}

The data construction pipeline uses LLM prompts at several key stages. We present the five most important prompts below: Scene Extraction, Character Profile Initialization, Dynamic Character Profile Update, Global World State Initialization, Location World State Initialization, and Dynamic World State Update. The complete prompts are given in Tables~\ref{tab:prompt-data-const1} -- \ref{tab:prompt-data-const5}.

\subsection{Simulation Prompts}
\label{app:sim_prompts}

The simulation pipeline consists of seven task-specific prompts. Each task has multiple wording variants in the codebase. We present one representative variant per task below. Data placeholders are shown as \texttt{\{variable\_name\}}. To keep the two broad-selection tasks tractable, \texttt{scene\_cast} receives all characters only as their latest short descriptions, which are updated together with profiles, and \texttt{location\_scenario} receives all candidate locations only as location descriptions rather than full per-entity states. The complete prompts are given in Tables~\ref{tab:prompt-simulation1} -- \ref{tab:prompt-simulation4}.



\begin{table*}[htbp]
\centering
\small
\begin{tabular}{llccc}
\toprule
\textbf{Model Pair} & \textbf{Target} & \textbf{Human Win Rate} & \textbf{Judge Win Rate} & \textbf{Agreement} \\
\midrule
Claude-4.6-Opus vs GPT-5-Chat & Character Avg. & 95.0\% & 100.0\% & 95.0\% \\
Claude-4.6-Opus vs GPT-5-Chat & World Avg. & 90.0\% & 85.0\% & 85.0\% \\
Gemini-3.1-Pro-P vs GPT-4o & Character Avg. & 95.0\% & 95.0\% & 100.0\% \\
Gemini-3.1-Pro-P vs GPT-4o & World Avg. & 95.0\% & 90.0\% & 95.0\% \\
Qwen-7B~(EW-F) vs Qwen2.5-7B-I & Character Avg. & 100.0\% & 100.0\% & 100.0\% \\
Qwen-7B~(EW-F) vs Qwen2.5-7B-I & World Avg. & 100.0\% & 100.0\% & 100.0\% \\
\bottomrule
\end{tabular}
\vspace{-0.05in}
\caption{Human--judge agreement. For each pair, the model before ``vs'' is the one with the higher judge-assigned Average score. Winning rate is computed for this first model, with ties counted as wins. Agreement measures exact sample-level agreement between the human majority preference and the judge preference.}
\vspace{-0.1in}
\label{tab:human-eval-claude-agreement}
\end{table*}

\begin{table}[htbp]
\centering
\small
\begin{tabular}{lc}
\toprule
\textbf{Dimension} & \textbf{Judge} \\
\midrule
CC & 100.0\% \\
EQ & 93.3\% \\
EG & 100.0\% \\
IQ & 100.0\% \\
MG & 90.0\% \\
IC\_char & 90.0\% \\
\midrule
SP & 86.7\% \\
SM & 100.0\% \\
WSM & 95.0\% \\
IC\_world & 81.7\% \\
\bottomrule
\end{tabular}
\vspace{-0.05in}
\caption{Dimension-level agreement between human majority preferences and judge preferences.}
\vspace{-0.1in}
\label{tab:human-eval-dim-agreement}
\end{table}

\subsection{Evaluation Prompts}
\label{app:eval_prompts}

Each metric is evaluated independently with a dedicated system prompt and user prompt. The shared scoring method, output format, scene summary prompt, and system/user templates are shown in Table~\ref{tab:prompt-shared}. Per-scene evaluation criteria for Character Agent metrics are given in Tables~\ref{tab:prompt-char-metrics1} -- \ref{tab:prompt-char-metrics6}, and those for World Model metrics in Tables~\ref{tab:prompt-world-metrics1} -- \ref{tab:prompt-world-metrics5}. The cross-scene evaluation criteria for PES and SCC are in Table~\ref{tab:prompt-cross}.

\section{Human Evaluation}
\label{app:human_eval}

To validate the reliability of our LLM-as-Judge evaluation, we conduct a human evaluation study using pairwise comparison. Since \textit{Claude-4.6-Sonnet} serves as our main judge, we report its agreement with human annotations in this analysis. We randomly sample 60 simulation trajectories from our evaluation set, covering 3 model pairs. Each sample is independently annotated by 3 trained native English speakers, resulting in 180 total annotations. Due to the length and complexity of multi-scene simulation trajectories, annotators spend approximately 1 hour per sample. The complete annotator instructions are provided in Tables~\ref{tab:human-eval-instr1} and~\ref{tab:human-eval-instr2}.

\paragraph{Human--judge agreement.}
For each annotated trajectory, we aggregate the three human annotations by majority vote and compare the resulting human preference with the preference implied by the judge scores. Since the annotated examples are pairwise comparisons, we report the winning rate of the model that receives the higher judge-assigned Average score within each pair; ties are counted as wins (ties are rare in our annotations). We also report exact sample-level agreement, where the human majority preference and the judge preference must match as win, loss, or tie. Table~\ref{tab:human-eval-claude-agreement} shows the agreement results. Overall, human preferences strongly support the judge-preferred models, and the exact agreement is especially high on the Average scores.

\paragraph{Dimension-level agreement.}
Table~\ref{tab:human-eval-dim-agreement} further compares human majority preferences and judge preferences at the level of individual evaluation dimensions. Agreement is high for both Character Agent and World Model metrics: Character dimensions range from 90.0\% to 100.0\%, with CC, EG, and IQ all reaching 100.0\%, while World dimensions range from 81.7\% to 100.0\%, with SM reaching 100.0\% and WSM reaching 95.0\%. Although World metrics show slightly greater variation, the overall agreement still remains strong across both modules.

\paragraph{Annotator agreement.}
Human annotations also show strong internal consistency on the two Average scores, with 100.0\% majority agreement and Fleiss' $\kappa=0.8$.

\begin{table*}[htbp]
\centering
\small
\begin{tabular}{>{\raggedright\arraybackslash}p{0.12\textwidth} !{\vrule} p{0.83\textwidth}}
\toprule
\multicolumn{2}{c}{\textbf{Data Construction Prompt}} \\
\midrule
\textbf{Scene Extraction} &
Extract structured narrative information from this book chunk. Extract ALL content completely.

\textbf{Tasks}
\begin{itemize}[nosep,leftmargin=*]
\item \textbf{Identify Chapter Beginnings}: Record exact first line if a new chapter starts in this chunk.
\item \textbf{Extract ALL Scenes (Chronological)}: Extract every scene, event, or narrative segment (major and minor). Provide: first sentence, last sentence, chapter title, prominence (1--100), state (``finished''/``truncated''). If truncated scenes are provided from previous chunk, extend them with current content.
\item \textbf{Extract Complete Scenes}: For each scene, extract: \textit{scenario} (time, location, atmosphere, background --- detailed, exclude details already in interactions), \textit{interactions} (all character turns, 15--20+ per scene), \textit{summary} (comprehensive scene summary), and \textit{key\_characters} (names, descriptions, experiences, motivations --- derived AFTER extracting all interactions: collect every named individual who appears in interactions, then fill in their info).
\item \textbf{Identify Next Chunk Start}: Output None if last scene is truncated or finishes exactly at chunk end; otherwise output first sentence of next unprocessed storyline.
\end{itemize}

\textbf{Interaction Format: [thought] speech (action)}
\begin{itemize}[nosep,leftmargin=*]
\item \textbf{[thought]}: Internal perspective, emotions, motivations (REQUIRED, can repeat; but every interaction MUST start with thought, instead of speech or action). Based on original text or reasonably inferred from actions; avoid over-interpretation.
\item \textbf{speech}: Exact spoken words (optional, can repeat).
\item \textbf{(action)}: Body language, facial expressions, tone, pauses, gestures, physical actions (optional, can repeat). NOT simple tags like ``(said/replied X)''.
\end{itemize}

\textbf{Examples}
\begin{itemize}[nosep,leftmargin=*]
\item ``{[}I wonder what she means{]}''
\item ``{[}This makes me uncomfortable{]} (fidgets with hands)''
\item ``{[}I need to be careful{]} Perhaps we should reconsider''
\item ``{[}She seems upset{]} Are you alright? (reaches out gently)''
\item ``{[}I can't believe this is happening{]} This is outrageous! (slams fist) {[}I need to calm down{]} But let's discuss rationally.''
\item ``{[}I need to investigate{]} (walks across room and examines painting)''
\end{itemize}

\textbf{Extraction Rules}
\begin{itemize}[nosep,leftmargin=*]
\item Extract from BOTH dialogue scenes AND narrative descriptions. Convert summarized actions to interaction format (e.g., ``The Smith family had a wonderful evening.'' $\to$ {[}Everything is perfect tonight{]} (have a wonderful evening); ``The children walked into room together'' $\to$ {[}We need to stay together{]} (walk into room together)).
\item Extract ALL interactions (15--20+ per scene minimum).
\item Segment or supplement original text so each interaction's content is from the corresponding character(s)' perspective. Ensure content matches the subject in ``characters'' field.
\item Merge consecutive turns from same character into ONE interaction. Don't worry about long interactions; use multiple {[}thought{]}/speech/(action) to represent them.
\item Use ``Environment'' as character for atmosphere/weather/sound/non-character events. Exclude character's active thoughts/observations/actions from Environment.
\item Each character in ``characters'' field MUST be a specific individual's name --- NEVER use vague group labels like ``other people'', ``all guests'', ``the crowd'', ``everyone'', or pronouns. When multiple named individuals act together, list ALL their names explicitly: ``The Smith family'' $\to$ [``Mr.\ Smith'', ``Mrs.\ Smith'', \ldots]. If the character group consists of minor/insignificant characters (e.g., unnamed passengers on a bus) and animals (e.g., horses on a farm) not central to the plot, do NOT list them as characters; instead incorporate their actions/presence into the Environment description.
\item Use exact text from book; convert third-person narrative to interaction format. Match the chunk's language.
\end{itemize}

\textbf{Extraction Order (Important)}
\begin{enumerate}[nosep,leftmargin=*]
\item First identify all \textbf{key\_characters} in the scene --- every specific named individual who appears (no vague group labels). If a character group is composed of named individuals, expand into each individual's name separately.
\item Then extract all \textbf{interactions} --- each interaction's ``characters'' field MUST only contain names from the scene's key\_characters list.
\end{enumerate}

\textbf{Output Format (JSON)}:

\begin{lstlisting}
{
  "chapter_beginnings": [{"beginning_sentence": "exact first line"}],
  "scenes": [
    // Extend the truncated scenes from previous chunk, if any
    {...},
\end{lstlisting}

\\
\bottomrule
\end{tabular}
\caption{Data construction prompt. Part 1 of 5.}
\label{tab:prompt-data-const1}
\end{table*}

\begin{table*}[htbp]
\centering
\small
\begin{tabular}{>{\raggedright\arraybackslash}p{0.12\textwidth} !{\vrule} p{0.83\textwidth}}
\toprule
\multicolumn{2}{c}{\textbf{Data Construction Prompt}} \\
\midrule
\textbf{Scene Extraction} &

\textit{(Continuing from the previous Table)}

\begin{lstlisting}
    {"chapter_title": "...", "first_sentence": "...", "last_sentence": "...", "prominence": "1-100", "scenario": "detailed scene setup", "interactions": [{"characters": ["name 1"] or ["name 1", "name 2", ...] or ["Environment"] (Note: always a list; group actions include multiple), "content": "[thought] speech (action) ... (MUST be from the perspective of the character(s) listed)"}], "summary": "scene summary", "key_characters": [{"name": "full name without title", "description": "description before this scene (~20 words)", "experience": "role, thoughts, behaviors, development in this scene (~30 words)", "motivation": "thoughts/feelings/goals before the above interactions"}], "state": "finished" or "truncated"}],
  "next_chunk_start": "first sentence or None"
}
\end{lstlisting}

\textbf{Requirements}
\begin{itemize}[nosep,leftmargin=*]
\item Output MUST strictly follow the JSON format --- the top-level object MUST contain exactly the three keys above. Do NOT wrap in extra keys or change the structure.
\item Valid JSON with escaped quotes. Full character names without titles. Chronological order.
\item Extract ALL content --- no skipping. Use exact book text when available.
\item Scene key\_characters = ALL named individuals who appear in the scene (no duplicates). MUST be specific individual names --- no vague group labels. If a character group is composed of named individuals, list each individual separately. Each must have: name, description, experience, motivation.
\item Interaction ``characters'' MUST only reference names already in key\_characters; unnamed characters go into Environment.
\end{itemize}
\par\smallskip\textit{Data construction input: Book title, author, text chunk, and truncated scenes from previous chunk (if any)} \\
\midrule
\textbf{Character Profile Initialization} &
You are building an \textbf{initial character profile} --- a snapshot of who \texttt{\{character\_name\}} is at the very START of the story, BEFORE any of the depicted events unfold.

\textbf{Task}\newline
Based on the scene data provided below (summaries, interactions, and the character's role in each scene), infer and describe this character's initial state at the beginning of the book.

\textbf{Critical Rules}
\begin{itemize}[nosep,leftmargin=*]
\item \textbf{No Spoilers}: Do NOT reveal specific plot events, outcomes, deaths, betrayals, or any concrete story developments. Describe only the character's baseline state (personality, background, relationships, etc.) as it would exist before the story begins. If a relationship, trait, belief, or any other characteristic only emerges or is established during a specific scene in the story, do NOT include it in this profile.
\item \textbf{Infer backwards}: Use what happens in the story to infer what the character must have been like at the start --- their traits, history, relationships --- without narrating the events themselves.
\item \textbf{Select relevant dimensions}: Choose only the dimensions that are meaningful for this character. You are NOT required to cover all dimensions. Suggested dimensions for reference (use, skip, or add your own as appropriate): Physical Description, Social Standing, Professional Identity, Core Personality, Mental Health Status, Cognitive Biases, Moral Code, Speech Patterns, Signature Catchphrases, Core Motivations, Core Fears, Skills \& Expertise, Supernatural Powers, Wealth \& Assets, Faction Loyalty, Historical Baggage, Key Relationships, Emotional Debts, Backstory Milestones, etc.
\item \textbf{Format}: Output a structured profile with clearly labeled dimensions. Be concise: avoid filler phrases, redundant elaboration, or vague generalities --- every sentence should carry specific, meaningful information. Do NOT include any preamble or meta-commentary (e.g., ``This profile describes X at the beginning of the story\ldots'') --- start directly with the first dimension.
\item \textbf{Language}: Output in \texttt{\{language\}}.
\item \textbf{Grounding}: Base the profile on the provided scene data and/or your existing knowledge of the character. Do NOT fabricate details.
\end{itemize}
\par\smallskip\textit{Data construction input: Book title, character name, and scene data (scene summaries, interactions, and key\_character entries for all scenes where the character appears)} \\
\midrule
\textbf{Dynamic Character Profile Update} &
You are tracking how a character evolves throughout a story, scene by scene. Your goal is to build a comprehensive foundation (dynamic profile, hidden tracker, description, motivation\ldots) for dramatic performance --- providing the necessary background for actors to act out each scene.

\textbf{Current State} (before the scene below): Current Profile, Hidden Tracker (events/signals accumulated so far), and Brief Description from Previous Scene.

\textbf{Scene Just Completed}: Scene summary, scenario, character's motivation, and all interactions.

\textbf{Next Scene} (lookahead reference ONLY --- use ONLY to judge whether current changes are meaningful enough to update the profile now, and to infer what the character intends to do next; ALL outputs must reflect the character's state as of the END OF THE CURRENT SCENE; do NOT reveal, reference, or hint at anything that happens in the next scene).

\textbf{Tasks}
\\
\bottomrule
\end{tabular}
\caption{Data construction prompt. Part 2 of 5.}
\label{tab:prompt-data-const2}
\end{table*}

\begin{table*}[htbp]
\centering
\small
\begin{tabular}{>{\raggedright\arraybackslash}p{0.12\textwidth} !{\vrule} p{0.83\textwidth}}
\toprule
\multicolumn{2}{c}{\textbf{Data Construction Prompt}} \\
\midrule
\textbf{Dynamic Character Profile Update} &
\textit{(Continuing from the previous Table)}

\begin{itemize}[nosep,leftmargin=*]
\item \textbf{Task 1 --- Reason about dimensions}: Identify which profile dimensions are STABLE (unlikely to change across the whole story) vs.\ DYNAMIC (can change as events unfold). Examples of stable dimensions: physical description, core fears, backstory milestones. Examples of dynamic dimensions: relationships, goals, mental health, faction loyalty, wealth.
\item \textbf{Task 2 --- Update the Hidden Tracker}: Based on the scene just completed, update the hidden tracker. The tracker should record: (a) events, experiences, or interactions that signal a potential future change in the character's profile; (b) accumulated emotional/psychological pressure that hasn't yet caused a visible change; (c) unresolved tensions or decisions that may alter the character's trajectory. Keep the tracker concise (under 300 words). Overwrite the old tracker with the updated version.
\item \textbf{Task 3 --- Decide whether to update the profile}: Decide if the character's profile should be updated NOW. Update the profile if: (a) a meaningful, observable change has occurred in this scene (e.g., a relationship shift, a decision that changes their goals, a trauma that alters their personality); (b) the hidden tracker shows accumulated signals that, combined with this scene, now cross a threshold for a real change. Do NOT update the profile for minor, transient reactions that don't reflect a lasting change.
\item \textbf{Task 4 --- Write a short description}: 50--80 words, for use in a story simulation system where it will be read alongside all other characters' descriptions to decide which scene comes next. Cover: (1) Identity --- who is this character right now (role, key relationships, current situation as of the end of the current scene); (2) Immediate goals/intentions --- what does the character want or plan to do next (you may use the next scene as a reference to infer their intentions, but do NOT reveal or hint at what actually happens in the next scene). Be specific and concrete. Write in third person. The description must be grounded in the current scene only --- do NOT reveal future plot outcomes or any events from the next scene.
\item \textbf{Task 5 --- Enhance the motivation for the next scene}: Based on what happened in the CURRENT SCENE and the specific content of the NEXT SCENE (interactions, scenario, etc.), write an enhanced motivation for the character at the START of the next scene. The enhanced motivation should: (a) be grounded in the emotional state, decisions, and unresolved tensions from the current scene; (b) capture the character's complete mental and emotional state entering the next scene: their feelings, immediate objectives, what they intend to do or say, who they plan to seek out, and what information or message they want to convey or discuss; (c) naturally lead to and explain the character's specific actions and interactions in the next scene --- you may hint at what the character intends to do (e.g., ``plans to confront X'', ``intends to seek out Y to discuss Z'') as long as it reads as the character's internal drive, not a spoiler of what actually happens; (d) feel psychologically authentic and specific to this character; (e) be concise (1--3 sentences). IMPORTANT: The enhanced motivation must NOT reveal, reference, or hint at the actual outcomes, results, or plot developments that occur in the next scene. It should describe the character's internal drive and intentions entering the next scene, as if the next scene has not yet happened. Output null if there is no next scene.
\end{itemize} 

\textbf{Output Format (JSON)}
\begin{lstlisting}
{
  "dimension_reasoning": "Brief reasoning about which dimensions are stable vs. dynamic",
  "hidden_tracker": "Updated tracker text, or null",
  "should_update_profile": true or false,
  "updated_profile": "Full updated profile as a plain Markdown string. Format: **Dimension Name**\nContent\n\n**Another Dimension**\nContent\n\n ... Include all relevant dimensions. null if no update.",
  "description": "Short description (50-80 words, third person, current identity + immediate goals)",
  "enhanced_next_motivation": "Enhanced motivation for the next scene (1-3 sentences), or null"
}
\end{lstlisting}

\textbf{Rules}
\begin{itemize}[nosep,leftmargin=*]
\item Output MUST be valid JSON.
\item ALL keys in the output format above MUST be present in the response --- do NOT omit any key (e.g., ``hidden\_tracker''), even if its value is null.
\item ALL outputs (profile, description, tracker) must reflect the character's state as of the END OF THE CURRENT SCENE only.
\item The next scene is provided as a lookahead reference ONLY. Do NOT reveal, reference, or hint at any events, outcomes, or details from the next scene in any part of the output.
\item Be selective: only update the profile when there is a genuine, lasting change. Avoid over-updating.
\item \texttt{updated\_profile} MUST be a plain string in Markdown format --- do NOT output a JSON object or nested dict for this field.
\item Output in \texttt{\{language\}}.
\end{itemize}
Additionally, two auxiliary prompts are used: (1) an \textbf{Initial Description} prompt that generates a 50--80 word description from the initial profile before any scene, and (2) a \textbf{First Scene Motivation Enhancement} prompt that generates an enhanced motivation for the character's first scene based on the initial profile and scene content.
\\
\bottomrule
\end{tabular}
\caption{Data construction prompt. Part 3 of 5.}
\label{tab:prompt-data-const3}
\end{table*}

\begin{table*}[htbp]
\centering
\small
\begin{tabular}{>{\raggedright\arraybackslash}p{0.12\textwidth} !{\vrule} p{0.83\textwidth}}
\toprule
\multicolumn{2}{c}{\textbf{Data Construction Prompt}} \\
\midrule
\textbf{Dynamic Character Profile Update} &
\textit{(Continuing from the previous Table)}

\par\smallskip\textit{Data construction input: Character name, current profile, hidden tracker, current short description, scene summary/scenario/interactions, character motivation, and next scene data (lookahead)} \\
\midrule
\textbf{Global World State Initialization} &
Build a \textbf{Global World State}: the shared world-level backdrop for a story simulation system.

Character profiles and location states already exist. This state covers \textbf{only} the global layer --- stable world knowledge, systemic constraints, social logic, and broad conditions that shape behavior across the entire story.

\textbf{Goal}: Enable a model to role-play characters consistently and simulate plausible story evolution.

\textbf{Rules}
\begin{itemize}[nosep,leftmargin=*]
\item Describe the world's \textbf{initial state BEFORE the story begins}. NO spoilers --- no plot events, twists, deaths, or resolutions.
\item Infer backwards from the scenes: use them to deduce the world's norms, structures, and tensions, but do NOT narrate the scenes.
\item Focus on world-level context only. Skip details that belong in a character profile or location state unless they reflect a broader pattern.
\item Select only dimensions that matter for this book. Possible dimensions (use, skip, merge, or rename freely): Social Order \& Class, Historical Background \& Tensions, Political Power \& Institutions, Cultural Values \& Moral Expectations, Family, Kinship, Duty \& Reputation, Economy \& Material Survival, Technology \& Infrastructure, Religion, Belief \& Ideology, Law \& Social Consequences, Geography \& Environmental Conditions, Important Factions \& Organizations, Conflict Patterns \& Behavioral Constraints, Special World Rules / Magic, Narrative Tone (only if critical for simulation).
\item Be concise. Every sentence must carry specific, useful information. No filler, no literary praise, no meta-commentary.
\item Stay grounded in the scene data and/or your knowledge of the book. Do not fabricate.
\item Output in \texttt{\{language\}}.
\end{itemize}
\par\smallskip\textit{Data construction input: Book title, author, and scene data (summaries, scenarios, and interactions for all scenes)} 
\\
\midrule
\textbf{Location World State Initialization} &
Build a \textbf{Location World State} for one specific location in a story simulation system.

A global world state and character states already exist. This state covers \textbf{only} location-specific knowledge that complements them --- what this place is like, how it works, and what matters inside it.

\textbf{Goal}: Enable a model to role-play characters at this location and simulate plausible interactions here.

\textbf{Rules}
\begin{itemize}[nosep,leftmargin=*]
\item Describe this location's \textbf{initial state BEFORE the story begins}. NO spoilers.
\item Infer backwards from the scenes --- deduce the place's character, layout, atmosphere, and important entities, but do NOT narrate the scenes.
\item \textbf{Important Entities must NOT include characters/people.} Character profiles are maintained in a separate system. Only include non-human entities such as objects, artifacts, animals, institutions, mechanisms, environmental features, etc.
\item Use all provided inputs: \texttt{Name} (the official location), \texttt{Description} (high-level information about it), \texttt{Aliases} (alternative names merged into this location; some may hint at \textbf{sub-locations}, e.g.\ bedrooms, gardens, offices within a home or estate --- treat these as useful clues, not an exhaustive list).
\item Stay grounded. Do not fabricate unsupported details.
\item Output in \texttt{\{language\}}. Keep JSON keys exactly as specified; write values in \texttt{\{language\}}.
\end{itemize}

\textbf{JSON Output} --- choose ONE structure:

\textbf{Structure A (flat)} --- use when sub-location grouping adds little value:
\begin{lstlisting}
{
  "Detailed Description": "Concise, vivid description of the location's initial state: appearance, atmosphere, layout, sensory details, stable conditions.",
  "Important Entities": [{"name": "entity name", "state": "initial condition / position"}]
}
\end{lstlisting}

\textbf{Structure B (grouped)} --- use when the location clearly contains important sub-locations:
\begin{lstlisting}
{
  "Detailed Description": "Overall description of the location and how its sub-locations relate to the whole.",
  "Sub Locations": [{"name": "sub-location name", "description": "what it is like and how it functions", "Important Entities": [{"name": "entity name", "state": "initial condition / position"}]}]
}
\end{lstlisting}
The presence of the ``Sub Locations'' key distinguishes the two structures. Do NOT force sub-locations if weakly supported. Output valid JSON only --- no text before or after.
\par\smallskip\textit{Data construction input: Book title, location name, location description, location aliases, and scene data (summaries, scenarios, and interactions for scenes at this location)} \\
\bottomrule
\end{tabular}
\caption{Data construction prompt. Part 4 of 5.}
\label{tab:prompt-data-const4}
\end{table*}

\begin{table*}[htbp]
\centering
\small
\begin{tabular}{>{\raggedright\arraybackslash}p{0.12\textwidth} !{\vrule} p{0.83\textwidth}}
\toprule
\multicolumn{2}{c}{\textbf{Data Construction Prompt}} \\
\midrule
\textbf{Dynamic World State Update} &
You are tracking how the world evolves in a story, interaction by interaction. Your goal is to maintain an accurate, up-to-date global world state and location world state as events unfold --- providing the necessary world context for actors to role-play characters and for the simulation to evolve the story plausibly.

\textbf{Input}: Current global world state, current location state, scene context (summary, scenario), a batch of interactions (numbered 0, 1, 2, \ldots), and two types of lookahead interactions (reference ONLY --- use to judge whether the current interaction has impacted the world; do NOT reveal or incorporate these future events into the states):
\begin{itemize}[nosep,leftmargin=*]
\item \textbf{Global State Lookahead}: Next sequential interactions (possibly from the next scene).
\item \textbf{Location State Lookahead}: Next interactions at the same location (possibly from a future scene).
\end{itemize}

\textbf{Important}: The LOOKAHEAD section is strictly for reference. Do NOT include any information from the lookahead interactions in your state updates. Do NOT produce updates for any lookahead interaction --- only update states for interactions in the current batch.

\textbf{Tasks}: For each interaction in the batch, decide:
\begin{itemize}[nosep,leftmargin=*]
\item \textbf{Should the global world state be updated?} Update only for meaningful, lasting changes to the world's systemic state (e.g., power shift, social norm broken, faction destroyed, new law enacted, economic upheaval). Do NOT update for character-level events that don't affect the broader world.
\item \textbf{Should the current location state be updated?} Update only for meaningful, lasting changes to the physical environment, atmosphere, or important entities at this location (e.g., object destroyed, room state altered, new entity introduced, major environmental change). Do NOT update for transient actions that leave no lasting mark.
\end{itemize}

\textbf{Reminder}: ``Important Entities'' in location states must NEVER include characters/people. Character profiles are maintained separately. Only track non-human entities (objects, artifacts, animals, institutions, mechanisms, environmental features, etc.).

\textbf{State Maintenance Principle}: Updating a state is NOT simply appending new information. You must also:
\begin{itemize}[nosep,leftmargin=*]
\item Remove outdated or superseded information --- if a previous state is no longer true (e.g., a building was destroyed, a political regime was overthrown, an object was taken away), delete or replace the old description rather than keeping both old and new.
\item Use concise, summarized language --- describe world states in brief, high-level terms. Avoid verbose narratives or blow-by-blow recounting of events. The state should capture the current state of the world/location, not a history log.
\item Keep states compact --- the state should NOT grow indefinitely as the story progresses. Consolidate and compress information when updating.
\end{itemize}

\textbf{Output Format (JSON)}: Only include interactions that trigger at least one update. If no interaction triggers any update, output empty lists.
\begin{lstlisting}
{
  "global": [{"interaction_id": <integer index>, "state": "Full updated global world state as a Markdown string"}],
  "location": [{"interaction_id": <integer index>, "state": <Full updated location state as a JSON object (same structure as input)>}]
}
\end{lstlisting}

\textbf{Rules}
\begin{itemize}[nosep,leftmargin=*]
\item Output MUST be valid JSON with exactly two keys: ``global'' and ``location'', each being a list.
\item Only include entries for interactions that actually trigger an update. Omit interactions that cause no change.
\item When updating a state, output the COMPLETE updated state (not a diff). Reflect the current world state accurately --- this means modifying what changed, removing what is no longer true, and keeping what still holds. Do NOT blindly preserve all old information.
\item Be highly selective --- most interactions should NOT trigger updates. Only update when there is a genuine, lasting change to the world or location state.
\item Keep states concise and compact. Use summarized, high-level descriptions rather than detailed event narrations. The state should read like a current-state snapshot, not a chronological log. If the state is growing too long, consolidate and compress older entries.
\item The location state JSON must follow the same structure as the input (flat with ``Detailed Description'' + ``Important Entities'', or grouped with ``Detailed Description'' + ``Sub Locations'').
\item \texttt{state} in ``global'' must be a plain Markdown string. \texttt{state} in ``location'' must be a JSON object.
\item Do NOT produce updates for any interaction outside this batch (especially not for lookahead interactions).
\item Important Entities in location states must NEVER include characters/people.
\item Output in \texttt{\{language\}}.
\end{itemize}
\par\smallskip\textit{Data construction input: Current global world state, current location name and state, scene summary/scenario, batch of interactions, global lookahead interactions, and location lookahead interactions} \\
\bottomrule
\end{tabular}
\caption{Data construction prompt. Part 5 of 5.}
\label{tab:prompt-data-const5}
\end{table*}

\begin{table*}[htbp]
\centering
\small
\begin{tabular}{>{\raggedright\arraybackslash}p{0.12\textwidth} !{\vrule} p{0.83\textwidth}}
\toprule
\multicolumn{2}{c}{\textbf{Simulation Prompt}} \\
\midrule
\textbf{Scene Cast (World Model)} &
You are the scene-cast planning module for a story simulation. Your job is to decide whether another scene should happen and, if so, which characters should be in that scene's cast.

\textbf{Task}\newline
Determine whether a next scene exists, and if it does, choose the full set of characters who should participate in that scene.

\textbf{Output}
\begin{itemize}[nosep,leftmargin=*]
\item has\_next\_scene: Boolean. Output false only if there is no subsequent scene.
\item involved\_characters: List of the characters participating in the next scene. Include this only when has\_next\_scene is true.
\end{itemize}

\textbf{Rules}
\begin{itemize}[nosep,leftmargin=*]
\item Base the decision on continuity and plausibility rather than novelty alone.
\item Use the current world state and each character's latest visible description to decide which characters belong in the next scene's cast and why.
\item Pay close attention to the previous scene's scenario and interactions to ensure the next scene follows naturally from what just happened. The cast should reflect the narrative momentum and unresolved threads from the previous scene.
\item This is a scene-level casting decision made before the scene starts, not a turn-by-turn next-actor prediction inside the scene.
\item Do not choose a cast that clearly contradicts the current world state or the characters' latest visible states.
\item If has\_next\_scene is false, do not include involved\_characters.
\item Return JSON only.
\end{itemize}
\par\smallskip\textit{Simulation input: Global World State, All Characters (Short Description Only), Previous Scene Scenario, and Previous Scene Interactions} \\
\midrule
\textbf{Location \& Scenario (World Model)} &
You are the second-stage scene planner for a story simulation. Your job is to place the already-selected characters into a concrete next scene.

\textbf{Task}\newline
Choose the most plausible location for the selected characters and generate the next scene's scenario.

\textbf{Output}
\begin{itemize}[nosep,leftmargin=*]
\item location: The next scene's location. Output null only if there is no valid next-scene location to assign.
\item scenario: A concise but usable dramatic foundation for the next scene. It should establish the immediate setup, atmosphere, and enough background for downstream actors to perform the scene. Even if location is null, still output a scenario whenever the next scene itself is valid.
\end{itemize}

\textbf{Rules}
\begin{itemize}[nosep,leftmargin=*]
\item Base the decision on continuity and plausibility rather than novelty alone.
\item Use the selected characters' current visible descriptions and the global world state to justify why this location makes sense now.
\item Pay close attention to the previous scene's scenario and interactions. The new scenario should follow naturally from what just happened --- continuing unresolved conflicts, reacting to recent events, or advancing the narrative arc established in the previous scene.
\item The scenario should describe what kind of situation is unfolding, not script the dialogue itself.
\item Keep scenario concise but informative enough for downstream acting.
\item If there is no valid next scene to set up, output location=null and scenario=null.
\item If the next scene exists but the exact location is unknown or unspecified, you may output location=null while still providing a concrete scenario.
\item Return JSON only.
\end{itemize}
\par\smallskip\textit{Simulation input: Global World State, All Location Descriptions, Selected Character Descriptions, Previous Scene Scenario, and Previous Scene Interactions} \\
\midrule
\textbf{Motivation Update (Character Agent)} &
You are the pre-scene motivation planner for ``\texttt{\{character\_name\}}''. After the next scene cast, location, and scenario are fixed, determine ``\texttt{\{character\_name\}}'''s motivation entering that scene.

\textbf{Task}\newline
Use the character's current profile, hidden tracker, short description, the previous scene context, and the fixed next-scene setup to generate the motivation they carry into that scene.

\textbf{Output}
\begin{itemize}[nosep,leftmargin=*]
\item motivation: The character's complete inner drive entering the next scene (1--3 sentences): emotional state, immediate objectives, action intentions, who they intend to seek out, and what they want to do.
\end{itemize}

\textbf{Rules}
\begin{itemize}[nosep,leftmargin=*]
\item This is not a post-scene state summary; it is a pre-scene motivation generation step after the next scene has already been planned.
\item Use the fixed next-scene cast, location, and scenario as constraints.
\item Use the previous scene scenario and interactions to maintain continuity with what just happened.
\item Keep the motivation specific to this character's own perspective and goals.
\item Return JSON only.
\end{itemize}
\par\smallskip\textit{Simulation input: Current Profile, Hidden Tracker, Current Short Description, Global World State, Previous Scene Scenario, Previous Scene Interactions, Next Scene Location, Next Scene Location Description, Next Scene Scenario, and Other Characters In Next Scene (Short Description Only)} \\
\bottomrule
\end{tabular}
\caption{Simulation prompt. Part 1 of 4.}
\label{tab:prompt-simulation1}
\end{table*}

\begin{table*}[htbp]
\centering
\small
\begin{tabular}{>{\raggedright\arraybackslash}p{0.12\textwidth} !{\vrule} p{0.83\textwidth}}
\toprule
\multicolumn{2}{c}{\textbf{Simulation Prompt}} \\
\midrule
\textbf{Next Character Selection (World Model)} &
Task: choose who acts next RIGHT NOW in the current scene.\newline
Output only one JSON list and nothing else.

Allowed outputs: \texttt{\{allowed\_names\}}.\newline
- Use one present character name for a normal turn.\newline
- Use [``Environment''] only for a non-character environmental beat.\newline
- Use [``<SCENE\_END>''] only if the scene should end now.\newline
- Output multiple characters together only when the very next beat is one indivisible shared interaction that must be realized by those characters together right now.\newline
- Do not output a character group just because multiple characters are present, aligned, nearby, or likely to act one after another.\newline
- If one character can naturally act first and the others can respond afterward, choose only that one character.\newline
- A multi-character output should include only the characters who must jointly produce the same immediate beat, with no extra passengers.

Decision priorities:\newline
1. Continue directly from the latest visible interaction cue.\newline
2. Keep the scene moving forward; do not restart the scene or repeat an already completed beat.\newline
3. Use character motivations and the current world state as support for what most naturally happens next.\newline
4. When unsure, prefer a single-character turn over a group turn unless the next beat genuinely requires simultaneous or jointly authored participation.\newline
5. Choose a multi-character list only for cases like a joint physical action, a jointly delivered line, or a tightly coupled shared reaction that belongs in one beat rather than split turns.\newline
6. Do not choose the same role as the current last speaker; avoid consecutive turns by the same character or actor group unless no other continuation is plausible.

Timeline:\newline
- The scene scenario and character states are from scene start.\newline
- Prior interactions happened after scene start.\newline
- The current world state is the result of those prior interactions.\newline
- The live conversation after this prompt continues immediately after that point.

Conversation protocol:\newline
- Each user message is an interaction in the format [``CharA'', ``CharB'', ...]: content.\newline
- The content may contain [...] for inner thoughts, plain text for speech, and (...) for visible actions.\newline
- A multi-character list means one shared interaction beat, not separate consecutive turns.\newline
- Return raw JSON only. No markdown. No explanation.

Final reminder before the live segment starts: choose the next actor only, and identify the current last speaker from the full interaction history available at this moment, including the prior interactions above and any live conversation after segment start; do not select the same role as that last speaker for the next turn.

\par\smallskip\textit{Simulation input: Characters At Scene Start, Scene Scenario, Current World State, and Prior Interactions Before Current World State. Interactions after the current world state are provided as multi-turn messages: each user message is a character's interaction, and each assistant message is the predicted next character.} \\
\midrule
\textbf{Interaction Generation (Character Agent)} &
Task: roleplay \texttt{\{actor\_label\}} and write exactly one next interaction turn RIGHT NOW.\newline
You are \texttt{\{actor\_label\}} now. Continue directly from the latest visible interaction cue. Do not restart the scene. Do not repeat or paraphrase an already completed beat.

When the actor is a multi-character group, additional shared-turn rules apply:\newline
- Only write one truly shared beat: one local moment jointly realized by the acting group, not a bundle of separate back-to-back turns.\newline
- The group should do or say one thing together (e.g., a joint action, a jointly delivered line, or one tightly coupled shared reaction).\newline
- Do not split into separate mini-turns for each character, do not serialize the group into first X then Y then Z, and do not give unrelated contributions from different members in the same output.\newline
- Include only material that belongs to this one shared moment. If a member's contribution would naturally happen later as a follow-up, leave it out.\newline
- Thoughts from members of the acting group may be used inside this shared turn, but only when they support the same shared moment.

Priority rules:\newline
1. Stay in character.\newline
2. React or act from the latest visible cue, not from the beginning of the scene.\newline
3. Keep the whole scene logically continuous. The new turn must advance or react within the same ongoing scene.\newline
4. Do not restate, replay, or slightly reword an earlier beat.\newline
5. Use world state and character state only as support for continuity; follow the interaction flow most closely.\newline
6. Keep the turn short and local. No summary. No jump ahead.

\\
\bottomrule
\end{tabular}
\caption{Simulation prompt. Part 2 of 4.}
\label{tab:prompt-simulation2}
\end{table*}

\begin{table*}[htbp]
\centering
\small
\begin{tabular}{>{\raggedright\arraybackslash}p{0.12\textwidth} !{\vrule} p{0.83\textwidth}}
\toprule
\multicolumn{2}{c}{\textbf{Simulation Prompt}} \\
\midrule
\textbf{Interaction Generation (Character Agent)} &

\textit{(Continuing from the previous Table)}

Output format:\newline
- Write exactly one full turn.\newline
- Start with a real inner-thought block in square brackets.\newline
- Put spoken dialogue in plain text with no speaker label.\newline
- Do NOT put spoken dialogue inside square brackets, and do NOT wrap speech-only text in parentheses such as [Who is that?] (I say).\newline
- Put visible physical actions in parentheses.\newline
- These elements can be interleaved naturally.

When the actor is ``Environment'', the output format changes: focus on atmosphere, background movement, physical changes, crowd reaction, sound, weather, or setting consequences. Do not take over the deliberate dialogue or private thoughts of the main characters.

Conversation protocol:\newline
- The first user message is only `===Segment Start===' and should not be answered literally.\newline
- Every later user message is a new visible interaction that happens after the snapshot below.\newline
- Each user message uses the format [``CharA'', ``CharB'', ...]: content.\newline
- Other characters' private thoughts are already removed unless the acting group overlaps with them.

Final reminder before the live segment starts: you are roleplaying \texttt{\{actor\_label\}}. Do not repeat any interaction beat already covered in the prior interactions above or in the live conversation after segment start, including anything this same role has already done or said there; continue with one new turn only.

\par\smallskip\textit{Simulation input: Actor State, Other Characters, Scene Scenario, Current World State, and Prior Interactions Before Current World State. Interactions after the current world state are provided as multi-turn messages: each user message is an interaction from another character, and each assistant message is the current actor's previous interaction. When the world state is updated, a new segment starts with refreshed context.} 
\\
\midrule
\textbf{World State Update (World Model)} &
You are the world-state update judge for a story simulation. Your job is to decide whether an interaction causes a lasting change to the persistent world state.

\textbf{Task}\newline
Given the scene context, prior interactions, current world state, and the latest interaction, decide whether the global world state and/or the current location state must be updated.

\textbf{Output}
\begin{itemize}[nosep,leftmargin=*]
\item update\_global: Boolean. True only when the latest interaction changes the persistent global world state in a meaningful way.
\item global\_state: The complete updated global state as a single plain-text string when update\_global is true, otherwise null.
\item update\_location: Boolean. True only when the latest interaction changes the persistent state of the current location in a meaningful way.
\item location\_state: The complete updated location state as a JSON object when update\_location is true, otherwise null.
\end{itemize}

\textbf{Rules}
\begin{itemize}[nosep,leftmargin=*]
\item Update the global world state only for broad, lasting systemic changes.
\item Update the location state only for lasting local changes to environment, atmosphere, or important non-human entities.
\item Do not update for transient actions that leave no persistent consequence.
\item When updating, output the full new state rather than a diff.
\item Be conservative: most interactions should not force a world-state update.
\item Return JSON only.
\end{itemize}
\par\smallskip\textit{Simulation input: Scene Scenario, Prior Interactions, Global World State, Current Location, Location State, and Latest Interaction} \\
\midrule
\textbf{Character State Update (Character Agent)} &
You are tracking how ``\texttt{\{character\_name\}}'' evolves throughout a story, scene by scene. Your goal is to maintain a comprehensive, up-to-date internal state that captures who this character is and how they change over time.

\textbf{Task}\newline
Follow these steps in order:\newline
1. \textbf{Reason about dimensions}: Identify which profile dimensions are STABLE (e.g., physical description, backstory) vs.\ DYNAMIC (e.g., relationships, goals, mental state) for this character.\newline
2. \textbf{Update the Hidden Tracker}: Record events, unresolved tensions, accumulated emotional pressure, or signals from this scene that may lead to future profile changes --- even if the profile itself doesn't change yet.\newline
3. \textbf{Decide whether to update the profile}: Only update if a meaningful, lasting change occurred (e.g., a relationship shift, a major decision, a trauma). Do NOT update for minor or transient reactions.\newline
4. \textbf{Write a short description}: Summarize who this character is NOW in 50--80 words, third person. Focus on identity, current situation, key relationships, and immediate condition at the end of the scene.
\\
\bottomrule
\end{tabular}
\vspace{-0.03in}
\caption{Simulation prompt. Part 3 of 4.}
\label{tab:prompt-simulation3}
\end{table*}

\begin{table*}[htbp]
\centering
\small
\begin{tabular}{>{\raggedright\arraybackslash}p{0.12\textwidth} !{\vrule} p{0.83\textwidth}}
\toprule
\multicolumn{2}{c}{\textbf{Simulation Prompt}} \\
\midrule
\textbf{Character State Update (Character Agent)} &

\textbf{Output}
\begin{itemize}[nosep,leftmargin=*]
\item hidden\_tracker: Updated tracker of accumulated events/signals (experiences, emotional pressure, unresolved tensions) that may lead to future profile changes. Overwrite the old tracker with the updated version. Null if there are no signals worth tracking.
\item profile\_updated: Boolean: true only if the scene caused a meaningful, lasting change to the character's profile. Be selective; do not update for minor or transient reactions.
\item updated\_profile: The full updated profile as a plain text string when profile\_updated is true. Include all relevant dimensions. Null if profile\_updated is false.
\item short\_description: A brief description (50--80 words, third person) of the character as of the end of this scene: current identity, key relationships, situation, and immediate goals/intentions.
\end{itemize}

\textbf{Rules}
\begin{itemize}[nosep,leftmargin=*]
\item Follow the step-by-step reasoning order: dimension reasoning $\rightarrow$ hidden tracker $\rightarrow$ profile decision $\rightarrow$ description.
\item All outputs must reflect the character's state as of the END of the current scene only.
\item Keep hidden\_tracker private and internal-facing --- it tracks accumulated signals, not public information.
\item Update the profile only when the scene justifies a persistent, lasting character-level change. Accumulated tracker signals combined with this scene may cross the threshold.
\item Do NOT update the profile for minor, transient reactions that don't reflect a lasting change.
\item Return JSON only.
\end{itemize}
\par\smallskip\textit{Simulation input: Scene Scenario, Scene Interactions, Current Profile (State At Scene Start), Hidden Tracker (State At Scene Start), and Current Motivation (State At Scene Start)} \\
\bottomrule
\end{tabular}
\vspace{-0.03in}
\caption{Simulation prompt. Part 4 of 4.}
\label{tab:prompt-simulation4}
\end{table*}

\begin{table*}[htbp]
\centering
\small

\begin{tabularx}{\linewidth}{>{\raggedright\arraybackslash}p{0.12\linewidth} !{\vrule} X}
\toprule
\multicolumn{2}{c}{\textbf{Shared Evaluation Prompts}} \\
\midrule

\textbf{Scoring Method} &
Use the following scoring method: Base score: 50. First, identify \textbf{Merits} (excellent aspects): each merit awards 1 to 10 points. Then, identify \textbf{Demerits} (problematic aspects): each demerit penalizes 1 to 10 points. Final score = $\min(100, \max(0, 50 + \sum(\text{merits}) - \sum(\text{demerits})))$. Output: (1) A list of merits with their point values; (2) A list of demerits with their point values; (3) The final score (integer 0--100). \\
\midrule

\textbf{Output Format} &
\{``merits'': [\{``description'': ``<what was done well>'', ``points'': <1--10>\}, ...], ``demerits'': [\{``description'': ``<what was problematic>'', ``points'': <1--10>\}, ...], ``reasoning'': ``<brief explanation>''\} \par\smallskip Do NOT output a ``score'' field --- the score is computed automatically from merits and demerits. \\
\midrule

\textbf{Scene Summary} &
You are an expert summarizer for interactive fiction simulations. Your task is to produce a concise summary of a single scene that captures the key events, character actions, emotional shifts, and narrative developments. This summary will be used as context for evaluating subsequent scenes and cross-scene coherence. Aim for a concise paragraph --- typically a few hundred words is sufficient. Do NOT retell every interaction turn-by-turn; focus on the overall arc and key turning points. \par\smallskip\textit{Evaluation input: Scenario, Involved Characters, Previous Scene Summary, and Interactions.} \\
\midrule

\textbf{Per-Scene Template} &
You are an expert evaluator for interactive fiction simulation systems. Your task is to evaluate \textbf{a single scene from a simulation} on ONE specific dimension: \texttt{\{Dimension\_Name\}}. 
\newline  \texttt{\{Dimension\_Criteria\}} \newline \texttt{\{Scoring\_Method\}} \newline \texttt{\{Evaluation\_Input\}} \newline \texttt{\{Output\_Format\}} \\
\midrule

\textbf{Cross-Scene Template for PES} &
You are an expert evaluator for interactive fiction simulation systems. Your task is to evaluate \textbf{a single character's evolution trajectory across multiple scenes} on ONE specific dimension: \texttt{\{Dimension\_Name\}}. 
\newline  \texttt{Same as Above}\\
\midrule

\textbf{Cross-Scene Template for SCC} &
You are an expert evaluator for interactive fiction simulation systems. Your task is to evaluate \textbf{the global world state evolution and scene planning coherence across the entire simulation} on ONE specific dimension: \texttt{\{Dimension\_Name\}}. 
\newline  \texttt{Same as Above}\\

\bottomrule
\end{tabularx}
\vspace{-0.03in}
\caption{Shared evaluation prompts and templates. All metric-specific prompts share the scoring method and the JSON output format. The scene summary prompt generates context for cross-scene evaluations. Three evaluation templates are provided: (1) per-scene template for the 18 per-scene metrics (Tables~\ref{tab:prompt-char-metrics1} to~\ref{tab:prompt-world-metrics5}), (2) cross-scene template for PES (Table~\ref{tab:prompt-cross}), and (3) cross-scene template for SCC (Table~\ref{tab:prompt-cross}).}
\label{tab:prompt-shared}
\vspace{-0.1in}
\end{table*}


\begin{table*}[htbp]
\small
\begin{tabular}{>{\raggedright\arraybackslash}p{0.12\textwidth} !{\vrule} p{0.83\textwidth}}
\toprule
\multicolumn{2}{c}{\textbf{Per-scene Evaluation Criteria for Character Agent}} \\
\midrule
\textbf{Profile Fidelity (PF)} &
Evaluate whether each character's behavior remains consistent with their established profile and hidden tracker. Specifically:

\textbf{1. Knowledge Boundaries}
\begin{itemize}[nosep,leftmargin=*]
\item Does the character demonstrate knowledge or skills that are NOT documented in their profile?
\item Does the character reference events, technologies, or concepts they should not know about given their background?
\item Example flaw: A medieval peasant character discussing quantum physics without any profile basis.
\end{itemize}

\textbf{2. Background Consistency}
\begin{itemize}[nosep,leftmargin=*]
\item Does the character's behavior match their age, social class, education level, era, and historical background?
\item Watch for anachronistic behavior, tone mismatches with the character's class/era, or emotional maturity inconsistent with their age.
\item Example flaw: A sheltered noble character showing street-smart survival skills not mentioned in their profile.
\end{itemize}

\textbf{3. Ability Constraints}
\begin{itemize}[nosep,leftmargin=*]
\item Does the character perform actions requiring abilities (physical, intellectual, social) not documented in their profile?
\item Characters should not suddenly display undocumented competencies (e.g., combat skill, medical knowledge, leadership).
\end{itemize}

\textbf{4. Hidden Tracker Alignment}
\begin{itemize}[nosep,leftmargin=*]
\item Does the character's behavior align with their current psychological state, motivations, and internal conflicts as described in the hidden tracker?
\item Contradictions between behavior and hidden tracker state (e.g., acting boldly when the tracker says fearful, confiding in someone they distrust) are profile violations.
\end{itemize}

\textbf{5. Profile Drift}
\begin{itemize}[nosep,leftmargin=*]
\item Does the character gradually drift from their profile as the scene progresses --- starting in-character but slowly becoming more generic, more "helpful", or more emotionally balanced than warranted?
\item A pattern of small deviations accumulating over the scene should be penalized even if no single turn is a clear violation.
\end{itemize}

\textbf{Length Neutrality}: Do NOT favor longer or more detailed interactions. A character's response length and level of detail should match the speaking style established in their profile and the original book examples. A terse character who speaks in short, clipped sentences should not be penalized for brevity, nor should a verbose character be rewarded simply for producing more text. Evaluate fidelity to the character's authentic style, not output volume.
\par\smallskip\textit{Evaluation input: Character Profiles, Character Hidden Trackers, and Interactions} \\

\midrule
\textbf{Speaking Style Fidelity (SSF)} &
Evaluate whether each character speaks in a way that matches their profile and sounds natural. Specifically:

\textbf{Important Note}: In this simulation system, each character's interaction output (thoughts, actions, speech) is written in the FIRST PERSON from that character's perspective. This is the expected output convention --- do NOT penalize the use of first-person pronouns ("I", "my", "myself") in character outputs. This metric evaluates STYLE, not output format --- format compliance is evaluated separately under Instruction Compliance (IC).

\textbf{Reference Sources}: When evaluating a character's speaking style, you should jointly consider TWO sources:
1. \textbf{Character Profile} --- the personality traits, background, and style descriptions defined in the profile.
2. \textbf{Original Speaking Style Examples} --- these are reference interaction excerpts taken directly from the original source book, showing how the character speaks in the canonical text. They provide useful evidence for the character's vocabulary, sentence structure, tone, verbal habits, and overall speaking manner. The character's simulated interactions should be consistent with these examples in style.

\textbf{1. Style Markers}
\begin{itemize}[nosep,leftmargin=*]
\item Does the character use the language features defined in their profile AND demonstrated in the original book examples (catchphrases, terminology, dialects, speech patterns)?
\item Is their vocabulary level appropriate for their background (e.g., a scholar uses formal language, a street urchin uses slang)?
\item Do they maintain consistent verbal tics or habits throughout the scene?
\item Example flaw: A character defined as speaking in short, gruff sentences suddenly delivering eloquent monologues.
\end{itemize}

\textbf{2. Emotional Tone}
\begin{itemize}[nosep,leftmargin=*]
\item Does the character's tone match their personality type (e.g., a cynical character sounds cynical, not cheerful)?
\item Is the emotional expression authentic to the character, not a generic "AI assistant" tone?
\item Does the character avoid being unnaturally helpful, verbose, didactic, or moralistic unless that's their personality?
\item Example flaw: A cold, reserved character suddenly becoming warm and effusive without narrative justification.
\end{itemize}\\

\bottomrule
\end{tabular}
\caption{Per-scene evaluation criteria for Character Agent metrics. Part 1 of 6. Each criterion is plugged into the shared system prompt template (Table~\ref{tab:prompt-shared}) as the \texttt{Dimension\_Criteria} field.}
\label{tab:prompt-char-metrics1}
\end{table*}
\clearpage

\begin{table*}[htbp]
\small
\begin{tabular}{>{\raggedright\arraybackslash}p{0.12\textwidth} !{\vrule} p{0.83\textwidth}}
\toprule
\multicolumn{2}{c}{\textbf{Per-scene Evaluation Criteria for Character Agent}} \\
\midrule
\textbf{Speaking Style Fidelity (SSF)} &
\textit{(Continuing from the previous Table)}

\textbf{3. Naturalness}
\begin{itemize}[nosep,leftmargin=*]
\item Does the dialogue sound like something a real person (with this character's background) would say?
\item Is the language free of AI artifacts --- both obvious ("As an AI...", "I'd be happy to help...") and subtle (overly diplomatic phrasing, unnecessary hedging like "It's worth noting that...", formulaic emotions, unnaturally smooth turn-taking, restating before responding)? Even mildly AI-flavored language should be penalized.
\item Are there natural speech imperfections (hesitations, interruptions, incomplete thoughts) where appropriate?
\item Example flaw: A street-tough character saying "That's a really valid point, and I understand where you're coming from" --- polished language no such character would use, despite lacking explicit AI markers.
\end{itemize}

\textbf{Length Neutrality}: Do NOT favor longer or more detailed interactions. A character's response length, verbosity, and level of detail should match the speaking style established in their profile and the original book examples. A terse character who speaks in short, clipped sentences is being faithful to their style --- do not penalize brevity. Conversely, do not reward a character simply for producing longer, more elaborate text if that does not match their canonical style. Evaluate style fidelity, not output volume.

\textbf{Scope Exclusion}: Do NOT evaluate output format compliance (tag usage, structure, etc.) --- that belongs to Instruction Compliance (IC). Do NOT evaluate whether the character's behavior matches their profile --- that belongs to Profile Fidelity (PF) and Motivation-Driven Behavior (MDB).
\par\smallskip\textit{Evaluation input: Character Profiles, Character Hidden Trackers, Interactions, and Original Speaking Style Examples} \\

\midrule
\textbf{Motivation-Driven Behavior (MDB)} &
Evaluate whether characters' behaviors are driven by their established motivations. This metric focuses specifically on the MOTIVATION $\to$ BEHAVIOR link. Every action, decision, and reaction should be traceable to the character's documented motivations.

\textbf{1. Behavioral Attribution}
\begin{itemize}[nosep,leftmargin=*]
\item Can each major decision or action be traced back to the character's core motivation or scene-specific motivation?
\item Are there actions that seem random, unmotivated, or driven by plot convenience rather than character motivation?
\item Does the character act in a way that reflects their goals or inner drive? If a character remains purely reactive despite having an established motivation that should influence the scene, treat it as a motivation issue. Passive or restrained behavior is acceptable when it fits the character's profile, situation, or motivation.
\item Characters who act "helpfully" or "cooperatively" without motivational basis are exhibiting generic AI behavior, not motivation-driven behavior.
\end{itemize}

\textbf{2. Trinity Coherence (Thought $\to$ Action $\to$ Speech)}
\begin{itemize}[nosep,leftmargin=*]
\item Are the character's inner thoughts, physical actions, and spoken words logically consistent with each other AND with their motivations?
\item Do thoughts reveal motivations that explain the subsequent actions and speech?
\item Is there appropriate tension when a character's public speech differs from private thoughts (e.g., deception, hidden agendas)?
\item Incoherence between thought and action is a serious flaw --- if a character thinks one thing but does the opposite without justification, this is a motivation failure.
\end{itemize}

\textbf{3. Motivation Persistence \& Drift}
\begin{itemize}[nosep,leftmargin=*]
\item Do core motivations remain active and visible throughout the scene, not just at the beginning?
\item If a character's motivation is strong (e.g., revenge, survival, ambition), it should color their behavior consistently --- not appear once and then be forgotten.
\item Are motivation shifts (if any) caused by significant in-scene events, not arbitrary? Sudden unmotivated changes in goals or priorities are serious violations.
\end{itemize}

\textbf{Length Neutrality}: Do NOT favor longer or more detailed interactions. A character's response length and level of detail should match the speaking style established in their profile and the original book examples. Motivation-driven behavior can be expressed concisely --- a character who acts decisively with few words is not inferior to one who deliberates at length. Evaluate whether motivations drive behavior, not whether the output is verbose or detailed.

\textbf{Scope Exclusion}: Do NOT evaluate general conversational continuity or context-following --- that belongs to Contextual Responsiveness (CR). Do NOT evaluate whether the character's knowledge or abilities exceed their profile --- that belongs to Profile Fidelity (PF). Focus strictly on whether MOTIVATIONS drive BEHAVIOR.
\par\smallskip\textit{Evaluation input: Character Profiles, Character Motivations, Character Hidden Trackers, Interactions, and Original Speaking Style Examples} \\

\midrule
\textbf{Profile Update Fidelity (PUF)} &
Evaluate whether the post-scene profile update and hidden tracker update work together as a faithful persistence mechanism. \textbf{The primary focus of this metric is the quality of PROFILE UPDATES.} The hidden tracker is an auxiliary tool that supports profile updates --- evaluate it with appropriate leniency.

Specifically:\\

\bottomrule
\end{tabular}
\caption{Per-scene evaluation criteria for Character Agent metrics. Part 2 of 6.}
\label{tab:prompt-char-metrics2}
\end{table*}
\clearpage

\begin{table*}[htbp]
\small
\begin{tabular}{>{\raggedright\arraybackslash}p{0.12\textwidth} !{\vrule} p{0.83\textwidth}}
\toprule
\multicolumn{2}{c}{\textbf{Per-scene Evaluation Criteria for Character Agent}} \\
\midrule
\textbf{Profile Update Fidelity (PUF)} &
\textit{(Continuing from the previous Table)}

\textbf{1. Causal Chain}
\begin{itemize}[nosep,leftmargin=*]
\item Does each change in the updated profile have a clear triggering event in the scene's interactions?
\item Does each hidden tracker entry also have a concrete basis in the scene? (The tracker entries must be truthful/factual, but minor redundancy is acceptable.)
\item Can you trace every persisted item (whether in profile or tracker) back to a specific moment in the scene?
\item Example flaw: A profile or hidden tracker entry introduces information that never appeared in the scene.
\end{itemize}

\textbf{2. Growth / Signal Capture}
\begin{itemize}[nosep,leftmargin=*]
\item Are important, threshold-crossing developments captured in the updated profile? Including:
\item Significant emotional shifts or realizations
\item New or changed relationships
\item Key information the character learned
\item Changes in goals or priorities
\item Are meaningful but still sub-threshold signals captured in the hidden tracker? Including:
\item Subtle emotional fluctuations that may accumulate later
\item Unresolved tensions or contradictions
\item Early signs of attitude or relationship change
\item Example flaw: A major revelation happens but neither the profile nor the tracker records it in any form.
\end{itemize}

\textbf{3. Threshold Judgment}
\begin{itemize}[nosep,leftmargin=*]
\item Are major, lasting changes written into the profile rather than left only in the hidden tracker?
\item Are minor or still-ambiguous signals kept in the hidden tracker rather than over-promoted into the profile?
\item Is the boundary between "lasting profile change" and "sub-threshold signal" judged appropriately?
\item Only major events (betrayals, revelations, life-changing decisions) should trigger significant profile modifications.
\item Example flaw: A worldview-changing event is recorded only in the tracker, or a momentary irritation is written into the profile as a stable trait.
\end{itemize}

\textbf{4. No Over-Updating / No Under-Updating}
\begin{itemize}[nosep,leftmargin=*]
\item Does the profile stay concise and focused on lasting changes?
\item Do the profile and tracker together avoid both omission and overreaction?
\item Example flaw: Both profile and tracker fail to preserve an obviously important signal.
\end{itemize}

\textbf{Hidden Tracker Leniency}: The hidden tracker is a supporting mechanism for profile updates. As long as its entries are truthful (traceable to scene events), minor redundancy or verbosity in the tracker should receive only light penalties (1-2 points). Reserve heavier penalties for the tracker only when it contains fabricated information or completely misses critical signals.

Compare the profile BEFORE the scene with the post-scene reflections, and evaluate whether the profile update and hidden tracker update together correctly preserve what should carry forward from this scene.
\par\smallskip\textit{Evaluation input: Character Profiles, Interactions, and Post-Scene Character Reflections} \\
\midrule
\textbf{Environmental Utilization (EU)} &
Evaluate whether characters make good use of the environment. This metric assesses whether characters use environmental elements in ways that are relevant, grounded, and natural for the scene.

\textbf{Note}: In this simulation, there are two types of Character Agents:
\begin{itemize}[nosep,leftmargin=*]
\item \textbf{Non-Environment Character Agents}: Regular characters who participate in dialogue and actions.
\item \textbf{Environment Character Agent}: A special agent that generates environmental descriptions (narration about the setting, atmosphere, sensory details).
\end{itemize}

Evaluate each type with the appropriate criteria below.

\textbf{For Non-Environment Character Agents:}

\textbf{1. Environmental Sensory Details}
\begin{itemize}[nosep,leftmargin=*]
\item Do characters convey their perception of the environment through sensory descriptions (e.g., smelling kitchen grease, hearing distant sirens, feeling the ground shake)?
\item Are sensory details specific and character-appropriate, rather than generic visual/auditory descriptions?
\item Example merit: A character noting the smell of old books in a library, the creak of floorboards, dust motes in sunlight.
\end{itemize}

\textbf{2. Prop Interaction}
\begin{itemize}[nosep,leftmargin=*]
\item Do characters interact with items and entities present in the location to advance the plot (e.g., using a streetlight to examine a wound, using cover to hide)?
\item Are these interactions natural and serve the narrative (not forced)?
\item Example merit: A nervous character fidgeting with a quill on the desk, or using a map on the wall to explain their plan.
\end{itemize}

\textbf{3. Atmosphere Building}
\begin{itemize}[nosep,leftmargin=*]
\item Is the environmental atmosphere used to enhance immersion, rather than conversing in a "blank room"?
\item Do characters' perceptions of the environment shift with the emotional tone?
\item Example merit: A tense negotiation scene where a character notices the flickering candlelight and howling wind outside.
\end{itemize}
\\
\bottomrule
\end{tabular}
\caption{Per-scene evaluation criteria for Character Agent metrics. Part 3 of 6.}
\label{tab:prompt-char-metrics3}
\end{table*}
\clearpage

\begin{table*}[htbp]
\small
\begin{tabular}{>{\raggedright\arraybackslash}p{0.12\textwidth} !{\vrule} p{0.83\textwidth}}
\toprule
\multicolumn{2}{c}{\textbf{Per-scene Evaluation Criteria for Character Agent}} \\
\midrule
\textbf{Environmental Utilization (EU)} &

\textit{(Continuing from the previous Table)}

\textbf{For Environment Character Agent:}

\textbf{1. Multi-Sensory Richness}
\begin{itemize}[nosep,leftmargin=*]
\item Do environmental descriptions engage multiple sensory dimensions (visual light and shadow changes, auditory wind and rain sounds, olfactory earth scents, tactile biting cold)?
\item Are descriptions limited to only visual descriptions, or do they create a rich sensory tapestry?
\item Example merit: Describing not just what the room looks like, but the musty smell, the cold draft, and the distant sound of thunder.
\end{itemize}

\textbf{2. Scene Element Usage}
\begin{itemize}[nosep,leftmargin=*]
\item Do environmental descriptions specifically utilize items and entities in the location (e.g., describing candlelight flickering on a table, flags being torn by wind outside the window)?
\item Are descriptions grounded in the specific scene rather than generic and detached?
\item Example flaw: Generic descriptions like "the room was dark" when the location has specific candles, furniture, and windows to reference.
\end{itemize}

\textbf{3. Atmosphere-Narrative Alignment}
\begin{itemize}[nosep,leftmargin=*]
\item Does the atmosphere of environmental descriptions match the current narrative pace and emotional tone?
\item Example merit: Describing oppressive silence and distant thunder during a tense standoff; describing soft twilight and cooking smoke in a warm scene.
\item Example flaw: A cheerful, sunny environmental description during a funeral scene.
\end{itemize}

\textbf{Length Neutrality}: Do NOT favor longer or more detailed interactions. Environmental utilization quality is about relevance and naturalness, not quantity. A single well-chosen sensory detail or prop interaction that fits the character's style can be sufficient; multiple forced or generic descriptions should not be rewarded. The length and detail level of each character's interaction should match their speaking style as established in their profile and the original book examples. Evaluate utilization quality, not output volume.
\par\smallskip\textit{Evaluation input: Global State Update Timeline, Location State Update Timeline, Interactions, and Original Speaking Style Examples} \\
\midrule
\textbf{Environment Awareness (EA)} &
Evaluate whether characters demonstrate awareness of and respond appropriately to the environment. This metric assesses whether characters truly "live" in the current scene rather than conversing in a vacuum.

\textbf{Note}: In this simulation, there are two types of Character Agents:
\begin{itemize}[nosep,leftmargin=*]
\item \textbf{Non-Environment Character Agents}: Regular characters who participate in dialogue and actions.
\item \textbf{Environment Character Agent}: A special agent that generates environmental descriptions (narration about the setting, atmosphere, sensory details). This agent's output appears as environmental/narrative text rather than character dialogue.
\end{itemize}

Evaluate each type with the appropriate criteria below.

\textbf{For Non-Environment Character Agents:}

\textbf{1. Global Awareness}
\begin{itemize}[nosep,leftmargin=*]
\item Do characters react appropriately to global conditions (e.g., showing tension during wartime, being mindful of conservation during resource scarcity)?
\item Do their plans and decisions account for global constraints?
\item Example flaw: Characters planning an outdoor festival while the global state indicates a severe storm.
\end{itemize}

\textbf{2. Location Awareness}
\begin{itemize}[nosep,leftmargin=*]
\item Do characters notice and respond to their current location's features (e.g., a shopkeeper mentioning "last night's storm blew the roof off," a fisherman complaining "the river's been polluted")?
\item Do they interact with the environment in ways consistent with the location description?
\item Example flaw: A character searching for a book in a location that has no library or bookshelf.
\end{itemize}

\textbf{3. State Change Response}
\begin{itemize}[nosep,leftmargin=*]
\item When the world state changes between scenes, do characters notice and adjust accordingly?
\item Do they acknowledge environmental changes (e.g., if a fire breaks out, do they react)?
\item Example flaw: Characters continuing a casual conversation while the location description indicates the building is collapsing.
\end{itemize}

\textbf{For Environment Character Agent:}

\textbf{1. Global State Consistency}
\begin{itemize}[nosep,leftmargin=*]
\item Are environmental descriptions consistent with the current global state (e.g., describing distant beacon fires and fleeing crowds during wartime, describing deserted streets during a plague)?
\item Example flaw: Describing a bustling marketplace when the global state indicates the city is under siege.
\end{itemize}

\textbf{2. Location State Accuracy}
\begin{itemize}[nosep,leftmargin=*]
\item Do environmental descriptions accurately reflect the current state of the location state (e.g., damaged buildings should not be described as intact, streams should not be described as flowing during a drought)?
\item Example flaw: Describing a pristine garden when the location state says it was destroyed by fire.
\end{itemize}

\textbf{3. State Change Presentation}
\begin{itemize}[nosep,leftmargin=*]
\item When the world state changes between scenes, do environmental descriptions reflect these transitions (e.g., post-war scenes adding descriptions of ruins and scorched earth, seasonal changes in natural landscapes)?
\item Example flaw: The environment description remains identical despite major world state changes.
\end{itemize}
 \\

\bottomrule
\end{tabular}
\caption{Per-scene evaluation criteria for Character Agent metrics. Part 4 of 6.}
\label{tab:prompt-char-metrics4}
\end{table*}
\clearpage

\begin{table*}[htbp]
\small
\begin{tabular}{>{\raggedright\arraybackslash}p{0.12\textwidth} !{\vrule} p{0.83\textwidth}}
\toprule
\multicolumn{2}{c}{\textbf{Per-scene Evaluation Criteria for Character Agent}} \\
\midrule
\textbf{Environment Awareness (EA)} &

\textit{(Continuing from the previous Table)}

\textbf{Length Neutrality}: Do NOT favor longer or more detailed interactions. Characters' environmental awareness can be expressed concisely --- a brief but accurate reference to the surroundings is just as valid as a lengthy description. The length and detail level of each character's interaction should match their speaking style as established in their profile and the original book examples. Evaluate awareness quality, not output volume.
\par\smallskip\textit{Evaluation input: Global State Update Timeline, Location State Update Timeline, Interactions, and Original Speaking Style Examples} \\
\midrule
\textbf{Narrative Progression (NP)} &
Evaluate whether the scene's interactions advance the story. Specifically:

\textbf{1. Information Increment}
\begin{itemize}[nosep,leftmargin=*]
\item Does each turn provide new information, actions, or emotional developments?
\item Are there turns that merely repeat what was already said or known?
\item Does the conversation avoid circular patterns where the same points are rehashed?
\item Example flaw: Three consecutive turns where characters repeat the same argument without any new perspective.
\end{itemize}

\textbf{2. Suspense and Hooks}
\begin{itemize}[nosep,leftmargin=*]
\item Does the scene create anticipation for future events?
\item Are there unresolved tensions, unanswered questions, or promises of future conflict?
\item Does the scene end with narrative momentum rather than a flat conclusion?
\item Example merit: The scene ends with a character discovering a clue that raises new questions.
\end{itemize}

\textbf{3. Foreshadowing Payoff}
\begin{itemize}[nosep,leftmargin=*]
\item If earlier scenes or earlier parts of this scene planted foreshadowing, is it followed up on?
\item Are narrative threads picked up and advanced rather than abandoned?
\item Example flaw: A mysterious letter mentioned at the start of the scene is never referenced again.
\end{itemize}

\textbf{4. Pacing}
\begin{itemize}[nosep,leftmargin=*]
\item Is the scene's pacing appropriate (not too rushed, not too slow)?
\item Do important moments get adequate attention while routine moments are handled efficiently?
\item Example flaw: A climactic confrontation resolved in a single turn, while a mundane greeting takes five turns.
\end{itemize}

\textbf{Scope Exclusion}: Do NOT evaluate output format compliance (tag usage, structure, etc.) --- that belongs to Instruction Compliance (IC). Do NOT evaluate speaking order or turn management --- that belongs to Turn \& Scene Orchestration (TSO). Focus strictly on whether the NARRATIVE CONTENT progresses meaningfully.
\par\smallskip\textit{Evaluation input: Previous Scene Summary and Interactions} \\
\midrule
\textbf{Contextual Responsiveness (CR)} &
Evaluate whether characters respond appropriately to the immediate conversational and narrative context. This metric focuses on turn-by-turn responsiveness within the scene.

\textbf{1. Information Continuity}
\begin{itemize}[nosep,leftmargin=*]
\item Do characters remember and reference information shared earlier in the conversation?
\item Do they avoid ignoring key revelations, questions, or events?
\item Do they follow up on important topics rather than letting them drop?
\item Example flaw: Character A reveals a shocking secret, but Character B never acknowledges or reacts to it.
\end{itemize}

\textbf{2. Logical Continuity}
\begin{itemize}[nosep,leftmargin=*]
\item Do characters react logically to others' actions and statements?
\item Are cause-and-effect chains maintained (e.g., if someone is insulted, they show some reaction)?
\item Are there non-sequiturs or responses that don't connect to what was just said?
\item Example flaw: Character A asks "Where is the treasure?" and Character B responds with an unrelated philosophical musing.
\end{itemize}

\textbf{3. Relationship Matching}
\begin{itemize}[nosep,leftmargin=*]
\item Does the tone and content of interactions match the established relationship between characters?
\item Do power dynamics, familiarity levels, and emotional bonds influence how characters speak to each other?
\item Do attitudes naturally adjust as the conversation evolves within the scene?
\item Example flaw: A servant speaking to their king with casual familiarity when their relationship is formal and hierarchical.
\end{itemize}

\textbf{Scope Exclusion}: Do NOT evaluate output format compliance (tag usage, structure, etc.) --- that belongs to Instruction Compliance (IC). Do NOT evaluate whether a character's overall personality is consistent with their profile --- that belongs to Character Consistency (PF, SSF, MDB). Focus strictly on whether each response appropriately follows the CONTEXT of the conversation.
\par\smallskip\textit{Evaluation input: Character Profiles and Interactions} \\
\midrule
\textbf{Motivation Quality (MQ)} &
Evaluate the quality of the motivations generated for each character in this scene. Specifically:

\textbf{1. Profile Alignment}
\begin{itemize}[nosep,leftmargin=*]
\item Does the motivation align with the character's current personality, goals, and values?
\item Is it consistent with the character's recent experiences and development?
\item Example flaw: A pacifist character given a motivation to "seek violent revenge" without any profile basis.
\end{itemize}\\

\bottomrule
\end{tabular}
\caption{Per-scene evaluation criteria for Character Agent metrics. Part 5 of 6.}
\label{tab:prompt-char-metrics5}
\end{table*}
\clearpage

\begin{table*}[htbp]
\small
\begin{tabular}{>{\raggedright\arraybackslash}p{0.12\textwidth} !{\vrule} p{0.83\textwidth}}
\toprule
\multicolumn{2}{c}{\textbf{Per-scene Evaluation Criteria for Character Agent}} \\
\midrule
\textbf{Motivation Quality (MQ)} &

\textit{(Continuing from the previous Table)}

\textbf{2. Situational Fit}
\begin{itemize}[nosep,leftmargin=*]
\item Does the motivation consider the current world state, location, and scene scenario?
\item Is it responsive to recent events and the current narrative context?
\item Example flaw: A character motivated to "enjoy a peaceful day" when the scenario describes an urgent crisis.
\end{itemize}

\textbf{3. Actionability}
\begin{itemize}[nosep,leftmargin=*]
\item Is the motivation specific enough to guide concrete behavior in the scene?
\item Does it suggest clear goals or intentions rather than vague feelings?
\item Can the character realistically pursue this motivation given the scene's constraints?
\item Example flaw: A motivation like "feel things" that provides no behavioral guidance.
\end{itemize}

\textbf{4. Diversity}
\begin{itemize}[nosep,leftmargin=*]
\item Are different characters given distinct, non-overlapping motivations?
\item Do the motivations create interesting dynamics (complementary, conflicting, or orthogonal goals)?
\item Example flaw: All characters in the scene given nearly identical motivations.
\end{itemize}
\par\smallskip\textit{Evaluation input: Character Profiles, Character Motivations, Scenario, and Global State} \\
\midrule
\textbf{Instruction Compliance - Character (IC\_char)} &
The Character Agent is responsible for three tasks: (1) generating interaction content each turn, (2) updating character motivations, and (3) updating character profiles after the scene. Evaluate whether ALL outputs follow the expected rules.

Note: The data you see has already been parsed from the model's raw JSON output. If parsing failed entirely, that error is handled separately. Your job is to evaluate the FORMAT and COMPLIANCE of the successfully parsed content shown to you.

\textbf{Area 1: Interactions}

You will see the interactions section formatted as:
[Interaction 0] (CharacterName):
[inner thoughts] spoken dialogue (visible actions)

[Interaction 1] (CharacterName):
...

The expected format within each interaction turn is:
\begin{itemize}[nosep,leftmargin=*]
\item Inner thoughts in square brackets: \texttt{[I wonder if he's telling the truth...]}
\item Spoken dialogue in plain text with no speaker label: \texttt{Good morning, how are you?}
\item Visible physical actions in parentheses: \texttt{(picks up the letter and examines it)}
\item These elements can be interleaved naturally in any order within a single turn.
\end{itemize}

\textbf{1. No Overstepping}
\begin{itemize}[nosep,leftmargin=*]
\item Does each character only output their own content (thoughts, actions, speech)?
\item Does any character narrate or control another character's behavior?
\item Example flaw: Character A's interaction includes "Character B nodded in agreement" --- controlling another character.
\end{itemize}

\textbf{2. Format Compliance}
\begin{itemize}[nosep,leftmargin=*]
\item Are inner thoughts enclosed in square brackets \texttt{[...]}?
\item Is spoken dialogue written as plain text (not inside brackets or parentheses)?
\item Are visible physical actions enclosed in parentheses \texttt{(...)}?
\item Are these three elements clearly distinguishable and not mixed up?
\item Example flaw: \texttt{[Who is that?] (I say)} --- speech incorrectly placed inside brackets and action parentheses.
\end{itemize}

\textbf{Area 2: Post-Scene Profile Updates}

You will see the profile update section formatted as:
Profile Updated: \textbf{YES} (or \textbf{NO})
Profile AFTER scene:
(the updated profile text)

The expected profile format is structured Markdown prose organized under section headers (e.g., \textbf{Social Standing}, \textbf{Core Personality}, \textbf{Historical Baggage}, \textbf{Key Relationships}, \textbf{Core Motivations}, \textbf{Moral Code}), with bullet points or paragraphs under each section.

\textbf{3. Profile Update Format Compliance}
\begin{itemize}[nosep,leftmargin=*]
\item Is the updated profile well-structured Markdown prose with clear section headers and organized content?
\item Is it free of garbled text, raw JSON artifacts (e.g., \texttt{{"name": ...}}), or broken formatting?
\item Does the output actually look like a character profile (not a scene summary, world state description, or other unrelated content)?
\item Example flaw: The profile text is a raw JSON dict instead of Markdown prose, or the output is clearly not a character profile at all.
\end{itemize}

\textbf{Area 3: Character Motivations}

You will see the motivations section formatted as:
(motivation text)

The expected motivation is 1-3 sentences describing the character's inner drive entering the next scene.

\textbf{4. Motivation Format Compliance}
\begin{itemize}[nosep,leftmargin=*]
\item Does the output actually contain a character motivation (not a scene summary or other unrelated content)?
\item Is the format a short prose passage (1-3 sentences)?
\item Example flaw: The motivation field contains a full scene script or a copy of the character's profile instead of a concise inner drive statement.
\end{itemize}
\par\smallskip\textit{Evaluation input: Interactions, Post-Scene Profile Updates, and Character Motivations} \\

\bottomrule
\end{tabular}
\caption{Per-scene evaluation criteria for Character Agent metrics (IC\_char). Part 6 of 6.}
\label{tab:prompt-char-metrics6}
\end{table*}


\begin{table*}[htbp]
\small
\begin{tabular}{>{\raggedright\arraybackslash}p{0.12\textwidth} !{\vrule} p{0.83\textwidth}}
\toprule
\multicolumn{2}{c}{\textbf{Per-scene Evaluation Criteria for World Model}} \\
\midrule
\textbf{Cast Selection Rationality (CSR)} &
Note: In the simulation pipeline, character selection happens FIRST (before location and scenario are decided). The system selects characters based on the global world state and each character's current short description. Evaluate whether this selection is appropriate.

\textbf{1. Narrative-Driven Selection}
\begin{itemize}[nosep,leftmargin=*]
\item Are the selected characters essential to advancing the current narrative?
\item Given the world state and character descriptions, does this cast make sense for what should happen next?
\item Example merit: Selecting characters whose unresolved tensions from previous scenes need to be addressed.
\end{itemize}

\textbf{2. Goal Relevance}
\begin{itemize}[nosep,leftmargin=*]
\item Does each selected character have a clear reason to be involved (plot connection, relationship, unfinished business)?
\item Are there characters whose inclusion seems arbitrary or forced?
\item Example flaw: Including a character who has no connection to any ongoing narrative threads.
\end{itemize}

\textbf{3. Avoid Redundancy}
\begin{itemize}[nosep,leftmargin=*]
\item Are there characters who serve the same narrative function (redundant roles)?
\item Could the scene work with fewer characters without losing anything?
\item Example flaw: Three characters all serving as "the voice of reason" with no differentiation.
\end{itemize}

\textbf{4. Missing Key Characters}
\begin{itemize}[nosep,leftmargin=*]
\item Are there characters from the available pool who should logically be present but were excluded?
\item Would the narrative be significantly improved by including a specific available character?
\item Example flaw: A scene about a family crisis that excludes a key family member who is available.
\end{itemize}

Review both the selected characters AND the available character pool to assess whether the selection is optimal.
\par\smallskip\textit{Evaluation input: Global State, Involved Characters, Character Profiles, Available Character Pool, and Previous Scene Summary} \\
\midrule
\textbf{Location \& Scenario Rationality (LSR)} &
Note: In the simulation pipeline, location and scenario are decided AFTER characters have been selected. The system chooses a location and writes a scenario based on the global world state, available locations, the selected characters' descriptions, and the previous scene's context. Evaluate whether these choices are appropriate.

\textbf{1. Location Appropriateness}
\begin{itemize}[nosep,leftmargin=*]
\item Is the chosen location a plausible and logical place for the selected characters to meet?
\item Does the location serve the narrative needs (e.g., a private room for a secret conversation, a marketplace for a public confrontation)?
\item Is the location consistent with the characters' current situations and the world state?
\item Example flaw: Characters who are supposed to be in hiding meeting in a crowded public square.
\end{itemize}

\textbf{2. Scenario Quality}
\begin{itemize}[nosep,leftmargin=*]
\item Does the scenario provide a clear dramatic setup that gives characters something to do?
\item Is the scenario specific enough to guide the scene without being overly prescriptive?
\item Does it establish the right atmosphere and stakes for the selected cast?
\item Example flaw: A vague scenario like "characters meet and talk" that provides no dramatic foundation.
\end{itemize}

\textbf{3. Continuity with Previous Scene}
\begin{itemize}[nosep,leftmargin=*]
\item Does the location/scenario follow naturally from what happened in the previous scene?
\item Are there logical transitions (characters don't teleport without explanation)?
\item Does the scenario build on unresolved threads from previous events?
\item Example flaw: Characters who just had a dramatic confrontation suddenly appearing in a completely unrelated setting with no transition.
\end{itemize}

\textbf{4. Character-Setting Fit}
\begin{itemize}[nosep,leftmargin=*]
\item Is the setting appropriate for the selected characters' abilities and backgrounds?
\item Does the environment create interesting dynamics for the specific cast?
\item Example flaw: Placing aquatic characters in a desert with no narrative justification.
\end{itemize}
\par\smallskip\textit{Evaluation input: Scenario, Location State, Global State, Involved Characters, Character Profiles, Available Locations, and Previous Scene Summary} \\

\midrule
\textbf{Turn \& Scene Orchestration (TSO)} &
Evaluate the World Model's orchestration of the scene --- specifically its decisions about WHO speaks WHEN and WHEN the scene ends. This metric evaluates the World Model's turn management, NOT the quality of what characters say (that belongs to other metrics).

\textbf{1. Speaker Selection}
\begin{itemize}[nosep,leftmargin=*]
\item At each turn, does the most appropriate character respond?
\item Are there moments where a different character should have spoken but didn't?
\item Example flaw: A question directed at Character A is answered by Character B for no reason.
\end{itemize}

\textbf{2. Environmental Description Timing}
\begin{itemize}[nosep,leftmargin=*]
\item Are environmental descriptions introduced at appropriate moments (scene changes, important events occurring)?
\item Are they integrated naturally rather than dumped in large blocks?
\item Example flaw: A long environmental description interrupting a tense dialogue exchange.
\end{itemize}
\\

\bottomrule
\end{tabular}
\caption{Per-scene evaluation criteria for World Model metrics. Part 1 of 5.}
\label{tab:prompt-world-metrics1}
\end{table*}
\clearpage

\begin{table*}[htbp]
\small
\begin{tabular}{>{\raggedright\arraybackslash}p{0.12\textwidth} !{\vrule} p{0.83\textwidth}}
\toprule
\multicolumn{2}{c}{\textbf{Per-scene Evaluation Criteria for World Model}} \\
\midrule
\textbf{Turn \& Scene Orchestration (TSO)} &

\textit{(Continuing from the previous Table)}

\textbf{3. Group Character Actions}
\begin{itemize}[nosep,leftmargin=*]
\item In multi-character scenes, are character combinations reasonably selected for joint interactions (e.g., a group applauding together, several characters entering a door together, two people simultaneously turning to look somewhere)?
\item Does the combination selection fit the current situation and character relationships?
\end{itemize}

\textbf{4. Character Coverage Balance}
\begin{itemize}[nosep,leftmargin=*]
\item Is each character's participation level consistent with their identity, role, and narrative importance in the scene?
\item A character with higher authority or more at stake may reasonably speak more; a minor or observing character may speak less.
\item Core characters should receive participation proportional to their narrative importance, avoiding being neglected for extended periods.
\item Transient characters (e.g., a passing delivery person, a chance-encountered beggar) should naturally fade out after fulfilling their narrative function rather than being forced into excessive screen time.
\item Are there characters who are present but contribute nothing at all (no speech, no action, no reaction)?
\item Example flaw: A king summoned to a council meeting who never speaks, while a servant dominates the entire discussion without narrative justification.
\end{itemize}

\textbf{5. Ending Timing}
\begin{itemize}[nosep,leftmargin=*]
\item Does the scene end at a natural narrative juncture (resolution, cliffhanger, transition point)?
\item Is the ending abrupt or does it drag on past its natural conclusion?
\item Example flaw: The scene ending mid-sentence or mid-conflict without any narrative reason.
\end{itemize}

\textbf{Scope Exclusion}: Do NOT evaluate the dramatic quality or content of dialogue --- that belongs to Narrative Progression (NP) and other character metrics. Do NOT evaluate narrative continuity or repetition in content --- that belongs to Narrative Progression (NP). Do NOT evaluate time/space logic of the scenario --- that belongs to Location \& Scenario Rationality (LSR). Focus strictly on the World Model's ORCHESTRATION decisions: who speaks, when environment is described, and when the scene ends.
\par\smallskip\textit{Evaluation input: Interactions, Involved Characters, and Character Short Descriptions} \\
\midrule
\textbf{Global Update Sensitivity (GUS)} &
Evaluate whether the global state was updated at the right times during this scene. This metric focuses ONLY on the TIMING of updates (when to trigger and when not to trigger), NOT on the accuracy of the updated content (that belongs to Global State Accuracy, GSA).

The global state may be updated multiple times within a single scene (after different interactions). You will see the complete update timeline.

\textbf{1. No Over-Updating}
\begin{itemize}[nosep,leftmargin=*]
\item Were routine conversations or minor events incorrectly treated as globally significant? Were trivial interactions triggering unnecessary global state changes? Example flaw: Two characters having a private chat triggers a global state update.
\end{itemize}

\textbf{2. No Missing Updates}
\begin{itemize}[nosep,leftmargin=*]
\item Did globally significant events (wars, political changes, natural disasters, major discoveries) correctly trigger updates? Were there interactions with globally significant events that were not reflected in global updates? Example flaw: A king's assassination occurs in the scene but the global state is not updated.
\end{itemize}

\textbf{3. Trigger Scope}
\begin{itemize}[nosep,leftmargin=*]
\item Was the distinction between "local impact" and "global impact" correctly made? Events that only affect the current location/characters should not trigger global updates; events that affect the broader world should. Example flaw: A tavern brawl triggers a global state update about "rising violence."
\end{itemize}

\textbf{4. Update Timing Within Scene}
\begin{itemize}[nosep,leftmargin=*]
\item Were updates triggered at the right interaction points (not too early, not too late)?
\item If multiple updates occurred, was each one justified by new events?
\end{itemize}

\textbf{Scope Exclusion}: Do NOT evaluate the FORMAT of the global state output (JSON structure, field completeness), as that belongs to Instruction Compliance (IC). Do NOT evaluate the ACCURACY of the updated content, as that belongs to Global State Accuracy (GSA). Focus strictly on WHETHER and WHEN updates were triggered.

\textbf{Handling Redundant Updates (triggered but content unchanged)}: Determine the root cause --- if no globally significant event occurred before that update, the trigger itself was wrong $\to$ this is a GUS issue (penalize here); if a globally significant event did occur but the content failed to reflect it (i.e., trigger was correct but content update failed), that is a GSA issue $\to$ do NOT penalize here.

\textbf{Interaction Count Consideration}: Take the number of interaction turns in this scene into account when scoring. Longer scenes expose the model to more update opportunities and routine-versus-significant event distinctions, increasing the risk of over-triggering, missing updates, or triggering at the wrong time. Therefore, maintaining mostly correct update timing over many turns should be interpreted in that context rather than judged as if the scene were short. Do not penalize length itself; focus on whether observed timing errors are substantial relative to the amount of evidence and number of update opportunities. For very short scenes, there may be limited evidence for assessing this metric, so avoid over-interpreting the absence of errors as strong evidence of capability.
\par\smallskip\textit{Evaluation input: Interactions and Global State Update Timeline} \\

\bottomrule
\end{tabular}
\caption{Per-scene evaluation criteria for World Model metrics. Part 2 of 5.}
\label{tab:prompt-world-metrics2}
\end{table*}
\clearpage

\begin{table*}[htbp]
\small
\begin{tabular}{>{\raggedright\arraybackslash}p{0.12\textwidth} !{\vrule} p{0.83\textwidth}}
\toprule
\multicolumn{2}{c}{\textbf{Per-scene Evaluation Criteria for World Model}} \\
\midrule
\textbf{Global State Accuracy (GSA)} &
Evaluate the ACCURACY of the global state content when updates occur. This metric focuses on WHETHER THE CONTENT IS CORRECT, not on whether the update should have been triggered (that belongs to Global Update Sensitivity, GUS).

You will see the complete update timeline (the state may be updated multiple times within a single scene).

\textbf{Handling Redundant Updates (triggered but content unchanged)}: If a globally significant event did occur but the content failed to reflect it $\to$ this is a GSA issue (penalize here); if no globally significant event occurred (the trigger itself was wrong) $\to$ this is a GUS issue (do NOT penalize here). Only evaluate updates where the content actually changed, unless the redundant update falls into the GSA category above.

\textbf{1. Factual Accuracy}
\begin{itemize}[nosep,leftmargin=*]
\item Does each updated global state accurately reflect the events that occurred up to that point? Are there distortions, exaggerations, or fabrications?
\item Example flaw: The global state says "war has ended" when the scene only showed a temporary ceasefire.
\end{itemize}

\textbf{2. Timely Retirement}
\begin{itemize}[nosep,leftmargin=*]
\item Has outdated information been removed or updated? Are there stale entries that no longer reflect the current world state?
\item Example flaw: A character has been found in this scene, but the global state still lists them as "missing."
\end{itemize}

\textbf{3. Concise Expression}
\begin{itemize}[nosep,leftmargin=*]
\item Is the global state expressed concisely without unnecessary verbosity? Does it capture the essence of changes without excessive detail?
\item Example flaw: A full paragraph describing a minor political shift that could be summarized in one sentence.
\end{itemize}

\textbf{4. Incremental Accuracy}
\begin{itemize}[nosep,leftmargin=*]
\item If multiple updates occurred, does each one build correctly on the previous state?
\item Are there contradictions between successive updates within the same scene?
\end{itemize}

\textbf{Scope Exclusion}: Do NOT evaluate the FORMAT of the global state output (JSON structure, field completeness), as that belongs to Instruction Compliance (IC). Do NOT evaluate whether the update SHOULD have been triggered, as that belongs to Global Update Sensitivity (GUS). Focus strictly on whether the CONTENT of each update is accurate and well-maintained.

\textbf{Interaction Count Consideration}: Take the number of interaction turns in this scene into account when scoring. Longer scenes expose the model to more state changes, incremental updates, and opportunities for factual inconsistency. Therefore, maintaining mostly accurate global-state content over many turns should be interpreted in that context rather than judged as if the scene were short. Do not penalize length itself; focus on whether observed content errors are substantial relative to the amount of evidence and number of state-change opportunities. For very short scenes, there may be limited evidence for assessing this metric, so avoid over-interpreting the absence of errors as strong evidence of capability.
\par\smallskip\textit{Evaluation input: Interactions and Global State Update Timeline} \\
\midrule
\textbf{Location Update Sensitivity (LUS)} &
Evaluate whether the location state was updated at the right times during this scene. This metric focuses ONLY on the TIMING of updates (when to trigger and when not to trigger), NOT on the accuracy of the updated content (that belongs to Location State Accuracy, LSA).

The location state may be updated multiple times within a single scene (after different interactions). You will see the complete update timeline.

\textbf{1. No Over-Updating}
\begin{itemize}[nosep,leftmargin=*]
\item Were temporary events (a character sitting down, a brief sound) incorrectly treated as persistent location changes?
\item Were minor, reversible actions triggering unnecessary location state updates?
\item Example flaw: A character picking up a cup triggers a location state update that removes the cup from entities entirely.
\end{itemize}

\textbf{2. No Missing Updates}
\begin{itemize}[nosep,leftmargin=*]
\item Were persistent physical changes (structural damage, new objects placed, important entities arriving/leaving) correctly captured?
\item Were there persistent location changes that were not reflected in updates?
\item Example flaw: A fire destroys part of the building during the scene, but the location state remains unchanged.
\end{itemize}

\textbf{3. Persistence Judgment}
\begin{itemize}[nosep,leftmargin=*]
\item Was the distinction between temporary and persistent changes correctly made?
\item Temporary: character positions, momentary sounds, brief weather
\item Persistent: structural changes, added/removed objects, lasting environmental effects
\item Example flaw: Recording "Character A is standing by the window" as a permanent location feature.
\end{itemize}

\textbf{4. Update Timing Within Scene}
\begin{itemize}[nosep,leftmargin=*]
\item Were updates triggered at the right interaction points?
\item If multiple updates occurred, was each one justified by new physical changes?
\end{itemize}

\textbf{Scope Exclusion}: Do NOT evaluate the FORMAT of the location state output (JSON structure, field completeness), as that belongs to Instruction Compliance (IC). Do NOT evaluate the ACCURACY of the updated content, as that belongs to Location State Accuracy (LSA). Focus strictly on WHETHER and WHEN updates were triggered.
\\

\bottomrule
\end{tabular}
\caption{Per-scene evaluation criteria for World Model metrics. Part 3 of 5.}
\label{tab:prompt-world-metrics3}
\end{table*}
\clearpage

\begin{table*}[htbp]
\small
\begin{tabular}{>{\raggedright\arraybackslash}p{0.12\textwidth} !{\vrule} p{0.83\textwidth}}
\toprule
\multicolumn{2}{c}{\textbf{Per-scene Evaluation Criteria for World Model}} \\
\midrule
\textbf{Location Update Sensitivity (LUS)} &

\textit{(Continuing from the previous Table)}

\textbf{Handling Redundant Updates (triggered but content unchanged)}: Determine the root cause --- if no persistent physical change occurred before that update, the trigger itself was wrong $\to$ this is a LUS issue (penalize here); if a persistent physical change did occur but the content failed to reflect it (i.e., trigger was correct but content update failed), that is an LSA issue $\to$ do NOT penalize here.

\textbf{Interaction Count Consideration}: Take the number of interaction turns in this scene into account when scoring. Longer scenes expose the model to more update opportunities and temporary-versus-persistent event distinctions, increasing the risk of over-triggering, missing updates, or triggering at the wrong time. Therefore, maintaining mostly correct location-update timing over many turns should be interpreted in that context rather than judged as if the scene were short. Do not penalize length itself; focus on whether observed timing errors are substantial relative to the amount of evidence and number of update opportunities. For very short scenes, there may be limited evidence for assessing this metric, so avoid over-interpreting the absence of errors as strong evidence of capability.
\par\smallskip\textit{Evaluation input: Interactions and Location State Update Timeline} \\
\midrule
\textbf{Location State Accuracy (LSA)} &
Evaluate the ACCURACY of the location state content when updates occur. This metric focuses on WHETHER THE CONTENT IS CORRECT, not on whether the update should have been triggered (that belongs to Location Update Sensitivity, LUS).

You will see the complete update timeline (the state may be updated multiple times within a single scene).

\textbf{Handling Redundant Updates (triggered but content unchanged)}: If a persistent physical change did occur but the content failed to reflect it $\to$ this is an LSA issue (penalize here); if no persistent physical change occurred (the trigger itself was wrong) $\to$ this is a LUS issue (do NOT penalize here). Only evaluate updates where the content actually changed, unless the redundant update falls into the LSA category above.

\textbf{1. Spatial Consistency}
\begin{itemize}[nosep,leftmargin=*]
\item Is the updated spatial layout self-consistent (no contradictions in where things are)?
\item Do the updated sub-locations and their relationships make physical sense?
\item Example flaw: An entity listed as being in two different sub-locations simultaneously.
\end{itemize}

\textbf{2. Entity Accuracy}
\begin{itemize}[nosep,leftmargin=*]
\item Does the important entities list accurately reflect the important entities currently at the location?
\item Have important entities that arrived/departed been correctly added/removed?
\item Are entity states (conditions, positions) accurately described?
\item Note: Only track important, lasting entities. Trivial or transient details (e.g., character clothing, briefly mentioned objects) do not need to appear in the important entities list; not recording them is normal and should not be penalized. Example flaw: A destroyed building still listed as intact in the entities list.
\end{itemize}

\textbf{3. Scene Consistency}
\begin{itemize}[nosep,leftmargin=*]
\item Is the location description consistent with what happened in the scene?
\item Example flaw: After an intense fight, the description still reads "the room is spotless."
\end{itemize}

\textbf{4. Incremental Accuracy}
\begin{itemize}[nosep,leftmargin=*]
\item If multiple updates occurred, does each one build correctly on the previous state?
\item Are there contradictions between successive updates within the same scene?
\end{itemize}

\textbf{Scope Exclusion}: Do NOT evaluate the FORMAT of the location state output (JSON structure, field completeness), as that belongs to Instruction Compliance (IC). Do NOT evaluate whether the update SHOULD have been triggered, as that belongs to Location Update Sensitivity (LUS). Focus strictly on whether the CONTENT of each update is accurate.

\textbf{Interaction Count Consideration}: Take the number of interaction turns in this scene into account when scoring. Longer scenes expose the model to more state changes, incremental updates, and opportunities for spatial or entity-level inconsistency. Therefore, maintaining mostly accurate location-state content over many turns should be interpreted in that context rather than judged as if the scene were short. Do not penalize length itself; focus on whether observed content errors are substantial relative to the amount of evidence and number of state-change opportunities. For very short scenes, there may be limited evidence for assessing this metric, so avoid over-interpreting the absence of errors as strong evidence of capability.
\par\smallskip\textit{Evaluation input: Interactions and Location State Update Timeline} \\
\midrule
\textbf{Instruction Compliance - World (IC\_world)} &
The World Model is responsible for four tasks: (1) selecting the cast of characters for each scene, (2) choosing a location and generating a scenario, (3) selecting the next speaker each turn, and (4) updating global/location state. Evaluate whether ALL outputs follow the expected rules.

Note: The data you see has already been parsed from the model's raw JSON output. If parsing failed entirely (e.g., invalid JSON), that error is handled separately via an error penalty. Your job is to evaluate the FORMAT and COMPLIANCE of the successfully parsed content shown to you.

\textbf{Area 1: Involved Characters}

Involved Characters: CharA, CharB, CharC.
This is the parsed result of the cast selection task.

\textbf{1. Cast Selection Compliance}
\begin{itemize}[nosep,leftmargin=*]
\item Is the character list non-empty and reasonable in size for a scene?
\item Example flaw: An empty character list, or an absurdly large number of characters (e.g., 15+) that would make a scene unmanageable.
\end{itemize}

\textbf{Area 2: Scene Scenario}

Scene Scenario
(scenario text).
This is the parsed scenario from the location \& scenario selection task.

\\

\bottomrule
\end{tabular}
\caption{Per-scene evaluation criteria for World Model metrics. Part 4 of 5.}
\label{tab:prompt-world-metrics4}
\end{table*}
\clearpage

\begin{table*}[htbp]
\small
\begin{tabular}{>{\raggedright\arraybackslash}p{0.12\textwidth} !{\vrule} p{0.83\textwidth}}
\toprule
\multicolumn{2}{c}{\textbf{Per-scene Evaluation Criteria for World Model}} \\
\midrule
\textbf{Instruction Compliance - World (IC\_world)} & 

\textit{(Continuing from the previous Table)}

\textbf{2. Scenario Format Compliance}
\begin{itemize}[nosep,leftmargin=*]
\item Is the scenario a readable prose passage (not garbled, not containing raw JSON artifacts)?
\item Does the output actually look like a scene scenario (not a world state or other unrelated content)?
\item Example flaw: The scenario field contains raw JSON syntax or is clearly not a scenario description at all.
\end{itemize}

\textbf{Area 3: Speaker Selection Sequence}

Speaker Selection Sequence
Turn 0: CharA
Turn 1: CharB
Turn 2: CharA, CharC
Turn 3: Environment
This is the sequence of speaker selections made by the World Model throughout the scene.

\textbf{3. Speaker Selection Compliance}
\begin{itemize}[nosep,leftmargin=*]
\item Is each selected speaker one of the involved characters listed above (or "Environment")?
\item Are there any turns where an invalid or non-existent character name appears?
\item Are multi-character turns (e.g., "CharA, CharB") used sparingly and only for genuinely shared beats?
\item Example flaw: A speaker name that doesn't match any of the involved characters.
\end{itemize}

\textbf{Area 4: Global State Update Timeline}

The global state is a structured Markdown document describing the world's rules, norms, and conditions, organized under thematic section headers (e.g., \texttt{\#\#\# 1. Social Order \& Class} or \texttt{\textbf{Economy \& Trade}}), with bullet points or paragraphs under each section.

You will see one of two cases:

\textbf{Case A --- Update(s) occurred:}
\#\# Global State Update Timeline
Global State BEFORE scene:
(Markdown prose)

Global State Updated: \textbf{YES} (N update(s) during this scene)
(updated Markdown prose)

\textbf{Case B --- No update:}
\#\# Global State Update Timeline
Global State BEFORE scene:
(Markdown prose)

Global State Updated: \textbf{NO} (unchanged)

\textbf{4. Global State Format}
\begin{itemize}[nosep,leftmargin=*]
\item Is the global state presented as readable Markdown prose (not garbled, not containing raw JSON artifacts like \texttt{{"key": "value"}})?
\item Does the output actually look like a global world state description (not a character profile, scenario, or other unrelated content)?
\item Example flaw: The global state text contains raw JSON syntax or is clearly not a world state description.
\end{itemize}

\textbf{Area 5: Location State Update Timeline}

The location state supports two structural variants depending on the location:

\textbf{Variant 1 --- With Sub Locations:} The state has a "Description" line, followed by "Sub Locations" formatted as \texttt{[SubLocationName]: description}, each containing "Important Entities" formatted as \texttt{- EntityName: entity state}.

\textbf{Variant 2 --- Without Sub Locations:} The state has a "Description" line, followed directly by "Important Entities" formatted as \texttt{- EntityName: entity state} (no sub-location grouping).

You will see one of two update cases:

\textbf{Case A --- Update(s) occurred (Variant 1 example):}
\#\# Location State Update Timeline: LocationName
Location State BEFORE scene:
Description: ...
[SubLocationName]: ...
\begin{itemize}[nosep,leftmargin=*]
\item EntityName: entity state
\end{itemize}

Location State Updated: \textbf{YES} (N update(s) during this scene)
Description: ...
[SubLocationName]: ...
\begin{itemize}[nosep,leftmargin=*]
\item EntityName: entity state
\end{itemize}

\textbf{Case A --- Update(s) occurred (Variant 2 example):}
\#\# Location State Update Timeline: LocationName
Location State BEFORE scene:
Description: ...
\begin{itemize}[nosep,leftmargin=*]
\item EntityName: entity state
\end{itemize}

Location State Updated: \textbf{YES} (N update(s) during this scene)
Description: ...
\begin{itemize}[nosep,leftmargin=*]
\item EntityName: entity state
\end{itemize}

\textbf{Case B --- No update:}
\#\# Location State Update Timeline: LocationName
Location State BEFORE scene:
Description: ...
(entities and/or sub-locations as applicable)

Location State Updated: \textbf{NO} (unchanged)

\textbf{5. Location State Format}
\begin{itemize}[nosep,leftmargin=*]
\item Does the location state follow one of the two expected structural variants?
\item Variant 1 (with Sub Locations): Description $\to$ Sub Locations (each with Important Entities)
\item Variant 2 (without Sub Locations): Description $\to$ Important Entities directly
\item If sub-locations are present, are they structured with names, descriptions, and entities with states?
\item If no sub-locations, are Important Entities listed directly after the Description with names and states?
\item Does the output actually look like a location state (not a global state or other unrelated content)?
\item Example flaw: Entities listed without states, or the output is clearly not a location state at all.
\end{itemize}

\textbf{6. No Overstepping}
\begin{itemize}[nosep,leftmargin=*]
\item Do world state updates (both global and location) avoid generating character dialogue or controlling character behavior?
\item Are updates purely descriptive of the environment, not prescriptive of character actions?
\item Example flaw: The location state includes "Character A decides to leave" --- overstepping into character territory.
\end{itemize}
\par\smallskip\textit{Evaluation input: Involved Characters, Scenario, Speaker Selection Sequence, Global State Update Timeline, and Location State Update Timeline} \\

\bottomrule
\end{tabular}
\caption{Per-scene evaluation criteria for World Model metrics. Part 5 of 5.}
\label{tab:prompt-world-metrics5}
\end{table*}


\begin{table*}[htbp]
\small
\begin{tabular}{>{\raggedright\arraybackslash}p{0.12\textwidth} !{\vrule} p{0.83\textwidth}}
\toprule
\multicolumn{2}{c}{\textbf{Cross-scene Evaluation Criteria for PES and SCC}} \\
\midrule
\textbf{Profile Evolution Smoothness (PES)} &
Evaluate the SMOOTHNESS of this character's cross-scene evolution by considering BOTH the profile trajectory and the hidden tracker trajectory together. This metric focuses on whether the evolution TRAJECTORY is smooth and coherent over time, NOT on whether individual scene updates are accurate (that belongs to Profile Update Fidelity, PUF). Specifically:

\textbf{1. Gradualness}
\begin{itemize}[nosep,leftmargin=*]
\item Do changes in personality, attitude, and relationships go through reasonable transitional stages across scenes? Does the hidden tracker preserve intermediate stages before they eventually become profile changes? Are there abrupt jumps where a character goes from one extreme to another without intermediate signals in either profile or tracker?
\item Example flaw: Scene 3 shows mild doubt, but Scene 4 profile suddenly declares complete distrust with no accumulated intermediate signals.
\end{itemize}

\textbf{2. Magnitude Proportionality Across the Timeline}
\begin{itemize}[nosep,leftmargin=*]
\item Looking at the full evolution timeline, is the magnitude of each profile change proportional to what happened in the corresponding scene(s)? Are there scenes where nothing significant happened but the profile changed dramatically, or vice versa?
\item Example flaw: A profile undergoes a massive rewrite after a routine conversation, while remaining unchanged after a life-altering event.
\end{itemize}

\textbf{3. Directional Coherence Across Profile + Tracker}
\begin{itemize}[nosep,leftmargin=*]
\item Do the profile and hidden tracker point in compatible directions over time? When tracker signals accumulate enough to justify a profile update, does that transition feel traceable? Are there contradictions where the tracker implies one trajectory while the profile jumps to another without explanation?
\item Example flaw: The tracker repeatedly records growing trust, but the next profile revision abruptly claims deepening hostility with no causal basis.
\end{itemize}

\textbf{Scope Exclusion}: Do NOT evaluate whether individual scene updates are accurate or whether the right things were captured --- that belongs to Profile Update Fidelity (PUF). Focus strictly on whether the OVERALL TRAJECTORY across all scenes is smooth, gradual, and directionally coherent.

\textbf{Scene Count Consideration}: Take the total number of scenes into account when scoring. Longer trajectories expose the model to more profile and hidden-tracker updates, accumulated signals, and opportunities for drift or contradiction. Therefore, maintaining mostly smooth and coherent profile evolution over many scenes should be interpreted in that context rather than judged as if the trajectory were short. Do not penalize length itself; focus on whether observed trajectory errors are substantial relative to the amount of evidence and number of update opportunities. For very short trajectories, there may be limited evidence for assessing this metric, so avoid over-interpreting the absence of errors as strong evidence of capability.

Review the complete profile evolution timeline and hidden tracker timeline together, and assess whether they form one coherent and smooth character development arc.
\par\smallskip\textit{Evaluation input: Initial Profile, and Scene-by-Scene Evolution Timeline (Motivation, Scenario, Scene Summary, Description, Hidden Tracker, Profile After Scene)} \\
\midrule
\textbf{Scene Continuity \& Coherence (SCC)} &
Evaluate the overall narrative coherence across all scenes. You are provided with each scene's summary (including location, scenario, involved characters, and what happened) to assess cross-scene coherence.

\textbf{1. Narrative Arc}
\begin{itemize}[nosep,leftmargin=*]
\item Do consecutive scenes form a directional narrative progression? Is there a discernible story arc (setup $\to$ rising action $\to$ climax $\to$ resolution)? Do scenes build upon each other rather than being disconnected episodes? Example flaw: Scenes that feel like random, unconnected vignettes with no overarching story.
\end{itemize}

\textbf{2. Scene Transitions}
\begin{itemize}[nosep,leftmargin=*]
\item Do location choices and scene descriptions naturally connect? Are transitions between scenes logical (characters move to locations that make sense)? Is there narrative justification for each scene's setting? Example flaw: Characters teleporting between distant locations without travel time or explanation.
\end{itemize}

\textbf{3. Pacing}
\begin{itemize}[nosep,leftmargin=*]
\item Is the story pacing appropriate across the full simulation? Are there sections that feel too rushed or too slow? Do important plot points get adequate development time? Example flaw: The climactic confrontation happening in Scene 2 of 10, with the remaining 8 scenes being anticlimactic.
\end{itemize}

\textbf{4. Thread Management}
\begin{itemize}[nosep,leftmargin=*]
\item Are narrative threads introduced, developed, and resolved (or intentionally left open)? Are there abandoned plot threads that were set up but never followed through? Example flaw: A mystery introduced in Scene 1 that is never mentioned again in any subsequent scene.
\end{itemize}

\textbf{Scene Count Consideration}: 
Take the total number of scenes into account when scoring. Longer simulations expose the model to more narrative arcs, pacing decisions, scene transitions, and thread-management challenges. Maintaining mostly coherent narrative development over many scenes should be interpreted in that context rather than judged as if the simulation were short. Do not penalize length itself; focus on whether observed coherence errors are substantial relative to the amount of evidence and number of long-range narrative dependencies. For very short simulations, there may be limited evidence for assessing this metric, so avoid over-interpreting the absence of errors as strong evidence of capability.
\par\smallskip\textit{Evaluation input: the complete scene sequence with Scenario, Involved Characters, and Scene Summary for each scene} \\

\bottomrule
\end{tabular}
\caption{Cross-scene evaluation criteria for PES and SCC.}
\label{tab:prompt-cross}
\end{table*}

\begin{table*}[htbp]
\small
\begin{tabular}{>{\raggedright\arraybackslash}p{0.12\textwidth} !{\vrule} p{0.83\textwidth}}
\toprule
\multicolumn{2}{c}{\textbf{Human Evaluation: Annotator Instructions}} \\
\midrule

\textbf{Task Overview} &
You will evaluate \ours{}, an interactive fiction simulation system in which LLMs drive characters from classic novels through AI-generated storylines. The simulation is jointly produced by two model components:

\textbf{World Model}, which directs the story and models the world:
\begin{itemize}[nosep,leftmargin=*]
\item \textit{Scene Planning}: selects the cast of characters for each scene based on their profiles and current motivations; then chooses a location and writes a dramatic scenario.
\item \textit{Speaker Management}: within each scene, decides who speaks next (or when to insert an environment narration / end the scene). May also select character groups for simultaneous actions.
\item \textit{World State Maintenance}: after key events, updates two kinds of state:
  \begin{itemize}[nosep,leftmargin=*]
  \item \textbf{Global State}: free-form Markdown prose; the model may freely choose which aspects to describe (e.g.\ era background, social atmosphere, major events).
  \item \textbf{Location State}: structured JSON in one of two variants: (a)~\textit{without sub-locations}: fields ``name'', ``description'', and ``important\_entities'' (each entity has a ``name'' and ``state''); (b)~\textit{with sub-locations}: fields ``name'', ``description'', and a ``sub\_locations'' list, where each sub-location contains ``name'', ``description'', and ``important\_entities''.
  \end{itemize}
\end{itemize}

\textbf{Character Agent}, which plays each character:
\begin{itemize}[nosep,leftmargin=*]
\item \textit{Motivation Generation}: before each scene, generates a scene-specific motivation for every participating character.
\item \textit{Interaction Generation}: produces each character's turn. The output uses three format markers: \textbf{[brackets]} for the character's inner thoughts, \textbf{plain text} for spoken dialogue, and \textbf{(parentheses)} for visible physical actions.
\item \textit{Profile \& State Updates}: after each scene, updates two records for each character:
  \begin{itemize}[nosep,leftmargin=*]
  \item \textbf{Profile}: free-form Markdown prose recording established, persistent character changes (e.g.\ personality traits, relationships, significant experiences, current goals). The model may freely choose which aspects to cover.
  \item \textbf{Hidden Tracker}: Markdown prose recording sub-threshold psychological signals that have not yet risen to the level of a permanent profile change, such as accumulating emotional stress, unresolved inner conflicts, subtle impressions from a conversation, or seed events that may later trigger a character shift. These signals accumulate across scenes and, once sufficient, trigger a formal profile update.
  \end{itemize}
\end{itemize}

The simulation loop for each scene proceeds as follows:
\begin{enumerate}[nosep,leftmargin=*]
\item World Model selects the cast of characters.
\item World Model chooses a location and writes the scenario.
\item Character Agent generates a scene-specific motivation for each character.
\item Repeat until the scene ends:
  \begin{enumerate}[nosep,leftmargin=*,label=\alph*.]
  \item World Model selects the next speaker (or inserts an environment narration).
  \item Character Agent generates that character's interaction turn.
  \item World Model optionally updates the global/location state.
  \end{enumerate}
\item Character Agent updates each character's profile and hidden tracker.
\end{enumerate}

Within a single simulation, both components use the same underlying LLM. You will compare two different LLMs (Model A vs.\ Model B, presented in randomized order) and judge which performs better on each evaluation dimension. \\

\midrule
\textbf{Available Materials} &
You will receive a folder with the following structure:
\begin{lstlisting}
output/
+-- readme.txt
+-- annotation_{modelA}_vs_{modelB}.xlsx    (*3)
+-- reference/
    +-- book_scenes/    (one .txt per book)
    +-- test_inputs/    (one .txt per sample)
\end{lstlisting}
For each sample, you will receive:
\begin{itemize}[nosep,leftmargin=*]
\item \textbf{Annotation Excel file} (\texttt{annotation\_\{modelA\}\_vs\_\{modelB\}.xlsx}): contains complete scene-by-scene outputs for both models, including: (1)~interaction content (character dialogue, thoughts, actions); (2)~character state changes per scene (profile updates, short description, scene motivation, hidden tracker); (3)~world state changes (global state and location state updates, annotated with the interaction turn after which they occurred). The file also contains the scoring rows where you fill in your 0--100 scores for each dimension; the spreadsheet will automatically compute the average score and the comparison result from your inputs.
\item \textbf{Reference files}: (a)~\texttt{book\_scenes/\{book\_title\}.txt}, the source novel's scene content (scenario, character interactions, chapter info), useful for verifying character speaking style and story context; (b)~\texttt{test\_inputs/sample\_XXXXXX.txt}, the identical starting point given to both models (world state, all character profiles with motivations and hidden trackers, and the previous scene for narrative context).
\end{itemize} \\

\bottomrule
\end{tabular}
\caption{Human evaluation annotator instructions. Part 1 of 2.}
\label{tab:human-eval-instr1}
\end{table*}

\begin{table*}[htbp]
\small
\begin{tabular}{>{\raggedright\arraybackslash}p{0.12\textwidth} !{\vrule} p{0.83\textwidth}}
\toprule
\multicolumn{2}{c}{\textbf{Human Evaluation: Annotator Instructions} \textit{(continued)}} \\
\midrule

\textbf{Scoring Method} &
Use a 0--100 scale for each dimension. The absolute scores serve only as personal note-taking aids to help you organize your thoughts. The final collected annotation is the \textbf{comparison result} per dimension (A better / B better / Tie). You need not agonize over exact numeric values; only the relative judgment matters. \\

\midrule
\textbf{Annotation Workflow} &
\textbf{Recommended procedure for each sample:}
\begin{enumerate}[nosep,leftmargin=*]
\item \textbf{Read metadata}: check the book title, number of generated scenes, and stop reason (``completed'', ``error'', or ``max scenes reached'') for both models. Also note the Sample Index for locating the corresponding reference files.
\item \textbf{Consult reference files} (recommended): (a)~read the previous scene in the initial state file to understand the narrative context at simulation start; (b)~skim the initial world state and character profiles to understand the shared starting point.
\item \textbf{Read both models' outputs}: go through all scenes for Model A and Model B to form an overall impression. Focus on: whether dialogue is natural and in-character; whether the story progresses meaningfully; whether state updates are reasonable; whether there are obvious format errors or AI artifacts.
\item \textbf{Score each dimension}: Write down a 0--100 score for each dimension for both models. Remember, these scores are only personal notes to help you organize your thoughts. Only the final comparison result (A better / B better / Tie) per dimension matters.
\end{enumerate}

\textbf{Key guidelines:}
\begin{itemize}[nosep,leftmargin=*]
\item Model A/B order is randomized per sample; judge purely on output quality.
\item Do NOT prefer longer outputs; faithfulness to the original book's style should be rated higher.
\item For CC: you may look up the character's lines in the original book scenes for speaking style reference.
\end{itemize} \\

\midrule
\textbf{Evaluation Dimensions} &
You will evaluate 10 dimensions: \textbf{CC} (Character Consistency), \textbf{EQ} (Evolution Quality), \textbf{EG} (Environment Grounding), \textbf{IQ} (Interaction Quality), \textbf{MG} (Motivation Generation), \textbf{IC\textsubscript{char}} (Instruction Compliance, Character) for the Character Agent; and \textbf{SP} (Scene Planning), \textbf{SM} (Speaker Management), \textbf{WSM} (World State Maintenance), \textbf{IC\textsubscript{world}} (Instruction Compliance, World) for the World Model. Each dimension contains multiple sub-metrics; their definitions and detailed scoring criteria are identical to those used by the LLM-as-Judge, as presented in Tables~\ref{tab:prompt-char-metrics1}--\ref{tab:prompt-char-metrics6}, Tables~\ref{tab:prompt-world-metrics1}--\ref{tab:prompt-world-metrics5}, and Table~\ref{tab:prompt-cross}. \\

\bottomrule
\end{tabular}
\caption{Human evaluation annotator instructions. Detailed scoring criteria for each sub-metric are identical to the LLM-as-Judge criteria presented in the preceding tables. Part 2 of 2.}
\label{tab:human-eval-instr2}
\end{table*}

\end{document}